\documentclass[journal]{IEEEtran}
\usepackage[comma,numbers,square,sort&compress]{natbib}
\usepackage{subfigure}
\usepackage[usenames,dvipsnames]{color}
\usepackage{colortbl}
\usepackage{graphicx}
\usepackage{float}
\usepackage{amsmath,latexsym,amssymb,amsthm,array,amsfonts,algorithm,algpseudocode,booktabs,graphicx,subfigure,multirow,cuted,stfloats}
\usepackage[sort&compress]{natbib}
\usepackage{url}
\usepackage{multirow}
\usepackage{lscape}
\usepackage{makecell}
\usepackage{slashbox}
\usepackage{bm}
\definecolor{mygray}{gray}{0.8}
\theoremstyle{definition}
\newcommand{\tabincell}[2]{\begin{tabular}{@{}#1@{}}#2\end{tabular}}

\usepackage{array}
\newlength\savewidth
\newcommand\shline{\noalign{\global\savewidth\arrayrulewidth
                            \global\arrayrulewidth 1.2pt}%
                   \hline
                   \noalign{\global\arrayrulewidth\savewidth}}

\newtheorem{remark}{\it Remark}

\begin{document}
\hyphenpenalty=5000
\tolerance=1200
%
\title{Differential Evolution with Event-Triggered Impulsive Control}

\author{Wei Du,
        Sunney Yung Sun Leung,
        Yang Tang,
        and Athanasios V. Vasilakos

\thanks{W. Du is with the Key Laboratory of Advanced Control and Optimization for Chemical Processes, Ministry of Education, East China University of Science and Technology, Shanghai 200237, China and the Institute of Textile and Clothing, The Hong Kong Polytechnic University, Hong Kong, China (e-mail: duwei0203@gmail.com).}

\thanks{S. Y. S. Leung is with the Institute of Textile and Clothing, The Hong Kong Polytechnic University, Hong Kong, China (e-mail: sunney.leung@polyu.edu.hk).}

\thanks{Y. Tang is with the Key Laboratory of Advanced Control and Optimization for Chemical Processes, Ministry of Education, East China University of Science and Technology, Shanghai 200237, China (e-mail: tangtany@gmail.com; yangtang@ecust.edu.cn).}

\thanks{Athanasios V. Vasilakos is with the Department of Computer Science, Electrical and Space Engineering, Lulea University of Technology, Lulea 97187, Sweden (e-mail: vasilako@ath.forthnet.gr).}}

\markboth{Preprint submitted to arXiv}%
{Shell \MakeLowercase{\textit{et al.}}: Bare Demo of IEEEtran.cls
for Journals}

\maketitle

\begin{abstract}
Differential evolution (DE) is a simple but powerful evolutionary algorithm, which has been widely and successfully used in various areas. In this paper, an event-triggered impulsive control scheme (ETI) is introduced to improve the performance of DE. Impulsive control, the concept of which derives from control theory, aims at regulating the states of a network by instantly adjusting the states of a fraction of nodes at certain instants, and these instants are determined by event-triggered mechanism (ETM). By introducing impulsive control and ETM into DE, we hope to change the search performance of the population in a positive way after revising the positions of some individuals at certain moments.
At the end of each generation, the impulsive control operation is triggered when the update rate of the population declines or equals to zero. In detail, inspired by the concepts of impulsive control, two types of impulses are presented within the framework of DE in this paper: stabilizing impulses and destabilizing impulses. Stabilizing impulses help the individuals with lower rankings instantly move to a desired state determined by the individuals with better fitness values. Destabilizing impulses randomly alter the positions of inferior individuals within the range of the current population. By means of intelligently modifying the positions of a part of individuals with these two kinds of impulses, both exploitation and exploration abilities of the whole population can be meliorated. In addition, the proposed ETI is flexible to be incorporated into several state-of-the-art DE variants. Experimental results over the CEC 2014 benchmark functions exhibit that the developed scheme is simple yet effective, which significantly improves the performance of the considered DE algorithms.

\end{abstract}


%
\IEEEpeerreviewmaketitle

\section{Introduction}
Differential evolution (DE), firstly proposed by Storn and Price \cite{storn1995differential,storn1997differential}, has proven to be a reliable and powerful population-based evolutionary algorithm for global numerical optimization. Over the past decade, different variants of DE have been proposed to handle complicated optimization problems in various application fields \cite{plagianakos2008review}, such as engineering design \cite{kim2007differential}, image processing \cite{sarkar2013multi}, data mining \cite{das2008automatic}, robot control \cite{neri2010memetic}, and so on.

Generally, DE employs three main operators: mutation, crossover, and selection at each generation for the population production \cite{neri2010recent,das2011differential,ghosh2012convergence}. The mutation operator provides the individuals with a sudden change or perturbation, which helps explore the search space. In order to increase the diversity of the population, the crossover operator is implemented after the mutation operation. The selection operator chooses the better one between a parent and its offspring, which guarantees that the population never deteriorates. In addition to these three basic operators, there are three control parameters which greatly influence the performance of DE: the mutation scale factor \emph{F}, the crossover rate \emph{CR}, and the population size \emph{NP}. Most of the current research on DE has focused on four aspects to enhance the performance of DE: developing novel mutation operators \cite{tang2014differential,islam2012adaptive,zhang2009jade,gong2013differential,guo2014improving,cai2013differential,epitropakis2011enhancing,wangliao2014differential,das2009differential}, designing new parameter control strategies \cite{tang2014differential,islam2012adaptive,zhang2009jade,brest2006self,zhu2013adaptive,yu2014differential,gong2011enhanced,teo2006exploring,tirronen2009differential}, improving crossover operator \cite{islam2012adaptive,wang2012enhancing,wang2014differential,guo2015enhancing}, and pooling multiple mutation strategies \cite{mallipeddi2011differential,wang2011differential,qin2009differential,dorronsoro2011improving,wu2016differential}. These four categories of research on DE are described in detail as follows. 1) In recent years, some efficient mutation operators have been presented and incorporated into the DE framework. For instance, Zhang and Sanderson \cite{zhang2009jade} proposed a new mutation strategy ``DE/current-to-\emph{p}best" to improve the performance of the basic DE. Gong and Cai \cite{gong2013differential} developed a ranking-based mutation operator to assign better individuals to lead the population. Guo \emph{et} \emph{al.} \cite{guo2014improving} presented a successful-parent-selecting method, which adapts the selection of parents when stagnation is occurred. 2) Various parameter control schemes have been introduced to the DE algorithm. In \cite{brest2006self} and \cite{zhang2009jade}, \emph{F} and \emph{CR} can be evolved during the evolution of the population. In \cite{zhu2013adaptive}, an adaptive population tuning scheme was proposed to reassign computing resources in a more reasonable way. 3) Some researchers have made efforts to optimize the conventional crossover strategy. For example, Islam \emph{et} \emph{al.} \cite{islam2012adaptive} incorporated a greedy parent selection strategy with the traditional binomial crossover scheme to develop a \emph{p}-best crossover operation. Guo and Yang \cite{guo2015enhancing} utilized eigenvectors of covariance matrix to make the crossover rotationally invariant, which generates a better search behavior. 4) Several DE variants have been put forward, which employs more than one mutation operator to breed new solutions, such as EPSDE \cite{mallipeddi2011differential}, CoDE \cite{wang2011differential}, SaDE \cite{qin2009differential}, and so on.

Despite numerous efforts on improving DE from the above four aspects, there are some DE variants which take advantage of ideas from other disciplines. For instance, Rahnamayan \emph{et} \emph{al.} \cite{rahnamayan2008opposition} presented opposition-based DE (ODE), which adopts opposition-based learning, a new scheme in machine intelligence, to speed up the convergence rate of DE. Laelo and Ali \cite{kaelo2007differential} made use of the attraction-repulsion concept in electromagnetism to boost the performance of the original DE. Vasile \emph{et} \emph{al.} \cite{vasile2011inflationary} proposed a novel DE, which is inspired by discrete dynamical systems. These improvements on DE enlighten us to look through techniques in other areas, which might be introduced to the development of DE variants.

On another research frontier, as an important component in control theory, impulsive control has attracted much attention in recent years due to its high efficiency. As exemplified in \cite{zhang2014synchronization,tang2015leader}, impulsive effects can be detected in various dynamical systems, like communication networks, electronic systems, biological networks, and so on. Besides, impulsive control is able to manipulate the states of a network to a desired value by adding impulsive signals to some specific nodes at certain instants. In addition, another effective technique, event-triggered mechanism (ETM), has also been widely utilized \cite{tabuada2007event,wang2011event,heemels2013periodic,tang2016robust} in control theory. In ETM, the state of the controller is updated when the system's state exceeds a given threshold. By integrating ETM into impulsive control, the operation of impulsive control can only be activated when some predefined events are triggered. This way, ETM avoids the periodical execution of impulsive control, which efficiently saves computational resources.

Taking a look at how DE works in an optimization problem, the movement of the population in the evolution process can be treated as a complicated multi-agent system in control theory, where individuals in the population can be regarded as nodes in a network.
On one hand, in original DE algorithms and some popular DE variants, it may take a long time for certain individuals to reach the desired positions. For instance, the ``\emph{p}best" individual is utilized to guide the search of other individuals in JADE \cite{zhang2009jade}. However, this operation is carried out at each generation and forces the individuals to approach the desired state slowly, which deteriorates the search performance of the population in the limited computational resources.
On the other hand, in many DE variants, like jDE \cite{brest2006self}, JADE \cite{zhang2009jade}, CoDE \cite{wang2011differential} \emph{et} \emph{al.}, the diversity of the population is maintained only by mutation and crossover, which are indirect.
Inspired by how impulsive control manipulates a dynamical system, we introduce the concept of impulsive control into the design of DE, aiming at increasing the search efficiency and the diversity of the population by instantly letting selected individuals move close to the desired positions.
Besides, when DE is used for an optimization problem, the computational resources are often limited, measured by the maximum number of function evaluations (MAX\_FES). Therefore, it is reasonable to trigger the instantaneous movement of certain individuals by some predefined events, which follows the idea of ETM.

Motivated by the above discussion, by integrating ETM into impulsive control, we introduce an event-triggered impulsive control scheme (ETI) for DE in this paper. Similar to adjusting the states of some nodes in dynamical systems in control theory, impulsive control revises the positions of a fraction of population at certain moments, the purpose of which is to positively change the evolution state of the whole population. In detail, two varieties of impulses: stabilizing impulses and destabilizing impulses are presented to fit into the framework of DE.
In addition, based on both the fitness value and the number of consecutive stagnation generation, a novel measure $\emph{R}_\emph{i}$ is developed to pick the individuals to be injected with impulsive controllers.
When the update rate (\emph{UR}) of the population begins to diminish or reduces to zero, the individuals with large values of $\emph{R}_\emph{i}$ will be injected with impulsive controllers. Stabilizing impulses are adopted to force a number of individuals with lower rankings in the current population to get close to the individuals with better fitness values, which increases the exploitation ability of DE. Besides, destabilizing impulses are considered to randomly adjust the positions of inferior individuals within the range of the current population, which improves the exploration capability of DE. The major contributions of this paper are mainly threefold: 1) an event-triggered impulsive control scheme is introduced into the DE framework, which aims to improve the search ability of DE; 2) two kinds of impulses, stabilizing and destabilizing impulses, are developed respectively, to enhance the exploitation and exploration performance of DE; 3) the proposed scheme is simple but effective, which can improve the performance of most of the considered representative DE algorithms in this paper. It is worth pointing out that we have done some preliminary work in \cite{du2015improving}, in which an impulsive control framework (IPC) is proposed for DE. IPC differs from ETI in the following three major aspects: 1) ETI includes ETM to identify when the individuals should be injected with impulsive controllers, while no ETM is used in IPC. 2) Two types of impulses, stabilizing and destabilizing impulses, are proposed in ETI, while only stabilizing impulses are presented in IPC. 3) ETI adopts ranking assignment and an adaptive mechanism (described in Section III) to select the individuals taking impulsive control, while IPC just utilizes a non-adaptive piecewise threshold function for choosing the individuals.

This paper is organized as follows. In Section II, the original DE and the concepts of ETM and impulsive control are introduced. Our proposed scheme ETI is presented in Section III. Experimental results are reported in Section IV. Finally, concluding remarks are made in Section V.

\section{Related Work}
In this section, we firstly introduce the original DE. Then the concepts of event-triggered mechanism and impulsive control are briefly outlined.

A single objective optimization problem can be formulated as follows (without any loss of generality, in this paper, a minimization problem is considered with a decision space $\Omega$):
\begin{equation}
\textrm{minimize}\hspace{0.3cm} f(\textbf{x}), \hspace{0.3cm} \textbf{x}\in\Omega,
\label{eq1}
\end{equation}
where $\Omega$ is a decision space, $\textbf{x}=[x_1,x_2,...,x_D]^{T}$ is a decision vector, and \emph{D} is the dimension size, representing the number of the decision variables involved in the problem. For each variable $x_j$, it should obey a boundary constraint:
\begin{equation}
L_j\leq x_j \leq U_j, \hspace{0.3cm} j=1,2,...,D,
\label{eq2}
\end{equation}
where $L_j$ and $U_j$ are the lower and upper bounds for the \emph{j}th dimension, respectively.

\subsection{DE}
DE is a population-based evolutionary algorithm for a numerical optimization problem. It initializes a population of \emph{NP} individuals in a \emph{D}-dimensional search space. Each individual represents a potential solution to the optimization problem. After initialization, at each generation, three operators: mutation, crossover and selection are employed to generate the offspring for the current population.
\subsubsection{Mutation}
Mutation is the most consequential operator in DE. Each vector $\textbf{x}_{\emph{i,G}}$ in the population at the \emph{G}th generation is called \emph{target} vector. A mutant vector called \emph{donor} vector is obtained through the differential mutation operation. For simplicity, the notation ``DE/\emph{a}/\emph{b}" is used to represent different mutation operators, where ``DE" denotes the differential evolution, ``\emph{a}" stands for the base vector, and ``\emph{b}" indicates the number of difference vectors utilized. In DE, there are six mutation operators that are most widely used:

i) ``DE/rand/1"
\begin{equation}
\textbf{v}_{\emph{i,G}}=\textbf{x}_{r_1,\emph{G}}+F\cdot(\textbf{x}_{r_2,\emph{G}}-\textbf{x}_{r_3,\emph{G}}),
\label{eq3}
\end{equation}

ii) ``DE/rand/2"
\begin{equation}
\begin{split}
\textbf{v}_{\emph{i,G}}=\textbf{x}_{r_1,\emph{G}}+F\cdot(\textbf{x}_{r_2,\emph{G}}-\textbf{x}_{r_3,\emph{G}})\\+F\cdot(\textbf{x}_{r_4,\emph{G}}-\textbf{x}_{r_5,\emph{G}}),
\end{split}
\label{eq4}
\end{equation}

iii) ``DE/best/1"
\begin{equation}
\textbf{v}_{\emph{i,G}}=\textbf{x}_{\emph{best,G}}+F\cdot(\textbf{x}_{r_1,\emph{G}}-\textbf{x}_{r_2,\emph{G}}),
\label{eq5}
\end{equation}

iv) ``DE/best/2"
\begin{equation}
\begin{split}
\textbf{v}_{\emph{i,G}}=\textbf{x}_{\emph{best,G}}+F\cdot(\textbf{x}_{r_1,\emph{G}}-\textbf{x}_{r_2,\emph{G}})\\+F\cdot(\textbf{x}_{r_3,\emph{G}}-\textbf{x}_{r_4,\emph{G}}),
\end{split}
\label{eq6}
\end{equation}

v) ``DE/current-to-best/1"
\begin{equation}
\begin{split}
\textbf{v}_{\emph{i,G}}=\textbf{x}_{\emph{i,G}}+F\cdot(\textbf{x}_{\emph{best,G}}-\textbf{x}_{\emph{i,G}})\\+F\cdot(\textbf{x}_{r_1,\emph{G}}-\textbf{x}_{r_2,\emph{G}}),
\end{split}
\label{eq7}
\end{equation}

vi) ``DE/current-to-rand/1"
\begin{equation}
\begin{split}
\textbf{u}_{\emph{i,G}}=\textbf{x}_{\emph{i,G}}+K\cdot(\textbf{x}_{r_1,\emph{G}}-\textbf{x}_{\emph{i,G}})\\+\hat{F}\cdot(\textbf{x}_{r_2,\emph{G}}-\textbf{x}_{r_3,\emph{G}}),
\end{split}
\label{eq8}
\end{equation}
where $\textbf{x}_{\emph{best,G}}$ specifies the best individual in the current population; $r_1,r_2,r_3,r_4$ and $r_5$ $\in \{1,2,...,\emph{NP}\}$, and $r_1\neq r_2\neq r_3\neq r_4\neq r_5\neq i$. The parameter $\emph{F}>0$ is called \emph{scaling factor}, which scales the difference vector. It is worth mentioning that (\ref{eq8}) shows the rotation-invariant mutation \cite{price1999introduction}. \emph{K} is the combination coefficient, which should be selected with a uniform random distribution from [0, 1] and $\hat{\emph{F}}=K\cdot \emph{F}$. Since ``DE/current-to-rand/1" contains both mutation and crossover, it is not necessary for the offspring to go through the crossover operation.

\subsubsection{Crossover}
After mutation, a binomial crossover operation is implemented to generate the \emph{trial} vector $\textbf{u}_{i}=[u_{i1},u_{i2},...,u_{\emph{iD}}]^{T}$:
\begin{equation}
 u_{\emph{ij,G}}=\left\{
\begin{array}{ll}
v_{\emph{ij,G}},       & \textrm{if} \hspace{0.1cm} \textrm{rand}(0,1)\leq \emph{CR}  \hspace{0.1cm} \textrm{or} \hspace{0.1cm} j=j_{rand},\\
x_{\emph{ij,G}},       & \textrm{otherwise},
\end{array} \right.
\label{eq9}
\end{equation}
where rand$(0,1)$ is a uniform random number in the range $[0,1]$. $\emph{CR}\in[0,1]$ is called \emph{crossover probability}, which determines how much the trial vector is inherited from the mutant vector. $\emph{j}_{\emph{rand}}$ is an integer randomly selected from $1$ to \emph{D} and newly generated for each $i$, which ensures at least one dimension of the trial vector will be different from the corresponding target vector. If $u_{\emph{ij,G}}$ is out of the boundary, it will be reinitialized in the range $[L_j,U_j]$.

\subsubsection{Selection}
The selection operator employs a one-to-one swapping strategy, which picks the better one from each pair of $\textbf{x}_{\emph{i,G}}$ and $\textbf{u}_{\emph{i,G}}$ for the next generation:
\begin{equation}
 \textbf{x}_{\emph{i,G+1}}=\left\{
\begin{array}{ll}
\textbf{u}_{\emph{i,G}},       & \textrm{if} \hspace{0.1cm} f(\textbf{u}_{\emph{i,G}})\leq f(\textbf{x}_{\emph{i,G}}),\\
\textbf{x}_{\emph{i,G}},       & \textrm{otherwise}.
\end{array} \right.
\label{eq10}
\end{equation}

\subsection{Event-triggered mechanism (ETM)}
Event-triggered mechanism (ETM) is an effective strategy in control theory that determines when the state of a controller is updated. Typically, a controller's state is independent of a system's state except at periodic instants. When the communication resource is insufficient, the traditional time-triggered paradigm may not be efficient. While in ETM, the state of the controller is revised only when a system's state exceeds a predefined threshold, or a specified event occurs. This way, ETM is able to reduce the amount of unnecessary communications. It is of paramount importance to make use of ETM by devising suitable event-triggering conditions, which saves system resources and ensures stable performance at the same time. One can refer to the references \cite{tabuada2007event,wang2011event,heemels2013periodic} and therein.

\subsection{Impulsive control}
In various dynamical networks \cite{zou2015restoration}, like biological networks, communication networks, and electronic networks, the states of networks often undergo abrupt changes at some instants, which may be due to switching phenomena or control requirements; and these changes can be modelled by impulsive effects. Usually, impulses can be divided into two categories: stabilizing and destabilizing impulses \cite{zhang2014synchronization,tang2015leader}, which respectively make networks stable and unstable. For dynamical networks, impulsive control is capable of adjusting the states of a network by instantaneously regulating the states of a fraction of nodes at certain instants. Due to the high efficiency of impulsive control, it has attracted increasing attention in recent years. Besides, as shown in \cite{zhang2014synchronization}, if the impulsive strength of each node is distinct in networks, we call such kind of impulses as heterogeneous impulses in space domain.

In order to clearly explain the mechanism of impulsive control, here we consider the following complex nonlinear dynamical network model:
\begin{equation}
\dot{x}_i(t)=\tilde{f}(x_i(t))+\upsilon\sum\limits_{j=1}^{N}a_{ij}x_j(t),
\label{eq11}
\end{equation}
where $i=1,2,...,N,x_i(t)=[x_{i1}(t),x_{i2}(t),...,x_{in}(t)]^{T}\in\mathbb{R}^n$ is the state vector of the \emph{i}th node at time \emph{t}; $\tilde{f}_1(x_i(t))=(\tilde{f}_{11}(x_{i1}(t)),...,\tilde{f}_{1n}(x_{in}(t))^T\in\mathbb{R}^n;\upsilon>0$ denotes the coupling strength; $A=[a_{ij}]_{N\times N}$ is the coupling matrix, where $a_{ij}$ is defined as follows: if there is a connection from node $j$ to node $i$ ($i\neq j$), then $a_{ij}=a_{ji}>0$; otherwise $a_{ij}=0$; for $i=j$, $a_{ij}$ is defined as follows:
\begin{equation}
a_{ii}=-\sum_{j=1,j\neq i}^{N} a_{ij}.
\end{equation}

Assume that the nonlinear dynamical network in (\ref{eq11}) can be forced to the following reference state $s(t)$:
\begin{equation}
\dot{s}(t)=\tilde{f}(s(t)).
\label{eq12}
\end{equation}
Let $e_i(t)=x_i(t)-s(t)$, then we get the error dynamical system:
\begin{equation}
\dot{e}_i(t)=f(e_i(t))+\upsilon\sum\limits_{j=1}^{N}a_{ij}e_j(t),
\label{eq13}
\end{equation}
where $f(e_i(t))=\tilde{f}(x_i(t))-\tilde{f}(s(t))$. Consider heterogeneous impulsive effects in system (\ref{eq11}) or (\ref{eq13}), we obtain the following model:
\begin{equation}
 \left\{
\begin{array}{ll}
\dot{e}_i(t)=f(e_i(t))+\upsilon\sum\nolimits_{j=1}^{N}a_{ij}e_j(t),t\neq t_k,k\in\mathbb{N}_+,\\
e_i(t_k^+)=e_i(t_k^-)+\mu_{ik}e_i(t_k^-),
\end{array} \right.
\label{eq14}
\end{equation}
where $\mu_{ik}$ denote impulsive strengths; the impulsive instant sequence $\{t_k\}_{k=1}^\infty$ satisfies $0<t_1<t_2<,...,<t_k<...,\textrm{lim}_{t\rightarrow\infty}t_k=\infty$; $x_i(t_k^-)$ and $x_i(t_k^+)$ denote the limit from the left and the right at time $t_k$, respectively. Without loss of generality, in this paper, we assume that $x_i(t_k^+)=x_i(t_k)$, $i=1,2,...,N$ and $t_0=0$.

\section{An Event-Triggered Impulsive Control Scheme}
In this section, we propose an event-triggered impulsive control scheme (ETI) for DE. In the following, we firstly introduce the proposed ETI in detail, which involves four components, i.e., stabilizing impulses, destabilizing impulses, ranking assignment, and an adaptive mechanism. Afterwards, we combine our approach with DE to develop ETI-DE, the pseudo-code and the computational complexity analysis of which are also given.

\subsection{Our Approach}
In our approach, ETM and impulsive control are integrated into the framework of DE algorithms. ETM identifies when the individuals should be injected with impulsive controllers, while impulsive control alters the positions of partial individuals when triggering conditions are violated. Specifically, two types of impulses, i.e., stabilizing impulses and destabilizing impulses, are imposed on the selected individuals (sorted by an index according to the fitness value and the number of consecutive stagnation generation) when the update rate (\emph{UR}) of the population in the current generation decreases or equals to zero. \emph{UR} is illustrated by Eq. (\ref{equr}).
\begin{equation}
\emph{UR}=\frac{\emph{UP}}{\emph{NP}},
\label{equr}
\end{equation}
where \emph{NP} is the population size, and \emph{UP} is the number of the individuals that update in the current generation.
On one hand, when \emph{UR} begins to decrease, stabilizing impulses drive the individuals with lower rankings in the current population to approach the individuals with better fitness values. The purpose of stabilizing impulsive control is to help update some inferior individuals and to enhance the exploitation capability of the algorithm. On the other hand, when \emph{UR} drops to zero or stabilizing impulses fail to take effect, destabilizing impulses randomly adjust the positions of the inferior individuals within the area of the current population. This operation improves the diversity of the population and hence improves the exploration ability of DE.

\begin{remark}
Given an individual $\textbf{x}_{\emph{i}}$, we can also differentiate stabilizing and destabilizing impulses from the perspective of the impulsive strength $\emph{K}_{\emph{i}}=\textrm{diag}\{\emph{K}_{\emph{i1}},\emph{K}_{\emph{i2}},...,\emph{K}_{\emph{iD}}\}$, where $\textrm{diag}\{\emph{K}_{\emph{i1}},\emph{K}_{\emph{i2}},...,\emph{K}_{\emph{iD}}\}$ denotes a diagonal matrix whose diagonal entries starting in the upper left corner are $\emph{K}_{\emph{i1}},\emph{K}_{\emph{i2}},...,\emph{K}_{\emph{iD}}$: stabilizing impulses ($\emph{K}_{\emph{ij}}\in[-1,0]$) and destabilizing impulses ($\emph{K}_{\emph{ij}}\in(0,1)$), where \emph{D} is the dimension and $\emph{j}=1,2,...,\emph{D}$.
\end{remark}

\subsubsection{Stabilizing Impulses}
Stabilizing impulses are employed when \emph{UR} begins to decrease. As mentioned before, in control theory, stabilizing impulses can be employed to regulate the states of a network to a desired value. Normally, the desired state is set as the reference state for the nodes to be injected with stabilizing impulsive controllers. In the framework of DE, stabilizing impulses mainly focus on improving the exploitation ability of DE. In DE algorithms, it is well known that good individuals (i.e., with smaller fitness values) usually contain useful information, which may be helpful to other individuals' evolution. Hence, these good individuals can be regarded as references. So when stabilizing impulsive control is triggered during the evolution, we set the individuals with smaller fitness values in the current generation as the reference states. The pseudo-code of stabilizing impulsive control is exhibited in Algorithm S.1 of the supplementary file.

Assume that $\textbf{x}_{\emph{i,G}}$ is one of the individuals at the \emph{G}th generation that are chosen to undergo impulsive effects, where $\textbf{x}_{\emph{i,G}}=[x_{\emph{i}1,\emph{G}},x_{\emph{i}2,\emph{G}},...,x_{\emph{iD},\emph{G}}]^\emph{T}$ and \emph{D} is the dimension. We set $\textbf{s}_{\emph{i,G}}$ as the reference state for $\textbf{x}_{\emph{i,G}}$, which is randomly selected from the best individual (\emph{gbest}) or other individuals with smaller fitness values than $\textbf{x}_{\emph{i,G}}$ in the current population. For each $\textbf{x}_{\emph{i,G}}$, a uniform random individual $\textbf{x}_{\emph{k,G}}$ is firstly chosen from the current population. If $f(\textbf{x}_{\emph{i,G}})<f(\textbf{x}_{\emph{k,G}})$, which means the randomly selected individual is worse than $\textbf{x}_{\emph{i,G}}$, then $\textbf{x}_{\emph{gbest,G}}$ is the reference state for $\textbf{x}_{\emph{i,G}}$; if $f(\textbf{x}_{\emph{i,G}})\geq f(\textbf{x}_{\emph{k,G}})$, which means $\textbf{x}_{\emph{k,G}}$ is better than $\textbf{x}_{\emph{i,G}}$ in the current population, then $\textbf{x}_{\emph{k,G}}$ is set as the reference state for $\textbf{x}_{\emph{i,G}}$.

The error between $\textbf{x}_{\emph{i,G}}$ and its reference state $\textbf{s}_{\emph{i,G}}$ at the \emph{G}th generation can be obtained:
\begin{equation}
\textbf{e}_{\emph{i,G}}=\textbf{x}_{\emph{i,G}}-\textbf{s}_{\emph{i,G}}.
\label{eq15}
\end{equation}
Then at the end of the \emph{G}th generation, stabilizing impulses force the chosen individuals to approach their reference state. Here we get:
\begin{equation}
\textbf{x}_{\emph{i,G}^+}=\textbf{x}_{\emph{i,G}}+\emph{K}_{\emph{i,G}}\cdot\textbf{e}_{\emph{i,G}},
\label{eq16}
\end{equation}
where $\emph{K}_{\emph{i,G}}=\textrm{diag}\{\emph{K}_{\emph{i1,G}},\emph{K}_{\emph{i2,G}},...,\emph{K}_{\emph{iD,G}}\}$ is the impulsive strength for individual $\textbf{x}_\emph{i}$ at the \emph{G}th generation. $\emph{K}_{\emph{ij,G}}\in(-1,0)$ shows that in the \emph{j}th dimension, $x_{\emph{i,G}}$ lies on the line between the reference and the individual itself; $\emph{K}_{\emph{ij,G}}=0$ means that in the \emph{j}th dimension, $x_{\emph{i,G}}$ is not injected with impulsive controllers; $\emph{K}_{\emph{ij,G}}=-1$ indicates that in the \emph{j}th dimension, $x_{\emph{i,G}}$ reaches the reference state, $\emph{j}=1,2,...,\emph{D}$. $\emph{G}^+$ denotes that stabilizing impulses are imposed on $\textbf{x}_{\emph{i,G}}$ at the end of the \emph{G}th generation. Every time, \emph{DM} dimensions of $\textbf{x}_{\emph{i,G}}$ are selected in a uniformly random way to be injected with impulsive controllers, where $\emph{DM}\in\{1,2,...,\emph{D}\}$.
When $\textbf{x}_{\emph{gbest,G}}$ serves as the reference state, for the selected \emph{DM} dimensions, the impulsive strength $\hat{\emph{K}}_{\emph{i,G}}=\textrm{diag}\{\emph{K}_{\emph{i}1,\emph{G}},\emph{K}_{\emph{i}2,\emph{G}},...,\emph{K}_{\emph{iDM},\emph{G}}\}_{\emph{DM}\times \emph{DM}}$, and $\emph{K}_{\emph{ij,G}}$ is a uniform random number from $-1$ to 0, $\emph{j}=1,2,...,\emph{D}$; for the rest $\emph{D}-\emph{DM}$ dimensions, $\check{\emph{K}}_{\emph{i,G}}=\textrm{diag}\{0,0,...,0\}_{(\emph{D}-\emph{DM})\times (\emph{D}-\emph{DM})}$.
When $\textbf{x}_{\emph{k,G}}$ is as the reference state, for the selected \emph{DM} dimensions, the impulsive strength $\hat{\emph{K}}_{\emph{i,G}}=\textrm{diag}\{-1,-1,...,-1\}_{\emph{DM}\times \emph{DM}}$; for the rest $\emph{D}-\emph{DM}$ dimensions, $\check{\emph{K}}_{\emph{i,G}}=\textrm{diag}\{0,0,...,0\}_{(\emph{D}-\emph{DM})\times (\emph{D}-\emph{DM})}$. $\emph{K}_{\emph{i,G}}$ is obtained from combining $\hat{\emph{K}}_{\emph{i,G}}$ and $\check{\emph{K}}_{\emph{i,G}}$.
It is noticed that $\zeta$ is a flag to indicate whether stabilizing impulsive control is successful to improve the performance: when $\zeta=1$, it means that the stabilizing impulsive control takes effect, and a new individual is introduced to the population by replacing an old one; when $\zeta=0$, it shows that the stabilizing impulsive control fails to take effect.

\begin{remark}
According to \cite{zhang2013exponential}, if the impulsive strength of each node is distinct in networks, such kind of impulses is called heterogeneous impulses in space domain. For stabilizing impulses developed in this paper, the impulsive strengths are not only heterogeneous in each individual of the population but also nonidentical in each dimension of each individual. Hence, it is apparent that our proposed impulses generalize the heterogeneous impulses in \cite{zhang2013exponential}. Apart from enhancing the performance of DE algorithms, our proposed stabilizing impulses can also contribute to the design of impulsive control systems.
\end{remark}

\begin{remark}
In \cite{tang2015leader}, if impulses are injected into only a fraction of nodes, such kind of impulses is called partial mixed impulses. In this paper, stabilizing impulses are imposed on not only a group of individuals in the population, but also partial dimensions of each individual. Therefore, our presented stabilizing impulses can be regarded as a hierarchical partial mixed impulses when compared with the impulses in \cite{tang2015leader}. Besides, the proposed impulses will not only promote the development of new powerful DEs, but also shed light on the design of impulsive control systems.
\end{remark}

\subsubsection{Destabilizing Impulses}
When \emph{UR} drops to zero ($\emph{UR}=0$) or stabilizing impulses fail to take effect ($\zeta=0$), destabilizing impulses are introduced to provide some randomness during the evolution. When destabilizing impulses are triggered, the selected individuals can be moved to any position within the range of the current population. The pseudo-code of injecting destabilizing impulses is exhibited in Algorithm S.2 of the supplementary file. Assume that $\textbf{x}_{i,G}$ is one of the individuals at the $G$th generation that are chosen to receive destabilizing impulses, where $\textbf{x}_{\emph{i,G}}=[x_{\emph{i}1,\emph{G}},x_{\emph{i}2,\emph{G}},...,x_{\emph{iD},\emph{G}}]^\emph{T}$ . $\emph{min}_{\emph{j,G}}$ and $\emph{max}_{\emph{j,G}}$ are the minimum and maximum values of the \emph{j}th dimension in the population at the \emph{G}th generation, $\emph{j}=1,2,...,\emph{D}$. The lower and upper bounds of the range of the population at the \emph{G}th generation are:
\begin{equation}
\textbf{x}_{\emph{L,G}}=[\emph{min}_{\emph{1,G}},\emph{min}_{\emph{2,G}},...,\emph{min}_{\emph{D,G}}]^{T},
\label{eq17}
\end{equation}
\begin{equation}
\textbf{x}_{\emph{U,G}}=[\emph{max}_{\emph{1,G}},\emph{max}_{\emph{2,G}},...,\emph{max}_{\emph{D,G}}]^{T}.
\label{eq18}
\end{equation}
From Eqs. (\ref{eq17})-(\ref{eq18}), we can obtain the error between $\textbf{x}_{\emph{U,G}}$ and $\textbf{x}_{\emph{L,G}}$ at the \emph{G}th generation:
\begin{equation}
\textbf{e}_{\emph{i,G}}=\textbf{x}_{\emph{U,G}}-\textbf{x}_{\emph{L,G}}.
\label{eq19}
\end{equation}
Then at the end of the \emph{G}th generation, the positions of the chosen individuals are randomly updated in the specified range. Here we have:
\begin{equation}
\textbf{x}_{\emph{i,G}^+}=\textbf{x}_{\emph{L,G}}+\emph{K}_{\emph{i,G}}\cdot\textbf{e}_{\emph{i,G}},
\label{eq20}
\end{equation}
where $\emph{K}_{\emph{i,G}}=\textrm{diag}\{\emph{K}_{\emph{i1,G}},\emph{K}_{\emph{i2,G}},...,\emph{K}_{\emph{iD,G}}\}$ is the impulsive strength for individual $\textbf{x}_\emph{i}$ at the \emph{G}th generation. Similarly, \emph{DM} dimensions of $\textbf{x}_{\emph{i,G}}$ are selected at random to be injected with impulses. $\emph{K}_{\emph{ij,G}}$ is a uniform random number from 0 to 1, $\emph{j}=1,2,...,\emph{D}$.

\begin{remark}
Similar to our developed stabilizing impulses, destabilizing impulses are imposed on a part of individuals. Thus our proposed destabilizing impulses can be regarded as partial mixed impulses according to \cite{tang2015leader}. Besides, random dimensions of the individual are chosen to be injected with destabilizing impulses, the impulsive strengths of which range from $(0,1)$. Therefore, destabilizing impulses in this paper can also be understood as generalized heterogeneous impulses when compared with the impulses in \cite{zhang2013exponential}.
\end{remark}

\begin{remark}
Two kinds of impulses are proposed in the framework of DE. The idea of introducing impulsive control into DE comes from the fact that impulsive control takes effect in dynamical networks. In \cite{zhang2014synchronization}, stabilizing impulses are imposed on partial nodes of a network, and the desired state is set as the reference state for these nodes. The dynamical network with stabilizing impulses can be synchronized to a desired state.
In DE, stabilizing impulses act as impetus, which forces certain individuals to approach good individuals (references) in the population at certain instants. And this operation is expected to facilitate the fast convergence of the population.
In addition, destabilizing impulses introduce disturbances to a network in multi-agent systems or dynamical networks \cite{tang2015leader}. Similarly, in DE, a fraction of individuals are injected with destabilizing impulses at certain moments, which aims at bring some randomness into the evolution process. In dynamical networks, impulsive control is used to adjust the states of a network by instantly regulating the states of a fraction of nodes at certain instants. And in DE, impulsive control is expected to enhance the search performance of the whole population by instantaneously modifying the positions of a part of individuals at certain moments.
\end{remark}

\subsubsection{Ranking Assignment}
In the following, we need to consider which individuals should be injected with impulsive controllers. During the evolution process, we consider two measures to characterize the status of the individuals. The first one is the fitness value of each individual, while the second one is the number of each individual's consecutive stagnant generation. Fitness value is the most direct index to judge whether an individual should enter into the next generation or not. The number of consecutive stagnant generation reflects the degree of the activity of an individual in the evolution. If an individual does not evolve for a relatively long time, it might be necessary to introduce some additional operations to change its position. Based on these discussions, in this paper, we rank the population based on the fitness value and the number of consecutive stagnant generation, respectively. $\tilde{\emph{R}}_\emph{i}$ is the ranking of $\textbf{x}_{\emph{i,G}}$ according to the fitness value, and $\bar{\emph{R}}_\emph{i}$ is the ranking of $\textbf{x}_{\emph{i,G}}$ based on the number of consecutive stagnation generation.
These two rankings are both ordered in an ascending way (i.e., from the best fitness value to the worst and from the smallest number of consecutive stagnation generation to the largest). Then we combine $\tilde{\emph{R}}_\emph{i}$ and $\bar{\emph{R}}_\emph{i}$ to get $\emph{R}_\emph{i}$, which indicates that the individuals are sorted according to both the fitness value and the number of consecutive stagnation generation.
\begin{equation}
\emph{R}_\emph{i}=\tilde{\emph{R}}_\emph{i}+\bar{\emph{R}}_\emph{i}.
\label{eqra}
\end{equation}
$\emph{R}_\emph{i}$ not only reflects the fitness value of the individual $\textbf{x}_{\emph{i,G}}$ but also delivers the degree of the individual's activity.

When impulsive control is triggered during the evolution, we select the individuals with larger values of $\emph{R}_\emph{i}$ from the population as the candidates to undergo stabilizing or destabilizing impulses. By specially displacing the individuals with higher rankings (i.e., larger $\emph{R}_\emph{i}$), the evolution status of the population can be improved.

\subsubsection{An adaptive mechanism to determine the number of individuals taking impulsive control}
Finally, in order to further improve the performance of ETI, an adaptive mechanism is proposed to determine the number of the individuals that should be injected with impulsive controllers. We firstly discuss the number of individuals (\emph{M}) with stabilizing impulses. \emph{LN} and \emph{UN} represent the lower and upper bound of \emph{M}, respectively. When stabilizing impulsive control is triggered for the first time, $\emph{M}=\emph{LN}$. After $\textbf{x}_{\emph{i,G}}$ experiences the stabilizing impulse, we get $\textbf{x}_{\emph{i,G}^+}$. $\textbf{x}_{\emph{i,G}^+}$ can join the current population instead of $\textbf{x}_{\emph{i,G}}$ if and only if $f(\textbf{x}_{\emph{i,G}^+})<f(\textbf{x}_{\emph{i,G}})$. Every time $\textbf{x}_{\emph{i,G}}$ is replaced with $\textbf{x}_{\emph{i,G}^+}$ (i.e., $\zeta=1$, see step 36 in Algorithm S.1 of the supplementary file), \emph{M} keeps unchanged. If $\zeta=0$, $\emph{M}=\emph{M}+1$. We aim to increase the success rate of stabilizing impulsive control by having more individuals to be injected with stabilizing impulsive controllers. Besides, if a new \emph{gbest} is generated in the population, \emph{M} drops to a random integer between $[\emph{LN},\emph{M}]$. The reason for reducing \emph{M} to a random integer between $[\emph{LN},\emph{M}]$ instead of \emph{LN} is to increase the times of successful stabilizing impulsive control, especially in the later stage of the evolution.
Next, we explain how to choose the number of individuals that undergo destabilizing impulses. As introduced above, destabilizing impulses are added in two cases: when $\emph{UR}=0$ or $\zeta=0$. Unlike stabilizing impulses, with the purpose of introducing some randomness, the selection operation (i.e., compare the fitness values of $\textbf{x}_{\emph{i,G}}$ and $\textbf{x}_{\emph{i,G}^+}$) will not be used after injecting destabilizing impulses, which means that $\textbf{x}_{\emph{i,G}^+}$ replaces $\textbf{x}_{\emph{i,G}}$ directly. Therefore, in order not to bring too many individuals with large fitness values into the population, we randomly select the individuals from \emph{M} candidates to be injected with destabilizing impulses. The selection process is described in Algorithm S.3 of the supplementary file.

\begin{remark}
In this paper, stabilizing and destabilizing impulses are triggered separately based on the status of the individuals. Two measures are used to characterize the status of the individuals, and one of them is the fitness value of each individual. In recent years, fitness control adaptation works effectively for developing evolutionary algorithms, which performs corrections and anti-corrections \cite{tirronen2009differential,caponio2007fast,neri2007adaptive,caponio2009super,tirronen2008enhanced,neri2011disturbed}. In the literature above, the best, worst, and average fitness values in the population are utilized to construct some metrics, such as $\xi$ in \cite{caponio2007fast}, $\psi$ in \cite{neri2007adaptive}, $\chi$ in \cite{caponio2009super}, and so on. These parameters adaptively determine the activation time of each local searcher. Although our method is similar to the fitness control adaptation, there are some differences between the works in \cite{tirronen2009differential,caponio2007fast,neri2007adaptive,caponio2009super,tirronen2008enhanced,neri2011disturbed} and in our research. Firstly, in our research, we rank the fitness values of the whole population, instead of using some typical values (i.e., the best, worst, and average values). Secondly, apart from the ranking of fitness value, we also consider the ranking according to the number of consecutive stagnation generation of the whole population. These two rankings are combined into one measure \emph{R} in (22). Thirdly, the metrics $\xi$, $\psi$, and $\chi$ in \cite{tirronen2009differential,caponio2007fast,neri2007adaptive,caponio2009super,tirronen2008enhanced,neri2011disturbed} are used to activate different local searchers. While in our work, \emph{R} is the measure to select individuals as the candidate to undergo stabilizing and destabilizing impulses. Therefore, fitness values play different roles in literature \cite{tirronen2009differential,caponio2007fast,neri2007adaptive,caponio2009super,tirronen2008enhanced,neri2011disturbed} and in our work. And it can be generally recognized that our work also fits into the framework of fitness control adaptation.
\end{remark}

\begin{remark}
It is worth mentioning that the essential of ETI is to adjust the search strategies of individuals according to the evolutionary states. In recent studies on DE and particle swarm optimization (PSO) \cite{yu2014differential,zhan2009adaptive,wu2014superior,yang2015differential}, the algorithms proposed also select the search strategies and parameters based on the states of the individuals. However, our work is quite different from these studies in the following three aspects: 1) The measures are different when representing the states of the individuals. For example, in \cite{yu2014differential}, the measure uses fitness and distance information; in \cite{zhan2009adaptive}, distance information is utilized; in \cite{wu2014superior}, the measure takes advantage of fitness and position information; in \cite{yang2015differential}, position information is considered; while we use fitness and stagnation information in our paper. 2) The search strategies are different in the algorithms developed in \cite{yu2014differential,zhan2009adaptive,wu2014superior,yang2015differential} and our proposed ETI. For instance, \cite{yu2014differential} and \cite{zhan2009adaptive} developed parameter adaptation strategies; \cite{wu2014superior} used a mutation operator in PSO; \cite{yang2015differential} employed a restart strategy after stagnation; while our research makes certain individuals approach superior individuals or reinitialized. 3) Our ETI is presented within a DE framework and cannot be incorporated into PSO, which will be explained in Remark \ref{remarkpso}.
\end{remark}

\subsection{DE with An Event-Triggered Impulsive Control Scheme}
Combining the developed event-triggered impulsive control scheme (ETI) with DE, the ETI-DE is proposed. The pseudo-code of ETI-DE with ``DE/rand/1" mutation operator is given in Algorithm 1. From step 7 to step 28, it is the original DE algorithm with ``DE/rand/1" mutation operator. The rest steps in Algorithm 1 illustrate the mechanism of ETI. ETM determines the moment to add impulses to the individuals, and impulsive control modifies the positions of partial individuals at the end of a certain generation. In detail, step 35 to step 42 and step 49 to step 56 describe the mechanism of destabilizing impulses, which are triggered when $\emph{UR}=0$ or $\zeta=0$. While step 43 to step 48 shows the details of stabilizing impulsive control, which is triggered when \emph{UR} decreases and $\emph{UR}\neq0$. These two types of impulses are able to accelerate the convergence of the population by updating some inferior individuals, and improve the diversity of the population by introducing some randomness to the search. Furthermore, ETI is flexible to be integrated into other advanced DE variants, such as jDE \cite{brest2006self}, JADE \cite{zhang2009jade}, SaDE \cite{qin2009differential}, and so on.

\begin{algorithm}[!htbp]\scriptsize{
\caption{DE with event-triggered impulsive control scheme} \algblock{Begin}{End}
\begin{algorithmic}[1]
\Begin
\State /* \emph{UR} is the update rate of the population in each generation
\State /* $\emph{UR\_tp}$ stores the temporary value of \emph{UR}
\State /* \emph{gbest} is the best individual of the population in the current generation
\State /* $\emph{gbest\_tp}$ stores the temporary value of \emph{gbest}
\State /* \emph{rs} records the number of individuals to be injected with destabilizing impulses
\State Set $\emph{LN}=1$; $\emph{UN}=\emph{NP}$; $\emph{M}=\emph{LN}$; $\emph{UR}=0$; $\emph{F}=0.5$; $\emph{CR}=0.9$
\State Create a random initial population \{$\textbf{x}_{\emph{i},0}|\emph{i}=1,2,...,\emph{NP}$\}
\State Evaluate the fitness values of the population and record \emph{gbest}
\While {the maximum evaluation number is not achieved}
\State $\emph{UR\_tp}=\emph{UR}$; $\emph{gbest\_tp}=\emph{gbest}$
 \For {\emph{i} = 1 to \emph{NP}}
 \State Select randomly three individuals $r_1 \neq r_2 \neq r_3 \neq i$
 \State $\textbf{v}_{\emph{i,G}} = \textbf{x}_{r_1,\emph{G}}+F\cdot(\textbf{x}_{r_2,\emph{G}}-\textbf{x}_{r_3,\emph{G}})$
 \State Check the boundary of $\textbf{v}_{\emph{i},\emph{G}}$
 \State Generate $\emph{j}_{\textrm{rand}}=\textrm{randi}(\emph{D},1)$
 \For {\emph{j} = 1 to \emph{D}}
   \If {$j=j_{\textrm{rand}}$ \textbf{or} $\textrm{rand}<\emph{CR}$}
   \State $u_{\emph{ij,G}}=v_{\emph{ij,G}}$
   \Else
   \State $u_{\emph{ij,G}}=x_{\emph{ij,G}}$
   \EndIf
 \EndFor
 \State Evaluate the fitness value of $\textbf{u}_{\emph{i,G}}$
 \If {$f(\textbf{u}_{\emph{i,G}}) \leq f(\textbf{x}_{\emph{i,G}})$}
 \State $\textbf{x}_{\emph{i,G+1}} = \textbf{u}_{\emph{i,G}}$
 \EndIf
 \EndFor
\State Record the fitness value of the best individual as \emph{gbest}
\If {$\emph{gbest}<\emph{gbest\_tp}$}
\State $\emph{M}=\textrm{randi}([\emph{LN},\emph{M}],1)$
\EndIf
\State Calculate $\tilde{\emph{R}}_\emph{i}$ and $\bar{\emph{R}}_\emph{i}$ of the population, $\emph{R}_\emph{i}=\tilde{\emph{R}}_\emph{i}+\bar{\emph{R}}_\emph{i}$
\State Calculate \emph{UR} of the population
\If {$\emph{UR}=0$}
\State $\emph{M}=\textrm{min}(\emph{M},\emph{UN})$
\State Select \emph{M} individuals with the largest $\emph{R}_\emph{i}$-value as \{$\textbf{x}_{\emph{i,G}}|\emph{i}=1,2,...,\emph{M}$\}
\State \{$\textbf{x}_{\emph{i,G}}|\emph{i}=1,2,...,\emph{rs}$\} = \textbf{Random\_Selection\_of\_Individuals ()}
\For {$\emph{i}=1$ to \emph{rs}}
\State $\textbf{x}_{\emph{i,G}}$ = \textbf{Injecting\_Destabilizing\_Impulsive ()}
\State Evaluate the fitness value of $\textbf{x}_{\emph{i,G}}$
\EndFor
\ElsIf {$\emph{UR}\neq0$ \textbf{and} $\emph{UR}<\emph{UR\_tp}$}
\State $\emph{M}=\textrm{min}(\emph{M},\emph{UN})$
\State Select \emph{M} individuals with largest $\emph{R}_\emph{i}$-value as \{$\textbf{x}_{\emph{i,G}}|\emph{i}=1,2,...,\emph{M}$\}
\For {$\emph{i}=1$ to \emph{M}}
\State [$\textbf{x}_{\emph{i,G}}$, $\zeta_{\emph{i,G}}$] = \textbf{Stabilizing\_Impulsive\_Control ()}
\EndFor
\If {$\textrm{sum}(\zeta_{\emph{1,G}},\zeta_{\emph{2,G}},...,\zeta_{\emph{M,G}})=0$}
\State \{$\textbf{x}_{\emph{i,G}}|\emph{i}=1,2,...,\emph{rs}$\} = \textbf{Random\_Selection\_of\_Individuals ()}
\For {$\emph{i}=1$ to \emph{rs}}
\State $\textbf{x}_{\emph{i,G}}$ = \textbf{Injecting\_Destabilizing\_Impulsive ()}
\State Evaluate the fitness value of $\textbf{x}_{\emph{i,G}}$
\State $\emph{M}=\emph{M}+1$
\EndFor
\EndIf
\State Record the best individual of current population as $\emph{gbest\_tp}$
\If {$\emph{gbest\_tp}<\emph{gbest}$}
\State $\emph{M}=\textrm{randi}([\emph{LN},\emph{M}],1)$
\EndIf
\EndIf
\EndWhile
\End
\end{algorithmic}}
\end{algorithm}

In ETI-DEs, stabilizing impulses force the individual $\textbf{x}_{\emph{i,G}}$ to approach its reference state $\textbf{s}_{\emph{i,G}}$, and destabilizing impulses randomly adjust the positions of the individuals within the area of the current population. These two operations are carried out within the search range of the problem, so they will not generate invalid solutions during the evolution process. In the following experimental section, when ETI is incorporated into other DE variants, we will not change the original bound constraints handling methods of these DE variants. These methods define that at any stage of the search process, solutions outside the bounds are invalid, just like the situation \textbf{S1} in \cite{liao2014note}. So for the CEC 2014 benchmark functions used in Section IV, ETI-DEs will not search outside the region $[-100, 100]^{D}$.

\begin{remark}
The proposed ETI is presented within a DE framework. Meanwhile, because different evolutionary algorithms (EAs) have different structures, ETI cannot be directly incorporated into other EAs, and some related modifications are needed on ETI. For example, in genetic algorithm (GA), only a fraction of individuals are selected as parents at each generation, so the number of each individual's consecutive stagnant generation is meaningless. Therefore, it is necessary to propose another measure to denote the state of the individuals. In particle swarm optimization (PSO), each member of the swarm searches the space based on the historical information of itself (\emph{pbest}) and other members (\emph{gbest}). So ETI may use the stagnation information of \emph{pbest} and \emph{gbest} to be fitted into PSO. In our future work, we will investigate in detail whether ETI can work efficiently in other EAs. According to the explanations in \cite{neri2010recent}, in essence, ETI varies the moves and enriches the pool of search moves. In detail, stabilizing impulses introduce extra moves towards individuals with better fitness values, the goal of which is to increase the exploitative pressure. Destabilizing impulses bring in more explorative moves, which helps the population explore the search space.\label{remarkpso}
\end{remark}

Here, we discuss the complexity of ETI. Generally, the proposed ETI-DE does not significantly increase the overall computational complexity of the original DE algorithm. The additional complexity of ETI-DE is population sorting when calculating $\tilde{\emph{R}}$ and $\bar{\emph{R}}$, and implementing impulsive control. The complexity of population sorting is $\mathcal{O}(2\cdot\emph{NP}\cdot\textrm{log}(\emph{NP}))$, and the maximum complexity of impulsive control is $\mathcal{O}(\emph{NP}\cdot D)$. It is known that the complexity of the original DE is $\mathcal{O}(\emph{G}_{\emph{max}}\cdot \emph{NP}\cdot \emph{D})$, so the total complexity of ETI-DE is $\mathcal{O}(2\cdot\emph{G}_{\emph{max}}\cdot \emph{NP}\cdot(\emph{D}+\textrm{log}(\emph{NP}))$, which can be regarded as the same as the original DE. Therefore, our presented scheme does not seriously increase the computational cost of the original DE.

\section{Experimental Results and Analysis}
In this section, we carry out extensive experiments to evaluate the performance of our developed ETI-DE. The total 30 benchmark functions presented in the CEC 2014 competition on single objective real-parameter numerical optimization are selected as the test suite \cite{liang2013problem}. According to their characteristics, the functions can be divided into four groups: 1) unimodal functions (F01-F03); 2) simple multimodal functions (F04-F16); 3) hybrid functions (F17-F22); 4) composition functions (F23-F30). More details of these functions can be found in \cite{liang2013problem}.

\subsection{Parameter Settings}
In the following experiments, we incorporate the proposed event-triggered impulsive control scheme with two original DE algorithms and eight state-of-the-art DE variants. The parameters are set as follows:

1) DE/rand/1/bin with $\emph{F}=0.5$, $\emph{CR}=0.9$ \cite{cai2013differential};

2) DE/best/1/bin with $\emph{F}=0.7$, $\emph{CR}=0.5$ \cite{cai2013differential};

3) jDE with $\tau_1=0.1$, $\tau_2=0.1$ \cite{brest2006self};

4) JADE with $\mu_\emph{F}=0.5$, $\mu_\emph{CR}=0.5$, $\emph{c}=0.1$, $\emph{p}=0.05$ \cite{zhang2009jade};

5) CoDE with $\emph{F}=[1.0, 1.0, 0.8]$, $\emph{CR}=[0.1, 0.9, 0.2]$ \cite{wang2011differential};

6) SaDE with $\emph{LP}=50$ \cite{qin2009differential};

7) ODE with $\emph{F}=0.5$, $\emph{CR}=0.9$, $\emph{J}_\emph{r}=0.3$ \cite{rahnamayan2008opposition};

8) EPSDE with $\emph{F}=[0.1, 0.2, 0.3, 0.4, 0.5, 0.6, 0.7, 0.8, 0.9]$, $\emph{CR}=[0.4, 0.5, 0.6, 0.7, 0.8, 0.9]$ \cite{mallipeddi2011differential};

9) SHADE with initial $\emph{M}_{F}=0.5$, $\emph{M}_{CR}=0.5$, $\emph{H}=\emph{NP}$ \cite{tanabe2013success};

10) OXDE with $\emph{F}=0.5$, $\emph{CR}=0.9$ \cite{wang2012enhancing}.

For the incorporated ETI-DE algorithms, the lower and upper bounds of the number of individuals that take impulsive control are set: $\emph{LN}=1$, $\emph{UN}=\emph{NP}$. The maximum number of function evaluations (MAX\_FES) is set to $\emph{D}\cdot10000$. We run each function optimized by each algorithm 51 times for the experiments \cite{liang2013problem}. The simulations are performed on an Intel Core i7 personal computer with 2.10-GHz central processing unit and 8-GB random access memory.

\begin{remark}
In Algorithm S.1, \emph{DM} is the number of dimensions of an individual selected to undergo stabilizing impulsive control. Every time, \emph{DM} dimensions of an individual are chosen in a uniformly random way to be injected with impulsive controllers, where $\emph{DM}\in\{1,2,...,\emph{D}\}$. In this section, the sensitivity of \emph{DM} is studied beforehand by comparing the performance of ETI-DEs with random \emph{DM} and with $\emph{DM}=[\emph{D}/3,2\emph{D}/3,\emph{D}]$. We provide the experimental results in Table S.11 of the supplementary file. The results show that in 26 out of 30 cases, ETI-DEs with random \emph{DM} perform better than those with $\emph{DM}=[\emph{D}/3,2\emph{D}/3,\emph{D}]$. Therefore, it is reasonable to select random \emph{DM} dimensions of an individual to take stabilizing impulsive control, which introduces some randomness into the evolution.
\end{remark}

It is worth mentioning that the above ten algorithms were tested on various benchmark problems, which are different from the CEC 2014 test suite in our paper. To make the comparisons fair and meaningful, an appropriate tuning of the population size must be carried out. Therefore, a set of tests are conducted to select a proper population size for each algorithm. In detail, the ten DE algorithms with $\emph{NP}=[30, 50, 100, 150]$ are applied to optimizing the CEC 2014 test suite 51 times, respectively, and the Holm-Bonferroni procedure \cite{holm1979simple} with confidence level 0.05 is used to evaluate the performance of each algorithm with different \emph{NP} values. The results are listed in Tables S.1-S.10 of the supplementary file. According to the obtained results, for each algorithm, the \emph{NP} value with the highest rank (highlighted in \textbf{boldface}) is chosen as its population size in the following experiments of our research. That is, DE/rand/1/bin: $\emph{NP}=100$, DE/best/1/bin: $\emph{NP}=50$, jDE: $\emph{NP}=100$, JADE: $\emph{NP}=100$, CoDE: $\emph{NP}=50$, SaDE: $\emph{NP}=100$, ODE, $\emph{NP}=100$, EPSDE: $\emph{NP}=50$, SHADE: $\emph{NP}=150$, OXDE: $\emph{NP}=100$.

In our experiment studies, three performance evaluation criteria are used for comparing the performance of each algorithm, which are listed below:
\subsubsection{Error}
The average and standard deviation of the function error value $f(\textbf{x})-f(\textbf{x}^{*})$ are recorded, where $\textbf{x}^{*}$ is the global optimum of the test function and $\textbf{x}$ is the best solution found by the algorithm in a single run. And error value smaller than $10^{-8}$ will be taken as $0$ \cite{liang2013problem}.
\subsubsection{Convergence graphs}
The convergence graphs are plotted to illustrate the mean function error values derived from each algorithm in the comparison.
\subsubsection{Wilcoxon rank-sum test}
In order to show the significant difference between the original DE and its ETI-DE variant, a Wilcoxon rank-sum test at 5\% significance level is conducted. The cases are marked with ``$+$/$\approx$/$-$" when the performance of the ETI-DE variant is significantly better than, equal to, and worse than the DE algorithm without the proposed scheme, respectively.
\subsubsection{Holm-Bonferroni procedure}
In order to complete the statistical analysis, the Holm-Bonferroni procedure with confidence level 0.05 is performed.

\subsection{Comparison with Ten DE Algorithms}
In this section, we assess the effectiveness of our developed scheme by comparing ten popular DE algorithms and their corresponding ETI-based variants. The experimental results are provided in Tables S.12-S.14 of the supplementary file. ``$+$/$\approx$/$-$" indicates that the performance of DE algorithms with ETI is significantly better than, equal to, and worse than those without ETI. The better values compared between the DE variants and their corresponding ETI-based DEs are highlighted in \textbf{boldface}.

From Tables S.12-S.14, we can see that the ten ETI-DEs perform better than their corresponding original DE algorithms. For example, for all the 30 test functions, ETI-DE/rand/1/bin improves in 16 functions, ties in 9 functions, and only loses in 5 functions; when compared with DE/best/1/bin, ETI-DE/best/1/bin wins in 15 cases, ties in 10 cases, and merely loses in 5 cases; for jDE, the incorporation of the proposed scheme exhibits superior performance in 13 functions, and provides similar performance in 15 functions; for ETI-JADE, it outperforms JADE in 20 out of 30 functions, and ties in 9 functions; for ETI-CoDE, it obtains better results in 10 functions, while ties in 17 functions, and just loses in 3 functions; for ETI-SaDE, it wins, ties, and loses in 10, 16, and 4 cases, respectively; for ODE, the proposed scheme improves its performance in 15 functions and only gets worse in 1 function; for ETI-EPSDE, it improves in 21 function, ties in 6 functions, and simply loses in 3 functions; ETI-SHADE wins in 11 cases, ties in 15 cases, and only loses in 4 cases when compared with SHADE; for OXDE, ETI enhances its performance in 17 functions, and merely becomes worse in 3 functions. In general, ETI significantly improves the search ability of the ten popular DE variants. Furthermore, in Figs. S.1-S.4 of the supplementary file, we use the box plot to show the results of JADE, CoDE, SaDE and EPSDE with and without ETI on CEC 2014 test suite at $\emph{D}=30$. Combining Tables S.12-S.14 with Figs. S.1-S.4, we can see the effectiveness of our proposed ETI.

The results of the Holm-Bonferroni procedure are given in Table S.15 of the supplementary file, where we set ETI-SHADE as the reference algorithm. From the rank values in Table S.15, we can find that ETI improves the performance of all the ten DE variants on the CEC 2014 test suite at D=30.

To better illustrate the convergence performance of the ten DE algorithms and their corresponding ETI-DEs, we plot the convergence curves of these algorithms in Fig. S.5 of the supplementary file for six selected test functions, which are from the four groups of the test suite. From Fig. S.5, we can observe that our proposed ETI improves the convergence performance of the ten original DE algorithms by introducing two types of impulses.

In summary, the presented ETI is very powerful and the ten ETI-DEs possess strong capabilities of rapid convergence and accurate search for the test functions. The results of the Wilcoxon rank-sum test confirm that our scheme is of paramount importance to improve the performance of the considered DE algorithms.

\subsection{Effectiveness of Two Types of Impulses}
In light of the results shown in Tables S.12-S.14, it can be seen that the proposed ETI can significantly improve the performance of the ten DE algorithms. The core of ETI is the use of two types of impulses: stabilizing impulses and destabilizing impulses. Stabilizing impulses help a part of individuals get close to promising areas, which enhance the exploitation ability of the algorithm; while destabilizing impulses increase the diversity of the current population, which improves the exploration capability of the algorithm. In this section, we conduct four groups of experiments to examine how these two kinds of impulses separately take effect for the DE algorithms. Therefore, we consider the following four variants of ETI-DE:

1) ETI1-DE: the proposed scheme only with injecting destabilizing impulses when $\emph{UR}=0$ in each improved DE (steps 43-60 in Algorithm 1 are deleted);

2) ETI2-DE: the proposed scheme without injecting destabilizing impulses when $\emph{UR}=0$ in each improved DE (steps 35-42 in Algorithm 1 are deleted);

3) ETI3-DE: the proposed scheme without injecting destabilizing impulses when $\zeta=0$ in each improved DE (steps 49-56 in Algorithm 1 are deleted);

4) ETI4-DE: the proposed scheme without injecting any destabilizing impulses both when $\emph{UR}=0$ and $\zeta=0$ in each improved DE (steps 35-42 and 49-56 in Algorithm 1 are deleted).

Firstly, ETI1-DEs are compared with the ten ETI-DEs to show the effectiveness of stabilizing impulsive control. Secondly, ETI2-DEs are compared with the ten ETI-DEs to inspect the effectiveness of destabilizing impulses when $\emph{UR}=0$. Thirdly, ETI3-DEs are compared with the ten ETI-DEs to examine the effectiveness of destabilizing impulses when $\zeta=0$. Fourthly, ETI4-DEs are compared with the ten ETI-DEs to exhibit the effectiveness of destabilizing impulses, which are triggered in two cases: when $\emph{UR}=0$ and when $\zeta=0$. For saving space, we only list the \emph{win}-\emph{lose} results of four types of comparisons (ETI-DEs \emph{vs.} ETI1-DEs, ETI-DEs \emph{vs.} ETI2-DEs, ETI-DEs \emph{vs.} ETI3-DEs, ETI-DEs \emph{vs.} ETI4-DEs) according to the Wilcoxon rank-sum test in Table S.16 of the supplementary file.

Based on the \emph{win}-\emph{lose} results in Table S.16, the following conclusions can be drawn:

1) The first two rows indicate the positive contribution of stabilizing impulses. Stabilizing impulses rearrange the location distribution of the population by making the individuals with higher rankings (i.e., larger \emph{R}) reach the areas close to the individuals with better fitness values, which increases the search efficiency of the ten DEs.

2) The rest six rows confirm the effectiveness of destabilizing impulses. In detail, the third and fourth rows demonstrate that destabilizing impulses triggered by the condition $\emph{UR}=0$ is of great importance to almost all the ten algorithms. $\emph{UR}=0$ means that the whole population stops updating at the current generation, which is quite unfavorable for the evolution. Destabilizing impulses force the inferior individuals to leave their previous positions, in order to pull the whole population out of the impasse. The fifth and sixth rows display the effect of injecting destabilizing impulses when the stabilizing impulses fail to take effect. Specifically, ETI-DEs wins in 4 cases, ties in 3 cases, and loses in 3 cases. The last two rows show that without destabilizing impulses, the performance of the ten algorithms deteriorates. In control theory, destabilizing impulses introduce disturbances to a dynamical system. Similarly, in DE, destabilizing impulses bring some randomness into the evolution process when the population reaches an impasse.

\subsection{Effectiveness of Random Selection of the Reference State in Stabilizing Impulses}
In ETI, when $\textbf{x}_{\emph{i,G}}$ is chosen to undergo stabilizing impulses, the reference state $\textbf{s}_{\emph{i,G}}$ is randomly selected from the current best individual (\emph{gbest}) or other individuals with better fitness values than $\textbf{x}_{\emph{i,G}}$ in the current population (see Section III.A). This operation not only introduces the information of elitist, but also avoids the premature convergence for the population during the evolution. In order to show the effectiveness of this operation, in this section, we compare the performance of ETI-DE and an ETI-DE modified in the following way (referred as ETIgb-DE): the reference state $\textbf{s}_{\emph{i,G}}$ is only selected from \emph{gbest} in stabilizing impulses. The detailed experimental data are provided in Tables S.17-S.19 of the supplementary file.

The results show that ETI-DEs are significantly better than ETIgb-DEs on most test functions. In ETI-DEs, the reference state of stabilizing impulses is set as \emph{gbest} or other individuals with better fitness values than $\textbf{x}_{\emph{i,G}}$, instead of merely \emph{gbest}. The setting optimizes the state of the whole population and increases the diversity of stabilizing impulses at the same time, which avoids the premature convergence of the population.
Therefore, we can conclude that it does make the search more effective when $\textbf{s}_{\emph{i,G}}$ is randomly selected from \emph{gbest} or other individuals with better fitness values.

\subsection{Comparison with Other Restart Strategies}
To the best of our knowledge, restart strategies directly replace the selected individuals with other individuals, without comparing the fitness values of them. Till now, different restart strategies have been proposed in DE \cite{vasile2011inflationary,peng2009multi,li2011new,zhabitsky2013asynchronous,yang2013improved,yang2015differential}. It is noticed that in our approach, destabilizing impulses serve as a restart strategy, which randomly adjust the positions of the inferior individuals within the area of the current population. Therefore, our presented ETI can also be viewed as a restart-based strategy. In this section, firstly, we illustrate the differences between some popular restart strategies in DE and ours; secondly, we compare our ETI with a latest restart strategy published in 2015.

To better illustrate the difference between other restart strategies and destabilizing impulses presented in this paper, we list the details of each strategy in Table S.20 of the supplementary file. In \cite{vasile2011inflationary}, if no improvement is observed for $\emph{n}_{\emph{samples}}$ sample points, a restart mechanism might be activated. Accordingly, a bubble is defined around the best individual within the cluster $\textbf{x}_{\emph{best}}$; then local and global restart strategies are performed inside and outside the bubble, respectively. However, the size of the bubble $\Delta$ is quite critical to the performance of the restart strategies.
In \cite{peng2009multi}, when the current population converges at a local optimum, a restart is activated. In detail, the newly generated individuals are forced away from the $\delta$ hypersphere neighborhood areas of previous local optima. Similarly, the neighborhood size $\delta$ is also important.
In \cite{li2011new}, the restart strategy takes effect when the predefined clusters are ``dead". The strategy consists of two operations: restart by DE/rand/2 and restart by reinitializing certain individuals within the search range. Specified probability values are assigned to each operation, and it is necessary to determine these values beforehand.
In \cite{zhabitsky2013asynchronous}, when stagnation is diagnosed, the algorithm performs a restart by increasing the population size by a predefined multiplier \emph{k} and starting an independent search.
In \cite{yang2013improved}, when the population diversity is poor or the population stagnates by measuring the Euclidean distances between individuals of a population, the individuals are restarted within the initial search space, which is sampled by a random number $\emph{randN}_{\emph{j,G}}$ with normal distribution.
In \cite{yang2015differential}, a new diversity enhance mechanism named auto-enhanced population diversity (AEPD) is proposed, which is an improved version of \cite{yang2013improved}. When population convergence or stagnation is identified by AEPD, some individuals are reinitialized. Our ETI is inspired by the idea of event-triggered mechanism (ETM) and impulsive control in control theory, and two kinds of impulses are developed  to enhance the exploitation and exploration performance of DE. Selected inferior individuals are restarted by being injected with destabilizing impulses when \emph{UR} drops to zero or stabilizing impulses fail to take effect (see Section III). Furthermore, the proposed ETI sheds light on the understandings of ETM and impulsive control in evolutionary computation, which broadens the applications of ETM and impulsive control in wider areas. Compared with the other methods in Table S.20, AEPD in \cite{yang2015differential} and our ETI are easy to implement: 1) they do not introduce any calculation of distances of individuals; 2) they do not use the neighborhood, which avoids determining the value of neighborhood size. The computational complexity of the original DE is $\mathcal{O}(\emph{G}_{\emph{max}}\cdot \emph{NP}\cdot \emph{D})$. While for AEPD and ETI, it is $\mathcal{O}(\emph{G}_{\emph{max}}\cdot ((3\cdot\emph{NP}+2)\cdot\emph{D}$)) and $\mathcal{O}(2\cdot\emph{G}_{\emph{max}}\cdot \emph{NP}\cdot(\emph{D}+\textrm{log}(\emph{NP}))$, respectively, both of which do not seriously increase the computational cost of the original DE.

In the following, we compare the performance of AEPD published in \cite{yang2015differential} in 2015 and ETI by applying them to jDE and JADE, and check their performance on the CEC 2014 benchmark functions. The parameters of jDE and JADE are given in Section IV.A. It is worth noting that in AEPD \cite{yang2015differential}, the population size \emph{NP} is set to 20. And in our ETI-jDE and ETI-JADE, \emph{NP} is set to 100. Therefore, the experiments are divided into three groups: 1) to compare the performance of AEPD-DEs and ETI-DEs with $\emph{NP}=20$; 2) to compare the performance of AEPD-DEs and ETI-DEs with $\emph{NP}=100$; 3) to compare the performance of AEPD-DEs with $\emph{NP}=20$ and ETI-DEs with $\emph{NP}=100$. The detailed experimental data are provided in Tables S.21-S.22 of the supplementary file. The upper half of Table S.21 demonstrates the superior performance of AEPD-jDE and AEPD-JADE with $\emph{NP}=20$; the lower half of Table S.21 confirms the outstanding performance of ETI-jDE and ETI-JADE with $\emph{NP}=100$. For Table S.22, AEPD-DEs and ETI-DEs use the \emph{NP} values recommended in \cite{yang2015differential} ($\emph{NP}=20$) and in our paper ($\emph{NP}=100$), respectively. The results show that ETI-jDE and ETI-JADE with $\emph{NP}=100$ perform better than AEPD-jDE and AEPD-JADE with $\emph{NP}=20$, which identifies the effectiveness of our ETI.
Compared with AEPD, our proposed ETI takes advantage of the concepts of impulsive control and ETM in control theory, which optimizes the state of the whole population by instantly altering the positions of partial individuals. The stabilizing and destabilizing impulses are triggered at certain moments, which enhances the exploitation and exploration abilities respectively, and saves the computational resources.

\subsection{Effectiveness of Ranking Assignment}
As introduced before, we select the individuals to undergo impulsive control by ranking the population based on two indices: fitness value and the number of consecutive stagnant generation. These two indices identify the state of each individual during the evolution process. Hence, $\tilde{\emph{R}}_\emph{i}$ and $\bar{\emph{R}}_\emph{i}$ are acquired, which are integrated into $\emph{R}_\emph{i}$. In this section, we carry out three classes of experiments to demonstrate the effectiveness of the ranking assignment. The first experiment compares the performance of ranking the population according to $\emph{R}_\emph{i}$ and only according to $\tilde{\emph{R}}_\emph{i}$; the second experiment displays the difference between utilizing $\emph{R}_\emph{i}$ and merely utilizing $\bar{\emph{R}}_\emph{i}$; the last experiment picks random individuals to be injected with impulses instead of using $\emph{R}_\emph{i}$. The corresponding three variants of ETI-DEs are as follows: R1-ETI-DE, R2-ETI-DE, and NoR-ETI-DE. These three variants are compared with ETI-DEs, the results of which (\emph{win}-\emph{lose} results of ETI-DEs \emph{vs.} R1-ETI-DEs, ETI-DEs \emph{vs.} R2-ETI-DEs, ETI-DEs \emph{vs.} NoR-ETI-DEs according to the Wilcoxon rank-sum test) are exhibited in Table S.23 of the supplementary file. Similar to the analysis of the last experiment, we also obtain three conclusions from Table S.23:

1) From the first two rows, it is observed that the individuals injected with impulsive controllers cannot be merely selected according to fitness values. The number of consecutive stagnant generation is also needed to be taken into consideration. For an individual, which has been updated in recent generations, although its fitness value is inferior, it may be near a promising area. Therefore, it is still necessary to keep this individual in the population.

2) From the third and fourth rows, it is noted that the individuals with impulsive control cannot be chosen by only considering the number of consecutive stagnant generation either. For an individual with superior fitness value, which stagnates in recent generations, it may still be helpful for the update of other individuals.

3) The last two rows further explain the necessity of using both fitness values and the number of consecutive stagnant generation to rank the individuals, which will be added impulses later. These two measures reflect the two most important characteristics of an individual in the evolution process. By adding impulses to the inferior individuals graded by these two measures, the search performance of the population can be enhanced.

\begin{remark}
${\emph{R}}_\emph{i}$ is the sum of $\tilde{\emph{R}}_\emph{i}$ and $\bar{\emph{R}}_\emph{i}$, where $\tilde{\emph{R}}_\emph{i}$ and $\bar{\emph{R}}_\emph{i}$ indicate the rankings of the individual according to the fitness value and the number of consecutive stagnation generation, respectively. In the following, we use jDE and ETI-jDE on F02 (unimodal) and F13 (multimodal) to show the impact of ETI on the population during the evolution process. In Fig. S.6 of the supplementary file, we plot the evolution of ten random individuals' ${\emph{R}}_\emph{i}$ by jDE and ETI-jDE on F02 and F13, respectively. From the figure, it can be observed that ${\emph{R}}_\emph{i}$ by ETI-jDE changes more frequently than that by jDE, which verifies that the introduction of ETI enhances the movement of the population.
\end{remark}

\subsection{Parameter Sensitivity Study}
In the proposed ETI, there are three parameters: \emph{pr}, \emph{LN}, and \emph{UN}. \emph{pr} represents the probability of selecting individuals to be injected with destabilizing impulses. \emph{LN} and \emph{UN} are the lower and upper bounds of the number of individuals with stabilizing impulses. In ETI, \emph{pr} is set as $0.2$, \emph{LN} is set as $1$, and \emph{UN} is set as the same as the population size \emph{NP}. Here, we set $\emph{pr}=[0.2, 0.6, 1.0]$, $\emph{LN}=[1, 0.2\emph{NP}, 0.5\emph{NP}, 0.8\emph{NP}]$, and $\emph{UN}=[0.1\emph{NP}, 0.4\emph{NP}, 0.7\emph{NP}, \emph{NP}]$. And we compare the ten DE variants and their corresponding ETI-DEs with different parameter values, in order to investigate the effects of the parameters on the performance of ETI-DEs.

In Figs. S.7-S.9 of the supplementary file, we use bar graph to show the number of functions that ETI-DEs with different parameter values are significantly better than, equal to, and worse than the original DE algorithms, respectively.

For \emph{pr}, most ETI-DEs (e.g., ETI-DE/best/1/bin, ETI-jDE, ETI-JADE, ETI-SaDE, ETI-EPSDE, and ETI-OXDE) with $\emph{pr}=0.2$ have more winning functions than that with $\emph{pr}=[0.6, 1.0]$. Besides, most ETI-DEs (e.g., ETI-jDE, ETI-JADE, ETI-CoDE, ETI-ODE, ETI-SHADE, and ETI-OXDE) with $\emph{pr}=0.2$ have fewer losing functions than that with $\emph{pr}=[0.6, 1.0]$. Then we add the \emph{win/tie/lose} numbers for all the algorithms when using the same value of \emph{pr}. And we find that $\emph{pr}=0.2$ is a little bit better than $\emph{pr}=[0.6, 1.0]$. A larger \emph{pr} denotes that more individuals will be injected with destabilizing impulses. And the introduction of more randomness may be harmful to the current research. Therefore, we set $\emph{pr}=0.2$ in the proposed ETI.

For \emph{LN}, along with the growth of \emph{LN}, we can find a trend for nine of the total ten ETI-DEs except ETI-DE/best/1/bin: the number of winning functions decreases and the number of losing function increases. And for ETI-DE/best/1/bin, its performance is not sensitive to the change of \emph{LN}. Then we add the \emph{win/tie/lose} numbers for all the algorithms when using the same value of \emph{LN}. And we find that $\emph{LN}=1$ is much better than $\emph{LN}=[0.2\emph{NP}, 0.5\emph{NP}, 0.8\emph{NP}]$. A larger \emph{LN} indicates that more individuals are injected with stabilizing impulses, which will disrupt the ongoing search. Therefore, we choose $\emph{LN}=1$ for our proposed ETI.

For \emph{UN}, we find that most ETI-DEs (e.g., ETI-DE/best/1/bin, ETI-JADE, ETI-SaDE, ETI-ODE, ETI-EPSDE, ETI-SHADE, and ETI-OXDE) with $\emph{UN}=\emph{NP}$ show better performance than that with $\emph{UN}=0.1\emph{NP}$. In general, the results indicate that too small values of \emph{UN} (e.g. $0.1\emph{NP}$) are worse than large values (e.g. $0.4\emph{NP}$, $0.7\emph{NP}$ and $\emph{NP}$). And along with the increase of \emph{UN}, the performance of ETI-DEs with different \emph{UN} values becomes close. \emph{UN} is the upper bound of the number of individuals with stabilizing impulses. In the experiments, we find that when ETI is adopted, it is more difficult for the number of individuals injected with stabilizing impulsive controllers to reach \emph{UN} as \emph{UN} becomes larger. And this explains why the performance of ETI-DEs becomes similar along with the increase of \emph{UN}. We add the \emph{win/tie/lose} numbers for all the algorithms when using the same value of \emph{UN}. And we find that it is a good choice to select $\emph{UN}=\emph{NP}$ in our paper.

\subsection{Scalability Study}
In the aforementioned experiments, we evaluate the performance of the algorithms by running them on 30 test functions with $\emph{D}=30$ from CEC 2014 test suite. The results show that our proposed ETI can improve the performance of the original DE variants. In this section, we perform the scalability study of the ETI to further examine its effectiveness. The dimension of the test functions is set as $\emph{D}=50$ and $\emph{D}=100$. The detailed experimental data and the results of the Holm-Bonferroni procedure are provided in Tables S.24-S.31 of the supplementary file. Furthermore, in Figs. S.10-S.13 of the supplementary file, we use the box plot to show the results of JADE, CoDE, SaDE and EPSDE with and without ETI on CEC 2014 test suite at $\emph{D}=100$.
From the results, we can find out that our developed scheme still works effectively for problems with large dimension ($\emph{D}=50$ and $\emph{D}=100$).
\subsection{Working Mechanism of ETI}
The above experimental results reveal the effectiveness of our proposed ETI. In this subsection, we will investigate the working mechanism of ETI based on some experiments.

In DE, the update rate (\emph{UR}) of the population reflects the degree of the activity of the population in the evolution process. The decrease of \emph{UR} indicates that the population gradually encounters stagnation. The worst situation is that \emph{UR} drops to zero before the global best solution is found. As introduced in Section III, two types of impulses, i.e., stabilizing and destabilizing impulses, are imposed on the selected individuals when \emph{UR} of the population in the current generation decreases or equals to zero. When \emph{UR} declines, certain individuals will instantly approach the reference states by means of stabilizing impulses. And we hope new superior individuals can be found, and help other individuals update. When \emph{UR} reduces to zero, selected individuals will be instantaneously displaced to other areas with the help of destabilizing impulses. These individuals in the new positions are expected to influence the stagnant population in a positive way.

In the following, we use jDE and ETI-jDE on the Shifted and Rotated Katsuura Function (i.e., F12 of the CEC 2014 test suite) to explain how ETI works. The results are given in Fig. S.14. Figs. S.14(a)-S.14(b) show the change of the fitness value of \emph{gbest} when F12 is optimized by jDE and ETI-jDE, respectively. Figs. S.14(c)-S.14(d) display the change of the value of \emph{UR} when F12 is optimized by jDE and ETI-jDE, respectively. Take a look at the status of the population from the 1700th generation to the 1750th generation. In Fig. S.14(c), the generations when $\emph{UR}=0$ are marked by red dots; and we find that \emph{UR} drops to zero very frequently, which means the whole population lacks the impetus to evolve. In Fig. S.14(d), the generations when $\emph{UR}=0$ are also marked by red dots, and red circles are used to mark the moments when \emph{UR} reduces. By observing the magnified figure in Fig. S.14(d), \emph{UR} does not drop to zero frequently, and we attribute it to the destabilizing impulses in ETI. When \emph{UR} decreases in the current generation, it seldom reduces again in the next generation due to the stabilizing impulses in ETI. Stabilizing and destabilizing impulses increase the degree of activity of the population during the evolution process, and hence help the update of \emph{gbest} (see Fig. S.14(b)).

The advantages of ETI can be summarized as follows:

1) Stabilizing impulses in ETI force certain individuals approach the reference states, which enhances
the exploitation ability of the algorithm.

2) Destabilizing impulses in ETI bring in more explorative moves, which increases the exploration ability of the algorithm.

3) Event-triggered mechanism (ETM) determines the moment of adding two types of impulses, which avoids the periodical execution of impulsive control and saves the computational resources.

4) ETI is easy to implement and flexible to be incorporated into other state-of-the-art DE variants.

\section{Conclusion}
In this paper, an event-triggered impulsive control scheme (ETI) has been proposed to improve the search performance of DE algorithms. There are four components in ETI: stabilizing impulses, destabilizing impulses, ranking assignment, and an adaptive mechanism. Firstly, stabilizing impulses aims to force the selected individuals to approach some promising areas instantly in the search domain, which facilitates the convergence of the population. Secondly, destabilizing impulses instantaneously alter the positions of inferior individuals in a certain area, which maintains the diversity of the population. Thirdly, ranking assignment is used to select the individuals to be injected with impulsive controllers, which is based on the fitness value and the number of consecutive stagnant generation of each individual. Fourthly, an adaptive mechanism is presented to determine the number of individuals taking impulsive control. Besides, ETM identifies the moment of imposing these two kinds of impulses. Meanwhile, the proposed ETI does not significantly increase the computational complexity of DE algorithms.

Extensive experiments have been carried out based on the CEC 2014 test suite. Firstly, ETI has been incorporated into two original DE algorithms and eight state-of-the-art DE variants. A series of results demonstrate that ETI can greatly improve the performance of these DE algorithms. Then several comparative experiments have been conducted to show the effectiveness of two types of impulses, random selection of the reference state in stabilizing impulses, and ranking assignment. Besides, we compare ETI with other restart strategies, and investigate the influence of three parameters. Then the presented ETI-DEs have also shown their superiority in high-dimensional problems. Finally, the working mechanism of ETI has been investigated based on some experiments. It is worth mentioning that the experimental results shed light on the understandings of ETM and impulsive control in evolutionary computation, which broadens the applications of ETM and impulsive control in wider areas.

\section*{Acknowledgment}
The authors would like to thank Dr. Y. Wang, Dr. M. Yang, and Dr. J. Wang for providing the source code of CoDE, AEPD, and OXDE, respectively.

\ifCLASSOPTIONcaptionsoff
  \newpage
\fi

{\footnotesize\bibliography{reference/ref}
\bibliographystyle{ieeetr}}
\clearpage

\center{\LARGE{\textbf{Supplementary file of ETI-DE}}}

\vspace{1cm}

\section*{Algorithm Captions}
\begin{itemize}
\item[$\bullet$] \textbf{Algorithm S.1} Stabilizing\_Impulsive\_Control ().
\item[$\bullet$] \textbf{Algorithm S.2} Injecting\_Destabilizing\_Impulses ().
\item[$\bullet$] \textbf{Algorithm S.3} Random\_Selection\_of\_Individuals ().
\end{itemize}

\section*{Table Captions}
\begin{itemize}
\item[$\bullet$] \textbf{Table \ref{hb1}} Holm test on the fitness, reference algorithm = DE/rand/1/bin ($\emph{NP}=100$, rank=\textbf{2.97}) for CEC 2014 test suite at $\emph{D}=30$.
\item[$\bullet$] \textbf{Table \ref{hb2}} Holm test on the fitness, reference algorithm = DE/best/1/bin ($\emph{NP}=100$, rank=2.47) for CEC 2014 test suite at $\emph{D}=30$.
\item[$\bullet$] \textbf{Table \ref{hb3}} Holm test on the fitness, reference algorithm = jDE ($\emph{NP}=100$, rank=\textbf{2.73}) for CEC 2014 test suite at $\emph{D}=30$.
\item[$\bullet$] \textbf{Table \ref{hb4}} Holm test on the fitness, reference algorithm = JADE ($\emph{NP}=100$, rank=\textbf{3.00}) for CEC 2014 test suite at $\emph{D}=30$.
\item[$\bullet$] \textbf{Table \ref{hb5}} Holm test on the fitness, reference algorithm = CoDE ($\emph{NP}=100$, rank=2.23) for CEC 2014 test suite at $\emph{D}=30$.
\item[$\bullet$] \textbf{Table \ref{hb6}} Holm test on the fitness, reference algorithm = SaDE ($\emph{NP}=100$, rank=\textbf{2.97}) for CEC 2014 test suite at $\emph{D}=30$.
\item[$\bullet$] \textbf{Table \ref{hb7}} Holm test on the fitness, reference algorithm = ODE ($\emph{NP}=100$, rank=\textbf{3.17}) for CEC 2014 test suite at $\emph{D}=30$.
\item[$\bullet$] \textbf{Table \ref{hb8}} Holm test on the fitness, reference algorithm = EPSDE ($\emph{NP}=100$, rank=2.70) for CEC 2014 test suite at $\emph{D}=30$.
\item[$\bullet$] \textbf{Table \ref{hb9}} Holm test on the fitness, reference algorithm = SHADE ($\emph{NP}=100$, rank=3.13) for CEC 2014 test suite at $\emph{D}=30$.
\item[$\bullet$] \textbf{Table \ref{hb10}} Holm test on the fitness, reference algorithm = OXDE ($\emph{NP}=100$, rank=\textbf{3.07}) for CEC 2014 test suite at $\emph{D}=30$.
\item[$\bullet$] \textbf{Table \ref{tabledm}} \emph{WIN}-\emph{LOSE} results of ETI-DEs with random \emph{DM} and with $\emph{DM}=[\emph{D}/3,2\emph{D}/3,\emph{D}]$ (in Section IV.A) for functions F01-F30 at $\emph{D}=30$.
\item[$\bullet$] \textbf{Table \ref{table30d1}} Experimental results of DE/rand/1/bin, DE/best/1/bin, jDE, JADE and the related ETI-based variants for functions F01-F30 at $\emph{D}=30$.
\item[$\bullet$] \textbf{Table \ref{table30d2}} Experimental results of CoDE, SaDE, ODE, EPSDE and the related ETI-based variants for functions F01-F30 at $\emph{D}=30$.
\item[$\bullet$] \textbf{Table \ref{table30d3}} Experimental results of SHADE, OXDE and the related ETI-based variants for functions F01-F30 at $\emph{D}=30$.
\item[$\bullet$] \textbf{Table \ref{hbtest}} Holm test on the fitness, reference algorithm = ETI-SHADE (rank=14.47) for functions F01-F30 at $\emph{D}=30$.
\item[$\bullet$] \textbf{Table \ref{tablec1}} \emph{WIN}-\emph{LOSE} results of ETI1-DEs, ETI2-DEs, ETI3-DEs, ETI4-DEs (in Section IV.C) and their counterparts for functions F01-F30 at $\emph{D}=30$.
\item[$\bullet$] \textbf{Table \ref{tablegb1}} Experimental results of ETI-DE/rand/1/bin, ETI-DE/best/1/bin, ETI-jDE, ETI-JADE and the related ETIgb-DEs for functions F01-F30 at $\emph{D}=30$.
\item[$\bullet$] \textbf{Table \ref{tablegb2}} Experimental results of ETI-CoDE, ETI-SaDE, ETI-ODE, ETI-EPSDE and the related ETIgb-DEs for functions F01-F30 at $\emph{D}=30$.
\item[$\bullet$] \textbf{Table \ref{tablegb3}} Experimental results of ETI-SHADE, ETI-OXDE and the related ETIgb-DEs for functions F01-F30 at $\emph{D}=30$.
\item[$\bullet$] \textbf{Table \ref{restt}} Different restart strategies in DE.
\item[$\bullet$] \textbf{Table \ref{tablers1}} Experimental results of ETI-DEs and AEPD-DEs for functions F01-F30 at $\emph{D}=30$.
\item[$\bullet$] \textbf{Table \ref{tablers2}} Experimental results of ETI-DEs and AEPD-DEs for functions F01-F30 at $\emph{D}=30$.
\item[$\bullet$] \textbf{Table \ref{tablera}} \emph{WIN}-\emph{LOSE} results of R1-ETI-DEs, R2-ETI-DEs, NoR-ETI-DEs (in Section IV.F) and ETI-DEs for functions F01-F30 at $\emph{D}=30$.
\item[$\bullet$] \textbf{Table \ref{table50d1}} Experimental results of DE/rand/1/bin, DE/best/1/bin, jDE, JADE and the related ETI-based variants for functions F01-F30 at $\emph{D}=50$.
\item[$\bullet$] \textbf{Table \ref{table50d2}} Experimental results of CoDE, SaDE, ODE, EPSDE and the related ETI-based variants for functions F01-F30 at $\emph{D}=50$.
\item[$\bullet$] \textbf{Table \ref{table50d3}} Experimental results of SHADE, OXDE and the related ETI-based variants for functions F01-F30 at $\emph{D}=50$.
\item[$\bullet$] \textbf{Table \ref{table100d1}} Experimental results of DE/rand/1/bin, DE/best/1/bin, jDE, JADE and the related ETI-based variants for functions F01-F30 at $\emph{D}=100$.
\item[$\bullet$] \textbf{Table \ref{table100d2}} Experimental results of CoDE, SaDE, ODE, EPSDE and the related ETI-based variants for functions F01-F30 at $\emph{D}=100$.
\item[$\bullet$] \textbf{Table \ref{table100d3}} Experimental results of SHADE, OXDE and the related ETI-based variants for functions F01-F30 at $\emph{D}=100$.
\item[$\bullet$] \textbf{Table \ref{hbtest50d}} Holm test on the fitness, reference algorithm = ETI-SHADE (rank=15.70) for functions F01-F30 at $\emph{D}=50$.
\item[$\bullet$] \textbf{Table \ref{hbtest100d}} Holm test on the fitness, reference algorithm = ETI-SHADE (rank=15.50) for functions F01-F30 at $\emph{D}=100$.
\end{itemize}


\section*{Figure Captions}
\begin{itemize}
\item[$\bullet$] \textbf{Fig. \ref{jadebp30}} Box plots for the results of JADE with/without ETI on CEC 2014 test suite at $\emph{D}=30$: 1--JADE; 2--ETI-JADE.
\item[$\bullet$] \textbf{Fig. \ref{codebp30}} Box plots for the results of CoDE with/without ETI on CEC 2014 test suite at $\emph{D}=30$: 1--CoDE; 2--ETI-CoDE.
\item[$\bullet$] \textbf{Fig. \ref{sadebp30}} Box plots for the results of SaDE with/without ETI on CEC 2014 test suite at $\emph{D}=30$: 1--SaDE; 2--ETI-SaDE.
\item[$\bullet$] \textbf{Fig. \ref{epsdebp30}} Box plots for the results of EPSDE with/without ETI on CEC 2014 test suite at $\emph{D}=30$: 1--EPSDE; 2--ETI-EPSDE.
\item[$\bullet$] \textbf{Fig. \ref{figcf}} Evolution of the mean function error values obtained from the algorithms versus the number of FES on six 30-dimensional test functions. (a) F02; (b) F11; (c) F15; (d) F19; (e) F20; (f) F26.
\item[$\bullet$] \textbf{Fig. \ref{figR}}  Evolution of $\emph{R}_\emph{i}$ by jDE and ETI-jDE on F02 and F13 at $\emph{D}=30$. (a) Evolution of ten random individuals' $\emph{R}_\emph{i}$ by jDE on F02. (b) Evolution of ten random individuals' $\emph{R}_\emph{i}$ by ETI-jDE on F02. (c) Evolution of ten random individuals' $\emph{R}_\emph{i}$ by jDE on F13. (d) Evolution of ten random individuals' $\emph{R}_\emph{i}$ by ETI-jDE on F13.
\item[$\bullet$] \textbf{Fig. \ref{prplot}} The number of functions that ETI-DEs with different \emph{pr} values ($\emph{LN}=1,\emph{UN}=100$) are significantly better than, equal to and worse than the original DEs on CEC 2014 test suite at $\emph{D}=30$. (The results of adding the \emph{win/tie/lose} numbers for all the algorithms when using the same value of \emph{pr}: $\emph{pr}=0.2: 181/97/22; \emph{pr}=0.6: 177/91/32; \emph{pr}=1.0: 172/90/38$.)
\item[$\bullet$] \textbf{Fig. \ref{lnplot}} The number of functions that ETI-DEs with different \emph{LN} values ($\emph{pr}=0.2,\emph{UN}=100$) are significantly better than, equal to and worse than the original DEs on CEC 2014 test suite at $\emph{D}=30$. (The results of adding the \emph{win/tie/lose} numbers for all the algorithms when using the same value of \emph{LN}: $\emph{LN}=1: 181/97/22; \emph{LN}=20: 170/71/59; \emph{LN}=50: 149/55/96; \emph{LN}=80: 133/36/131$.)
\item[$\bullet$] \textbf{Fig. \ref{unplot}} The number of functions that ETI-DEs with different \emph{UN} values ($\emph{pr}=0.2,\emph{LN}=1$) are significantly better than, equal to and worse than the original DEs on CEC 2014 test suite at $\emph{D}=30$. (The results of adding the \emph{win/tie/lose} numbers for all the algorithms when using the same value of \emph{UN}: $\emph{UN}=5: 152/120/28; \emph{UN}=20: 172/105/23; \emph{UN}=50: 175/96/29; \emph{UN}=100: 181/97/22$.)
\item[$\bullet$] \textbf{Fig. \ref{jadebp100}} Box plots for the results of JADE with/without ETI on CEC 2014 test suite at $\emph{D}=100$: 1--JADE; 2--ETI-JADE.
\item[$\bullet$] \textbf{Fig. \ref{codebp100}} Box plots for the results of CoDE with/without ETI on CEC 2014 test suite at $\emph{D}=100$: 1--CoDE; 2--ETI-CoDE.
\item[$\bullet$] \textbf{Fig. \ref{sadebp100}} Box plots for the results of SaDE with/without ETI on CEC 2014 test suite at $\emph{D}=100$: 1--SaDE; 2--ETI-SaDE.
\item[$\bullet$] \textbf{Fig. \ref{epsdebp100}} Box plots for the results of EPSDE with/without ETI on CEC 2014 test suite at $\emph{D}=100$: 1--EPSDE; 2--ETI-EPSDE.
\item[$\bullet$] \textbf{Fig. \ref{wm}} Working mechanism of ETI by jDE and ETI-jDE on F12 at $\emph{D}=30$. (a) Change of the fitness value of \emph{gbest} of F12 optimized by jDE. (b) Change of the fitness value of \emph{gbest} of F12 optimized by ETI-jDE. (c) Change of the value of \emph{UR} of F12 optimized by jDE. (d) Change of the value of \emph{UR} of F12 optimized by ETI-jDE.
\end{itemize}

\clearpage

\begin{algorithm}[!t]
\setcounter{algorithm}{0}
\renewcommand\thealgorithm{S.\arabic{algorithm}}
\scriptsize{
\caption{Stabilizing\_Impulsive\_Control ()} \algblock{Begin}{End}
\begin{algorithmic}[1]
\Begin
\State /* $\textbf{x}_{\emph{i,G}}$ is the individual that undergoes stabilizing impulsive control
\State /* $\zeta_{\emph{i,G}}$ is a flag to indicate whether stabilizing impulsive control is to improve the fitness value
\State /* rand(a,b) uniformly generate a random number belonging to the interval (a,b)
\State /* \emph{DM} is the number of dimensions selected to undergo stabilizing impulsive control in $\textbf{x}_{\emph{i,G}}$
\State $\textbf{A}=\{1,2,...,\emph{D}\}$; $\textbf{B}=\emptyset$; $\textbf{C}=\emptyset$
\State Randomly select an individual $\textbf{x}_{\emph{k,G}}$ from the current population
\If {$f(\textbf{x}_{\emph{i,G}})<f(\textbf{x}_{\emph{k,G}})$}
\State Set $\textbf{x}_{\emph{gbest,G}}$ as the reference state $\textbf{s}_{\emph{i,G}}$
\State Generate $\textbf{B}$ by randomly selecting \emph{DM} elements from $\textbf{A}$
\For {$\emph{j}=1$ to $\emph{D}$}
\If {$\emph{j}\in\textbf{B}$}
\State $\emph{K}_{\emph{ij,G}}=\textrm{rand}(-1,0)$
\Else
\State $\emph{K}_{\emph{ij,G}}=0$
\EndIf
\EndFor
\State $\emph{K}_{\emph{i,G}}=\textrm{diag}\{\emph{K}_{\emph{i1,G}},\emph{K}_{\emph{i2,G}},...,\emph{K}_{\emph{i}\emph{D,G}}\}_{\emph{D}\times\emph{D}}$, $\emph{i}=1,2,...,\emph{NP}$
\State $\textbf{e}_{\emph{i,G}}=\textbf{x}_{\emph{i,G}}-\textbf{s}_{\emph{i,G}}$
\State $\textbf{x}_{\emph{i,G}^+}=\textbf{x}_{\emph{i,G}}+\emph{K}_{\emph{i,G}}\cdot\textbf{e}_{\emph{i,G}}$
\Else
\State Set $\textbf{x}_{\emph{k,G}}$ as the reference state $\textbf{s}_{\emph{i,G}}$
\State Generate $\textbf{C}$ by randomly selecting \emph{DM} elements from $\textbf{A}$
\For {$\emph{j}=1$ to $\emph{D}$}
\If {$\emph{j}\in\textbf{C}$}
\State $\emph{K}_{\emph{ij,G}}=-1$
\Else
\State $\emph{K}_{\emph{ij,G}}=0$
\EndIf
\EndFor
\State $\emph{K}_{\emph{i,G}}=\textrm{diag}\{\emph{K}_{\emph{i1,G}},\emph{K}_{\emph{i2,G}},...,\emph{K}_{\emph{i}\emph{D,G}}\}_{\emph{D}\times\emph{D}}$, $\emph{i}=1,2,...,\emph{NP}$
\State $\textbf{e}_{\emph{i,G}}=\textbf{x}_{\emph{i,G}}-\textbf{s}_{\emph{i,G}}$
\State $\textbf{x}_{\emph{i,G}^+}=\textbf{x}_{\emph{i,G}}+\emph{K}_{\emph{i,G}}\cdot\textbf{e}_{\emph{i,G}}$
\EndIf
\If {$f(\textbf{x}_{\emph{i,G}^+}) \leq f(\textbf{x}_{\emph{i,G}})$}
\State $\textbf{x}_{\emph{i,G}} = \textbf{x}_{\emph{i,G}^+}$
\State $\zeta_{\emph{i,G}}=1$
\Else
\State $\zeta_{\emph{i,G}}=0$
\EndIf
\End
\end{algorithmic}}
\end{algorithm}

\begin{algorithm}[!t]
\renewcommand\thealgorithm{S.\arabic{algorithm}}
\scriptsize{
\caption{\textcolor{black}{Injecting\_Destabilizing\_Impulses ()}} \algblock{Begin}{End}
\begin{algorithmic}[1]
\Begin
\State /* $\textbf{x}_{\emph{i,G}}$ is the individual that undergoes destabilizing impulses
\State /* $\emph{min}_{\emph{j,G}}$ and $\emph{max}_{\emph{j,G}}$ are the minimum and maximum values of the \emph{j}th dimension in the population at the \emph{G}th generation
\State /* rand(a,b) uniformly generate a random number belonging to the interval (a,b)
\State $\textbf{x}_{\emph{L,G}}=[\emph{min}_{\emph{1,G}},\emph{min}_{\emph{2,G}},...,\emph{min}_{\emph{D,G}}]^{T}$
\State $\textbf{x}_{\emph{U,G}}=[\emph{max}_{\emph{1,G}},\emph{max}_{\emph{2,G}},...,\emph{max}_{\emph{D,G}}]^{T}$
\For {$\emph{j}=1$ to $\emph{D}$}
\State $\emph{K}_{\emph{ij,G}}=\textrm{rand}(0,1)$
\EndFor
\State $\emph{K}_{\emph{i,G}}=\textrm{diag}\{\emph{K}_{\emph{i1,G}},\emph{K}_{\emph{i2,G}},...,\emph{K}_{\emph{i}\emph{D,G}}\}_{\emph{D}\times\emph{D}}$, $\emph{i}=1,2,...,\emph{NP}$
\State $\textbf{e}_{\emph{i,G}}=\textbf{x}_{\emph{U,G}}-\textbf{x}_{\emph{L,G}}$
\State $\textbf{x}_{\emph{i,G}^+}=\textbf{x}_{\emph{L,G}}+\emph{K}_{\emph{i,G}}\cdot\textbf{e}_{\emph{i,G}}$
\End
\end{algorithmic}}
\end{algorithm}

\begin{algorithm}[!t]
\renewcommand\thealgorithm{S.\arabic{algorithm}}
\scriptsize{
\caption{Random\_Selection\_of\_Individuals ()} \algblock{Begin}{End}
\begin{algorithmic}[1]
\Begin
\State /* \{$\textbf{x}_{\emph{i,G}}|\emph{i}=1,2,...,\emph{M}$\} are the candidates that undergo destabilizing impulses
\State /* \emph{pr} is the probability for selecting individuals to be injected with destabilizing impulses
\State /* $\epsilon$ is a flag for judging whether it is necessary to increase \emph{pr}
\For {$\emph{i}=1$ to \emph{M}}
\State $\emph{r}_\emph{i}=\textrm{rand}$
\EndFor
\State $\emph{pr}=0.2$, $\epsilon=0$
\While {$\epsilon=0$}
\For {$\emph{i}=1$ to \emph{M}}
\If {$\emph{r}_\emph{i}<\emph{pr}$}
\State $\textbf{x}_{\emph{i,G}}$ will undergo destabilizing impulsive control later
\State $\epsilon=1$
\EndIf
\EndFor
\State $\emph{pr}=\emph{pr}+0.2$
\State $\emph{pr}=\textrm{min}(\emph{pr},1.0)$
\EndWhile
\End
\end{algorithmic}}
\end{algorithm}

\renewcommand\thetable{S.\arabic{table}}
\begin{table*}[!htbp]
\footnotesize
\centering
\caption{Holm test on the fitness, reference algorithm = DE/rand/1/bin ($\emph{NP}=100$, rank=\textbf{2.97}) for CEC 2014 test suite at $\emph{D}=30$.}\label{hb1}
\begin{tabular}{|c|c|c|c|c|c|c|}
\hline
\textbf{\emph{j}} & \textbf{Optimizer} & \textbf{Rank} & \textbf{$\emph{z}_\emph{j}$} & \textbf{$\emph{p}_\emph{j}$} & \textbf{$\bm\delta/\emph{j}$} & \textbf{Hypothesis} \\\hline
1     & DE/rand/1/bin ($\emph{NP}=30$) & 1.87 & -3.30E+00 & 4.83E-04 & 5.00E-02 & Rejected \\\hline
2     & DE/rand/1/bin ($\emph{NP}=50$) & 2.77 & -6.00E-01 & 2.74E-01 & 2.50E-02 & Accepted \\\hline
3     & DE/rand/1/bin ($\emph{NP}=150$) & 2.40 & -1.70E+00 & 4.46E-02 & 1.67E-02 & Accepted \\\hline
\end{tabular}
\end{table*}

\renewcommand\thetable{S.\arabic{table}}
\begin{table*}[!htbp]
\footnotesize
\centering
\caption{Holm test on the fitness, reference algorithm = DE/best/1/bin ($\emph{NP}=100$, rank=2.47) for CEC 2014 test suite at $\emph{D}=30$.}\label{hb2}
\begin{tabular}{|c|c|c|c|c|c|c|}
\hline
\textbf{\emph{j}} & \textbf{Optimizer} & \textbf{Rank} & \textbf{$\emph{z}_\emph{j}$} & \textbf{$\emph{p}_\emph{j}$} & \textbf{$\bm\delta/\emph{j}$} & \textbf{Hypothesis} \\\hline
1     & DE/best/1/bin ($\emph{NP}=30$) & 2.77 & 9.00E-01 & 8.16E-01 & 5.00E-02 & Accepted \\\hline
2     & DE/best/1/bin ($\emph{NP}=50$) & \textbf{3.03} & 1.70E+00 & 9.55E-01 & 2.50E-02 & Accepted \\\hline
3     & DE/best/1/bin ($\emph{NP}=150$) & 1.73 & -2.20E+00 & 1.39E-02 & 1.67E-02 & Rejected \\\hline
\end{tabular}
\end{table*}

\renewcommand\thetable{S.\arabic{table}}
\begin{table*}[!htbp]
\footnotesize
\centering
\caption{Holm test on the fitness, reference algorithm = jDE ($\emph{NP}=100$, rank=\textbf{2.73}) for CEC 2014 test suite at $\emph{D}=30$.}\label{hb3}
\begin{tabular}{|c|c|c|c|c|c|c|}
\hline
\textbf{\emph{j}} & \textbf{Optimizer} & \textbf{Rank} & \textbf{$\emph{z}_\emph{j}$} & \textbf{$\emph{p}_\emph{j}$} & \textbf{$\bm\delta/\emph{j}$} & \textbf{Hypothesis} \\\hline
1     & jDE ($\emph{NP}=30$) & 2.20 & -1.60E+00 & 5.48E-02 & 5.00E-02 & Accepted \\\hline
2     & jDE ($\emph{NP}=50$) & \textbf{2.73} & 0.00E+00 & 5.00E-01 & 2.50E-02 & Accepted \\\hline
3     & jDE ($\emph{NP}=150$) & 2.33 & -1.20E+00 & 1.15E-01 & 1.67E-02 & Accepted \\\hline
\end{tabular}
\end{table*}

\renewcommand\thetable{S.\arabic{table}}
\begin{table*}[!htbp]
\footnotesize
\centering
\caption{Holm test on the fitness, reference algorithm = JADE ($\emph{NP}=100$, rank=\textbf{3.00}) for CEC 2014 test suite at $\emph{D}=30$.}\label{hb4}
\begin{tabular}{|c|c|c|c|c|c|c|}
\hline
\textbf{\emph{j}} & \textbf{Optimizer} & \textbf{Rank} & \textbf{$\emph{z}_\emph{j}$} & \textbf{$\emph{p}_\emph{j}$} & \textbf{$\bm\delta/\emph{j}$} & \textbf{Hypothesis} \\\hline
1     & JADE ($\emph{NP}=30$) & 1.67 & -4.00E+00 & 3.17E-05 & 5.00E-02 & Rejected \\\hline
2     & JADE ($\emph{NP}=50$) & 2.53 & -1.40E+00 & 8.08E-02 & 2.50E-02 & Accepted \\\hline
3     & JADE ($\emph{NP}=150$) & 2.80 & -6.00E-01 & 2.74E-01 & 1.67E-02 & Accepted \\\hline
\end{tabular}
\end{table*}

\renewcommand\thetable{S.\arabic{table}}
\begin{table*}[!htbp]
\footnotesize
\centering
\caption{Holm test on the fitness, reference algorithm = CoDE ($\emph{NP}=100$, rank=2.23) for CEC 2014 test suite at $\emph{D}=30$.}\label{hb5}
\begin{tabular}{|c|c|c|c|c|c|c|}
\hline
\textbf{\emph{j}} & \textbf{Optimizer} & \textbf{Rank} & \textbf{$\emph{z}_\emph{j}$} & \textbf{$\emph{p}_\emph{j}$} & \textbf{$\bm\delta/\emph{j}$} & \textbf{Hypothesis} \\\hline
1     & CoDE ($\emph{NP}=30$) & 3.27 & 3.10E+00 & 9.99E-01 & 5.00E-02 & Accepted \\\hline
2     & CoDE ($\emph{NP}=50$) & \textbf{3.50} & 3.80E+00 & 1.00E+00 & 2.50E-02 & Accepted \\\hline
3     & CoDE ($\emph{NP}=150$) & 1.00 & -3.70E+00 & 1.08E-04 & 1.67E-02 & Rejected \\\hline
\end{tabular}
\end{table*}

\clearpage

\renewcommand\thetable{S.\arabic{table}}
\begin{table*}[!htbp]
\footnotesize
\centering
\caption{Holm test on the fitness, reference algorithm = SaDE ($\emph{NP}=100$, rank=\textbf{2.97}) for CEC 2014 test suite at $\emph{D}=30$.}\label{hb6}
\begin{tabular}{|c|c|c|c|c|c|c|}
\hline
\textbf{\emph{j}} & \textbf{Optimizer} & \textbf{Rank} & \textbf{$\emph{z}_\emph{j}$} & \textbf{$\emph{p}_\emph{j}$} & \textbf{$\bm\delta/\emph{j}$} & \textbf{Hypothesis} \\\hline
1     & SaDE ($\emph{NP}=30$) & 1.73 & -3.70E+00 & 1.08E-04 & 5.00E-02 & Rejected \\\hline
2     & SaDE ($\emph{NP}=50$) & 2.53 & -1.30E+00 & 9.68E-02 & 2.50E-02 & Accepted \\\hline
3     & SaDE ($\emph{NP}=150$) & 2.77 & -6.00E-01 & 2.74E-01 & 1.67E-02 & Accepted \\\hline
\end{tabular}
\end{table*}

\renewcommand\thetable{S.\arabic{table}}
\begin{table*}[!htbp]
\footnotesize
\centering
\caption{Holm test on the fitness, reference algorithm = ODE ($\emph{NP}=100$, rank=\textbf{3.17}) for CEC 2014 test suite at $\emph{D}=30$.}\label{hb7}
\begin{tabular}{|c|c|c|c|c|c|c|}
\hline
\textbf{\emph{j}} & \textbf{Optimizer} & \textbf{Rank} & \textbf{$\emph{z}_\emph{j}$} & \textbf{$\emph{p}_\emph{j}$} & \textbf{$\bm\delta/\emph{j}$} & \textbf{Hypothesis} \\\hline
1     & ODE ($\emph{NP}=30$) & 1.57 & -4.80E+00 & 7.93E-07 & 5.00E-02 & Rejected \\\hline
2     & ODE ($\emph{NP}=50$) & 2.87 & -9.00E-01 & 1.84E-01 & 2.50E-02 & Accepted \\\hline
3     & ODE ($\emph{NP}=150$) & 2.40 & -2.30E+00 & 1.07E-02 & 1.67E-02 & Rejected \\\hline
\end{tabular}
\end{table*}

\renewcommand\thetable{S.\arabic{table}}
\begin{table*}[!htbp]
\footnotesize
\centering
\caption{Holm test on the fitness, reference algorithm = EPSDE ($\emph{NP}=100$, rank=2.70) for CEC 2014 test suite at $\emph{D}=30$.}\label{hb8}
\begin{tabular}{|c|c|c|c|c|c|c|}
\hline
\textbf{\emph{j}} & \textbf{Optimizer} & \textbf{Rank} & \textbf{$\emph{z}_\emph{j}$} & \textbf{$\emph{p}_\emph{j}$} & \textbf{$\bm\delta/\emph{j}$} & \textbf{Hypothesis} \\\hline
1     & EPSDE ($\emph{NP}=30$) & 2.30 & -1.20E+00 & 1.15E-01 & 5.00E-02 & Accepted \\\hline
2     & EPSDE ($\emph{NP}=50$) & \textbf{2.80} & 3.00E-01 & 6.18E-01 & 2.50E-02 & Accepted \\\hline
3     & EPSDE ($\emph{NP}=150$) & 2.20 & -1.50E+00 & 6.68E-02 & 1.67E-02 & Accepted \\\hline
\end{tabular}
\end{table*}

\renewcommand\thetable{S.\arabic{table}}
\begin{table*}[!htbp]
\footnotesize
\centering
\caption{Holm test on the fitness, reference algorithm = SHADE ($\emph{NP}=100$, rank=3.13) for CEC 2014 test suite at $\emph{D}=30$.}\label{hb9}
\begin{tabular}{|c|c|c|c|c|c|c|}
\hline
\textbf{\emph{j}} & \textbf{Optimizer} & \textbf{Rank} & \textbf{$\emph{z}_\emph{j}$} & \textbf{$\emph{p}_\emph{j}$} & \textbf{$\bm\delta/\emph{j}$} & \textbf{Hypothesis} \\\hline
1     & SHADE ($\emph{NP}=30$) & 1.37 & -5.30E+00 & 5.79E-08 & 5.00E-02 & Rejected \\\hline
2     & SHADE ($\emph{NP}=50$) & 2.17 & -2.90E+00 & 1.90E-03 & 2.50E-02 & Rejected \\\hline
3     & SHADE ($\emph{NP}=150$) & \textbf{3.33} & 6.00E-01 & 7.26E-01 & 1.67E-02 & Accepted \\\hline
\end{tabular}
\end{table*}

\renewcommand\thetable{S.\arabic{table}}
\begin{table*}[!htbp]
\footnotesize
\centering
\caption{Holm test on the fitness, reference algorithm = OXDE ($\emph{NP}=100$, rank=\textbf{3.07}) for CEC 2014 test suite at $\emph{D}=30$.}\label{hb10}
\begin{tabular}{|c|c|c|c|c|c|c|}
\hline
\textbf{\emph{j}} & \textbf{Optimizer} & \textbf{Rank} & \textbf{$\emph{z}_\emph{j}$} & \textbf{$\emph{p}_\emph{j}$} & \textbf{$\bm\delta/\emph{j}$} & \textbf{Hypothesis} \\\hline
1     & OXDE ($\emph{NP}=30$) & 1.90 & -3.50E+00 & 2.33E-04 & 5.00E-02 & Rejected \\\hline
2     & OXDE ($\emph{NP}=50$) & 2.70 & -1.10E+00 & 1.36E-01 & 2.50E-02 & Accepted \\\hline
3     & OXDE ($\emph{NP}=150$) & 2.33 & -2.20E+00 & 1.39E-02 & 1.67E-02 & Rejected \\\hline
\end{tabular}
\end{table*}

\renewcommand\thetable{S.\arabic{table}}
\begin{table*}[!htbp]
\footnotesize
\centering
\caption{\emph{WIN}-\emph{LOSE} results of ETI-DEs with random \emph{DM} and with $\emph{DM}=[\emph{D}/3,2\emph{D}/3,\emph{D}]$ (in Section IV.A) for functions F01-F30 at $\emph{D}=30$.}\label{tabledm}
\begin{tabular}{|c|c|c||c|c|c|}
\hline
 \multicolumn{3}{|c||}{\textbf{ETI-DE/rand/1/bin}} & \multicolumn{3}{c|}{\textbf{ETI-DE/best/1/bin}} \\\hline
 \tabincell{c}{random \emph{DM} \\\emph{vs.} $\emph{DM}=\emph{D}/3$} & \tabincell{c}{random \emph{DM} \\\emph{vs.} $\emph{DM}=2\emph{D}/3$} & \tabincell{c}{random \emph{DM} \\\emph{vs.} $\emph{DM}=\emph{D}$} & \tabincell{c}{random \emph{DM} \\\emph{vs.} $\emph{DM}=\emph{D}/3$} & \tabincell{c}{random \emph{DM} \\\emph{vs.} $\emph{DM}=2\emph{D}/3$} & \tabincell{c}{random \emph{DM} \\\emph{vs.} $\emph{DM}=\emph{D}$} \\\hline
 18-2 & 15-3 & 15-4 & 5-7 & 0-6 & 14-3 \\\shline
 \multicolumn{3}{|c||}{\textbf{ETI-jDE}} & \multicolumn{3}{c|}{\textbf{ETI-JADE}} \\\hline
 \tabincell{c}{random \emph{DM} \\\emph{vs.} $\emph{DM}=\emph{D}/3$} & \tabincell{c}{random \emph{DM} \\\emph{vs.} $\emph{DM}=2\emph{D}/3$} & \tabincell{c}{random \emph{DM} \\\emph{vs.} $\emph{DM}=\emph{D}$} & \tabincell{c}{random \emph{DM} \\\emph{vs.} $\emph{DM}=\emph{D}/3$} & \tabincell{c}{random \emph{DM} \\\emph{vs.} $\emph{DM}=2\emph{D}/3$} & \tabincell{c}{random \emph{DM} \\\emph{vs.} $\emph{DM}=\emph{D}$} \\\hline
 10-2 & 9-0 & 11-1 & 13-2 & 10-1 & 12-1 \\\shline
 \multicolumn{3}{|c||}{\textbf{ETI-CoDE}} & \multicolumn{3}{c|}{\textbf{ETI-SaDE}} \\\hline
 \tabincell{c}{random \emph{DM} \\\emph{vs.} $\emph{DM}=\emph{D}/3$} & \tabincell{c}{random \emph{DM} \\\emph{vs.} $\emph{DM}=2\emph{D}/3$} & \tabincell{c}{random \emph{DM} \\\emph{vs.} $\emph{DM}=\emph{D}$} & \tabincell{c}{random \emph{DM} \\\emph{vs.} $\emph{DM}=\emph{D}/3$} & \tabincell{c}{random \emph{DM} \\\emph{vs.} $\emph{DM}=2\emph{D}/3$} & \tabincell{c}{random \emph{DM} \\\emph{vs.} $\emph{DM}=\emph{D}$} \\
\end{tabular}
\end{table*}

\renewcommand\thetable{S.\arabic{table}}
\begin{table*}[!htbp]
\scriptsize
\centering
\caption{Experimental results of DE/rand/1/bin, DE/best/1/bin, jDE, JADE and the related ETI-based variants for functions F01-F30 at $\emph{D}=30$.}\label{table30d1}
\begin{tabular}{|c|cc|cc|}
\hline
\textbf{Function} & \textbf{DE/rand/1/bin} & \textbf{ETI-DE/rand/1/bin} & \textbf{DE/best/1/bin} & \textbf{ETI-DE/best/1/bin}  \\\hline
    F01   & \textbf{5.731E+04 $\pm$ 4.288E+04} & 9.916E+04 $\pm$ 6.514E+04$-$ & 1.308E+07 $\pm$ 5.457E+06 & \textbf{4.817E+05 $\pm$ 3.454E+05}$+$ \\
    F02   & 0.000E+00 $\pm$ 0.000E+00 & 0.000E+00 $\pm$ 0.000E+00$\approx$ & 0.000E+00 $\pm$ 0.000E+00 & 0.000E+00 $\pm$ 0.000E+00$\approx$ \\
    F03   & 0.000E+00 $\pm$ 0.000E+00 & 0.000E+00 $\pm$ 0.000E+00$\approx$ & 0.000E+00 $\pm$ 0.000E+00 & 0.000E+00 $\pm$ 0.000E+00$\approx$ \\\hline
    F04   & \textbf{2.340E-01 $\pm$ 1.217E-01} & 1.638E+00 $\pm$ 8.823E+00$-$ & 4.840E+01 $\pm$ 4.264E+01 & \textbf{1.467E+00 $\pm$ 8.864E+00}$+$ \\
    F05   & 2.094E+01 $\pm$ 5.340E-02 & \textbf{2.036E+01 $\pm$ 9.584E-02}$+$ & 2.090E+01 $\pm$ 6.099E-02 & \textbf{2.023E+01 $\pm$ 2.166E-01}$+$ \\
    F06   & \textbf{4.206E-01 $\pm$ 7.135E-01} & 6.459E-01 $\pm$ 9.502E-01$\approx$ & 1.750E+00 $\pm$ 1.490E+00 & \textbf{1.680E+00 $\pm$ 1.212E+00}$\approx$ \\
    F07   & 3.383E-04 $\pm$ 1.709E-03 & \textbf{0.000E+00 $\pm$ 0.000E+00}$\approx$ & 4.878E-03 $\pm$ 6.312E-03 & \textbf{3.769E-03 $\pm$ 4.837E-03}$\approx$ \\
\hline
\end{tabular}
\end{table*}

\renewcommand\thetable{S.\arabic{table}}
\begin{table*}[!htbp]
\scriptsize
\centering
\caption{Experimental results of CoDE, SaDE, ODE, EPSDE and the related ETI-based variants for functions F01-F30 at $\emph{D}=30$.}\label{table30d2}
\begin{tabular}{|c|cc|cc|}
\hline
\textbf{Function} & \textbf{CoDE} & \textbf{ETI-CoDE} & \textbf{SaDE} & \textbf{ETI-SaDE}  \\\hline
F01   & \textbf{4.761E+04 $\pm$ 5.099E+04} & 7.441E+04 $\pm$ 6.708E+04$-$ & \textbf{2.154E+05 $\pm$ 1.393E+05} & 3.172E+05 $\pm$ 2.056E+05$-$ \\
    F02   & 0.000E+00 $\pm$ 0.000E+00 & 0.000E+00 $\pm$ 0.000E+00$\approx$ & 0.000E+00 $\pm$ 0.000E+00 & 0.000E+00 $\pm$ 0.000E+00$\approx$ \\
    F03   & 0.000E+00 $\pm$ 0.000E+00 & 0.000E+00 $\pm$ 0.000E+00$\approx$ & 0.000E+00 $\pm$ 0.000E+00 & 0.000E+00 $\pm$ 0.000E+00$\approx$ \\\hline
    F04   & 1.246E+00 $\pm$ 8.878E+00 & \textbf{3.628E-03 $\pm$ 7.089E-03}$\approx$ & \textbf{4.273E+01 $\pm$ 3.633E+01} & 5.201E+01 $\pm$ 3.374E+01$\approx$ \\
    F05   & 2.055E+01 $\pm$ 4.724E-02 & \textbf{2.010E+01 $\pm$ 9.499E-02}$+$ & 2.057E+01 $\pm$ 5.161E-02 & \textbf{2.001E+01 $\pm$ 1.948E-02}$+$ \\
    F06   & \textbf{8.938E-01 $\pm$ 1.053E+00} & 9.090E-01 $\pm$ 1.437E+00$-$ & \textbf{1.719E+00 $\pm$ 1.057E+00} & 2.080E+00 $\pm$ 1.352E+00$\approx$ \\
    F07   & 0.000E+00 $\pm$ 0.000E+00 & 0.000E+00 $\pm$ 0.000E+00$\approx$ & \textbf{3.183E-03 $\pm$ 8.174E-03} & 3.955E-03 $\pm$ 8.490E-03$\approx$ \\
\hline
\end{tabular}
\end{table*}

\renewcommand\thetable{S.\arabic{table}}
\begin{table*}[!htbp]
\scriptsize
\centering
\caption{Experimental results of SHADE, OXDE and the related ETI-based variants for functions F01-F30 at $\emph{D}=30$.}\label{table30d3}
\begin{tabular}{|c|cc|cc|}
\hline
\textbf{Function} & \textbf{SHADE} & \textbf{ETI-SHADE} & \textbf{OXDE} & \textbf{ETI-OXDE}  \\\hline
F01   & 7.131E+00 $\pm$ 4.882E+01 & \textbf{4.551E+00 $\pm$ 1.395E+01}$+$ & \textbf{6.857E+04 $\pm$ 4.661E+04} & 1.223E+05 $\pm$ 1.088E+05$-$ \\
    F02   & 0.000E+00 $\pm$ 0.000E+00 & 0.000E+00 $\pm$ 0.000E+00$\approx$ & 0.000E+00 $\pm$ 0.000E+00 & 0.000E+00 $\pm$ 0.000E+00$\approx$ \\
    F03   & 0.000E+00 $\pm$ 0.000E+00 & 0.000E+00 $\pm$ 0.000E+00$\approx$ & 0.000E+00 $\pm$ 0.000E+00 & 0.000E+00 $\pm$ 0.000E+00$\approx$ \\\hline
    F04   & 0.000E+00 $\pm$ 0.000E+00 & 0.000E+00 $\pm$ 0.000E+00$\approx$ & \textbf{3.840E-01 $\pm$ 1.739E-01} & 1.790E+00 $\pm$ 8.803E+00$-$ \\
    F05   & \textbf{2.025E+01 $\pm$ 2.670E-02} & 2.027E+01 $\pm$ 6.343E-02$-$ & 2.094E+01 $\pm$ 5.352E-02 & \textbf{2.035E+01 $\pm$ 1.120E-01}$+$ \\
    F06   & 8.833E-02 $\pm$ 3.059E-01 & \textbf{9.006E-03 $\pm$ 6.431E-02}$\approx$ & \textbf{2.653E-01 $\pm$ 5.908E-01} & 4.959E-01 $\pm$ 7.787E-01$\approx$ \\
    F07   & 0.000E+00 $\pm$ 0.000E+00 & 0.000E+00 $\pm$ 0.000E+00$\approx$ & 1.933E-04 $\pm$ 1.380E-03 & \textbf{0.000E+00 $\pm$ 0.000E+00}$+$ \\
\hline
\end{tabular}
\end{table*}

\renewcommand\thetable{S.\arabic{table}}
\begin{table*}[!htbp]
\scriptsize
\centering
\caption{Holm test on the fitness, reference algorithm = ETI-SHADE (rank=14.47) for functions F01-F30 at $\emph{D}=30$.}\label{hbtest}
\begin{tabular}{|c|c|c|c|c|c|c|}
\hline
\textbf{\emph{j}} & \textbf{Optimizer} & \textbf{Rank} & \textbf{$\emph{z}_\emph{j}$} & \textbf{$\emph{p}_\emph{j}$} & \textbf{$\bm\delta/\emph{j}$} & \textbf{Hypothesis} \\\hline
    1     & ETI-CoDE & 14.10 & -2.40E-01 & 4.05E-01 & 5.00E-03 & Accepted \\\hline
    2     & ETI-OXDE & 13.93 & -3.49E-01 & 3.63E-01 & 2.63E-03 & Accepted \\\hline
    3     & ETI-DE/rand/1/bin & 13.57 & -5.89E-01 & 2.78E-01 & 2.50E-02 & Accepted \\\hline
    4     & ETI-jDE & 13.47 & -6.55E-01 & 2.56E-01 & 8.33E-03 & Accepted \\\hline
    5     & SHADE & 13.20 & -8.29E-01 & 2.03E-01 & 2.94E-03 & Accepted \\\hline
    6     & ETI-JADE & 13.03 & -9.38E-01 & 1.74E-01 & 6.25E-03 & Accepted \\\hline
    7     & ETI-ODE & 10.73 & -2.44E+00 & 7.26E-03 & 3.57E-03 & Accepted \\\hline
    8     & ETI-SaDE & 10.63 & -2.51E+00 & 6.05E-03 & 4.17E-03 & Accepted \\\hline
    9     & ETI-EPSDE & 10.63 & -2.51E+00 & 6.05E-03 & 3.13E-03 & Accepted \\\hline
    10    & CoDE  & 10.53 & -2.57E+00 & 5.01E-03 & 5.56E-03 & Rejected \\\hline
    11    & JADE  & 10.17 & -2.82E+00 & 2.44E-03 & 7.14E-03 & Rejected \\\hline
    12    & jDE   & 9.37  & -3.34E+00 & 4.21E-04 & 1.00E-02 & Rejected \\\hline
    13    & OXDE  & 9.13  & -3.49E+00 & 2.40E-04 & 2.78E-03 & Rejected \\\hline
    14    & DE/rand/1/bin & 8.83  & -3.69E+00 & 1.13E-04 & 5.00E-02 & Rejected \\\hline
    15    & ODE   & 8.27  & -4.06E+00 & 2.47E-05 & 3.85E-03 & Rejected \\\hline
    16    & SaDE  & 8.23  & -4.08E+00 & 2.25E-05 & 4.55E-03 & Rejected \\\hline
    17    & EPSDE & 7.97  & -4.26E+00 & 1.04E-05 & 3.33E-03 & Rejected \\\hline
    18    & ETI-DE/best/1/bin & 5.90  & -5.61E+00 & 1.02E-08 & 1.25E-02 & Rejected \\\hline
    19    & DE/best/1/bin & 3.83  & -6.96E+00 & 1.69E-12 & 1.67E-02 & Rejected \\\hline
\end{tabular}
\end{table*}

\renewcommand\thetable{S.\arabic{table}}
\begin{table*}[!htbp]
\scriptsize
\centering
\caption{\emph{WIN}-\emph{LOSE} results of ETI1-DEs, ETI2-DEs, ETI3-DEs, ETI4-DEs (in Section IV.C) and their counterparts for functions F01-F30 at $\emph{D}=30$.}\label{tablec1}
\begin{tabular}{|c|c|c|c|c|c|}
\hline
 \textbf{ETI-DEs} & \textbf{\emph{vs.} ETI1-DE/rand/1/bin} & \textbf{\emph{vs.} ETI1-DE/best/1/bin} & \textbf{\emph{vs.} ETI1-jDE} & \textbf{\emph{vs.} ETI1-JADE} & \textbf{\emph{vs.} ETI1-CoDE}  \\
 & 16-5 & 12-2 & 11-3 & 17-3 & 11-7 \\
 & \textbf{\emph{vs.} ETI1-SaDE} & \textbf{\emph{vs.} ETI1-ODE} & \textbf{\emph{vs.} ETI1-EPSDE} & \textbf{\emph{vs.} ETI1-SHADE} & \textbf{\emph{vs.} ETI1-OXDE} \\
 & 11-4 & 10-2 & 18-5 & 12-3 & 16-1 \\\hline
 \textbf{ETI-DEs} & \textbf{\emph{vs.} ETI2-DE/rand/1/bin} & \textbf{\emph{vs.} ETI2-DE/best/1/bin} & \textbf{\emph{vs.} ETI2-jDE} & \textbf{\emph{vs.} ETI2-JADE} & \textbf{\emph{vs.} ETI2-CoDE}  \\
 & 15-0 & 15-2 & 7-2 & 1-1 & 6-1 \\
 & \textbf{\emph{vs.} ETI2-SaDE} & \textbf{\emph{vs.} ETI2-ODE} & \textbf{\emph{vs.} ETI2-EPSDE} & \textbf{\emph{vs.} ETI2-SHADE} & \textbf{\emph{vs.} ETI2-OXDE} \\
 & 6-1 & 16-0 & 6-1 & 0-2 & 14-0 \\\hline
 \textbf{ETI-DEs} & \textbf{\emph{vs.} ETI3-DE/rand/1/bin} & \textbf{\emph{vs.} ETI3-DE/best/1/bin} & \textbf{\emph{vs.} ETI3-jDE} & \textbf{\emph{vs.} ETI3-JADE} & \textbf{\emph{vs.} ETI3-CoDE}  \\
 & 3-2 & 1-0 & 1-1 & 7-3 & 1-4 \\
 & \textbf{\emph{vs.} ETI3-SaDE} & \textbf{\emph{vs.} ETI3-ODE} & \textbf{\emph{vs.} ETI3-EPSDE} & \textbf{\emph{vs.} ETI3-SHADE} & \textbf{\emph{vs.} ETI3-OXDE}  \\
 & 1-2 & 1-1 & 4-2 & 5-8 & 1-1 \\\hline
  \textbf{ETI-DEs} & \textbf{\emph{vs.} ETI4-DE/rand/1/bin} & \textbf{\emph{vs.} ETI4-DE/best/1/bin} & \textbf{\emph{vs.} ETI4-jDE} & \textbf{\emph{vs.} ETI4-JADE} & \textbf{\emph{vs.} ETI4-CoDE}  \\
 & 18-2 & 16-4 & 10-1 & 12-3 & 12-1 \\
 & \textbf{\emph{vs.} ETI4-SaDE} & \textbf{\emph{vs.} ETI4-ODE} & \textbf{\emph{vs.} ETI4-EPSDE} & \textbf{\emph{vs.} ETI4-SHADE} & \textbf{\emph{vs.} ETI4-OXDE} \\
 & 10-3 & 16-1 & 16-3 & 8-8 & 16-1 \\\hline
\end{tabular}
\end{table*}

\renewcommand\thetable{S.\arabic{table}}
\begin{table*}[!htbp]
\scriptsize
\centering
\caption{Experimental results of ETI-DE/rand/1/bin, ETI-DE/best/1/bin, ETI-jDE, ETI-JADE and the related ETIgb-DEs for functions F01-F30 at $\emph{D}=30$.}\label{tablegb1}
\begin{tabular}{|c|cc|cc|}
\hline
\textbf{Function} & \textbf{ETI-DE/rand/1/bin} & \textbf{ETIgb-DE/rand/1/bin} & \textbf{ETI-DE/best/1/bin} & \textbf{ETIgb-DE/best/1/bin}  \\\hline
    F01   & \textbf{9.916E+04 $\pm$ 6.514E+04}$\approx$ & 1.315E+05 $\pm$ 1.155E+05 & 4.817E+05 $\pm$ 3.454E+05$\approx$ & \textbf{3.654E+05 $\pm$ 2.226E+05} \\
    F02   & 0.000E+00 $\pm$ 0.000E+00$\approx$ & 0.000E+00 $\pm$ 0.000E+00 & 0.000E+00 $\pm$ 0.000E+00$\approx$ & 0.000E+00 $\pm$ 0.000E+00 \\
    F03   & 0.000E+00 $\pm$ 0.000E+00$\approx$ & 0.000E+00 $\pm$ 0.000E+00 & 0.000E+00 $\pm$ 0.000E+00$\approx$ & 0.000E+00 $\pm$ 0.000E+00 \\\hline
    F04   & 1.638E+00 $\pm$ 8.823E+00$\approx$ & \textbf{3.795E-01 $\pm$ 1.412E-01} & \textbf{1.467E+00 $\pm$ 8.864E+00}$\approx$ & 7.973E+00 $\pm$ 2.159E+01 \\
    F05   & \textbf{2.036E+01 $\pm$ 9.584E-02}$+$ & 2.091E+01 $\pm$ 5.290E-02 & 2.023E+01 $\pm$ 2.166E-01$-$ & \textbf{2.017E+01 $\pm$ 2.595E-01} \\
    F06   & 6.459E-01 $\pm$ 9.502E-01$\approx$ & \textbf{3.888E-01 $\pm$ 6.374E-01} & \textbf{1.680E+00 $\pm$ 1.212E+00}$+$ & 2.480E+00 $\pm$ 1.665E+00 \\
    F07   & \textbf{0.000E+00 $\pm$ 0.000E+00}$\approx$ & 1.450E-04 $\pm$ 1.036E-03 & \textbf{3.769E-03 $\pm$ 4.837E-03}$\approx$ & 6.326E-03 $\pm$ 7.942E-03 \\
\hline
\end{tabular}
\end{table*}

\renewcommand\thetable{S.\arabic{table}}
\begin{table*}[!htbp]
\scriptsize
\centering
\caption{Experimental results of ETI-CoDE, ETI-SaDE, ETI-ODE, ETI-EPSDE and the related ETIgb-DEs for functions F01-F30 at $\emph{D}=30$.}\label{tablegb2}
\begin{tabular}{|c|cc|cc|}
\hline
\textbf{Function} & \textbf{ETI-CoDE} & \textbf{ETIgb-CoDE} & \textbf{ETI-SaDE} & \textbf{ETIgb-SaDE}  \\\hline
F01   & 7.441E+04 $\pm$ 6.708E+04$\approx$ & \textbf{5.576E+04 $\pm$ 3.649E+04} & \textbf{3.172E+05 $\pm$ 2.056E+05}$\approx$ & 3.225E+05 $\pm$ 2.123E+05 \\
    F02   & 0.000E+00 $\pm$ 0.000E+00$\approx$ & 0.000E+00 $\pm$ 0.000E+00 & 0.000E+00 $\pm$ 0.000E+00$\approx$ & 0.000E+00 $\pm$ 0.000E+00 \\
    F03   & 0.000E+00 $\pm$ 0.000E+00$\approx$ & 0.000E+00 $\pm$ 0.000E+00 & 0.000E+00 $\pm$ 0.000E+00$\approx$ & 0.000E+00 $\pm$ 0.000E+00 \\\hline
    F04   & \textbf{3.628E-03 $\pm$ 7.089E-03}$\approx$ & 5.557E-03 $\pm$ 1.431E-02 & 5.201E+01 $\pm$ 3.374E+01$\approx$ & \textbf{4.239E+01 $\pm$ 4.248E+01} \\
    F05   & \textbf{2.010E+01 $\pm$ 9.499E-02}$+$ & 2.067E+01 $\pm$ 7.434E-02 & \textbf{2.001E+01 $\pm$ 1.948E-02}$+$ & 2.057E+01 $\pm$ 6.361E-02 \\
    F06   & 9.090E-01 $\pm$ 1.437E+00$\approx$ & \textbf{6.458E-01 $\pm$ 9.917E-01} & \textbf{2.080E+00 $\pm$ 1.352E+00}$\approx$ & 2.179E+00 $\pm$ 1.414E+00 \\
    F07   & \textbf{0.000E+00 $\pm$ 0.000E+00}$\approx$ & 1.450E-04 $\pm$ 1.036E-03 & 3.955E-03 $\pm$ 8.490E-03$\approx$ & \textbf{2.411E-03 $\pm$ 7.842E-03} \\
\hline
\end{tabular}
\end{table*}

\renewcommand\thetable{S.\arabic{table}}
\begin{table*}[!htbp]
\scriptsize
\centering
\caption{Experimental results of ETI-SHADE, ETI-OXDE and the related ETIgb-DEs for functions F01-F30 at $\emph{D}=30$.}\label{tablegb3}
\begin{tabular}{|c|cc|cc|}
\hline
\textbf{Function} & \textbf{ETI-SHADE} & \textbf{ETIgb-SHADE} & \textbf{ETI-OXDE} & \textbf{ETIgb-OXDE}  \\\hline
F01   & 4.551E+00 $\pm$ 1.395E+01$\approx$ & \textbf{2.256E+01 $\pm$ 1.072E+02} & 1.223E+05 $\pm$ 1.088E+05$\approx$ & \textbf{1.120E+05 $\pm$ 8.970E+04} \\
    F02   & 0.000E+00 $\pm$ 0.000E+00$\approx$ & 0.000E+00 $\pm$ 0.000E+00 & 0.000E+00 $\pm$ 0.000E+00$\approx$ & 0.000E+00 $\pm$ 0.000E+00 \\
    F03   & 0.000E+00 $\pm$ 0.000E+00$\approx$ & 0.000E+00 $\pm$ 0.000E+00 & 0.000E+00 $\pm$ 0.000E+00$\approx$ & 0.000E+00 $\pm$ 0.000E+00 \\\hline
    F04   & 0.000E+00 $\pm$ 0.000E+00$\approx$ & 0.000E+00 $\pm$ 0.000E+00 & 1.790E+00 $\pm$ 8.803E+00$\approx$ & \textbf{7.235E-01 $\pm$ 1.335E+00} \\
    F05   & \textbf{2.027E+01 $\pm$ 6.343E-02}$+$ & 2.061E+01 $\pm$ 1.911E-01 & \textbf{2.035E+01 $\pm$ 1.120E-01}$+$ & 2.090E+01 $\pm$ 4.970E-02 \\
    F06   & \textbf{9.006E-03 $\pm$ 6.431E-02}$\approx$ & 6.801E-02 $\pm$ 2.811E-01 & 4.959E-01 $\pm$ 7.787E-01$\approx$ & \textbf{4.181E-01 $\pm$ 7.437E-01} \\
    F07   & 0.000E+00 $\pm$ 0.000E+00$\approx$ & 0.000E+00 $\pm$ 0.000E+00 & 0.000E+00 $\pm$ 0.000E+00$\approx$ & 0.000E+00 $\pm$ 0.000E+00 \\
\hline
\end{tabular}
\end{table*}

\renewcommand\thetable{S.\arabic{table}}
\begin{table*}[!htbp]
\scriptsize
\centering
\caption{Different restart strategies in DE.}\label{restt}
\begin{tabular}{m{0pt}p{40pt}p{280pt}p{70pt}}
\hline
\rule{0pt}{8pt}& \textbf{Reference}     &\textbf{Restart Strategy} & \textbf{Parameter} \\\hline
\rule{0pt}{8pt}& [35]         & local restart in a bubble surrounding the best individual within the cluster $\textbf{x}_{\emph{best}}$, and global restart sampling outside the bubble  & $\emph{n}_{\emph{samples}}, \Delta$  \\
\rule{0pt}{8pt}& [52]         & restart by reinitializing certain individuals away from previous local optima   & $\delta, \delta_{\emph{vib}}, \delta_{\emph{fit}}, \lambda, \emph{interval}_{\emph{restart}}$\\
\rule{0pt}{8pt}& [53]         & restart by DE/rand/2 and by reinitializing certain individuals within the search range  & $\emph{cc}, \emph{M1}, \emph{M2}, \emph{S}, \emph{K}$ \\
\rule{0pt}{8pt}& [54]         & restart by increasing the population size   & $\emph{k}, \emph{c}$ \\
\rule{0pt}{8pt}& [55]         & restart according to Eqs. (4)-(9) in [55]   & \emph{UN}  \\
\rule{0pt}{8pt}& [56]         & restart according to Eqs. (15)-(20) in [56]   & $\emph{T}, \emph{a}$  \\
\rule{0pt}{8pt}& our method         &  restart according to Eqs. (18)-(21) in this paper  & $\emph{pr}, \emph{LN}, \emph{UN}$  \\\hline
\end{tabular}
\end{table*}

\renewcommand\thetable{S.\arabic{table}}
\begin{table*}[!htbp]
\scriptsize
\centering
\caption{Experimental results of ETI-DEs and AEPD-DEs for functions F01-F30 at $\emph{D}=30$.}\label{tablers1}
\begin{tabular}{|c|cc|cc|}
\hline
\textbf{Function} & \textbf{ETI-jDE ($\emph{NP}=20$)} & \textbf{AEPD-jDE ($\emph{NP}=20$)} & \textbf{ETI-JADE ($\emph{NP}=20$)} & \textbf{AEPD-JADE ($\emph{NP}=20$)}  \\\hline
F01   & 4.733E+05  $\pm$  4.165E+05$-$ & \textbf{1.632E+05 $\pm$ 1.329E+05} & 2.363E+04  $\pm$  1.094E+05$\approx$ & \textbf{4.800E+03  $\pm$  3.662E+03}  \\
    F02   & 2.537E-08  $\pm$  1.811E-07$-$ & \textbf{0.000E+00 $\pm$ 0.000E+00} & 0.000E+00 $\pm$ 0.000E+00$\approx$ & 0.000E+00 $\pm$ 0.000E+00  \\
    F03   & 0.000E+00 $\pm$ 0.000E+00$\approx$ & 0.000E+00 $\pm$ 0.000E+00 & 1.026E-02  $\pm$  7.328E-02$\approx$ & \textbf{4.078E-03  $\pm$  2.912E-02}  \\\hline
    F04   & \textbf{2.348E+01 $\pm$ 3.066E+01}$+$ & 5.970E+01  $\pm$  4.132E+01 & 4.691E+00  $\pm$  2.360E+01$-$ & \textbf{2.683E+00 $\pm$ 1.256E+01} \\
    F05   & \textbf{2.006E+01 $\pm$ 7.790E-02}$+$ & 2.032E+01  $\pm$  6.389E-02 & \textbf{2.001E+01 $\pm$ 3.049E-02}$+$ & 2.007E+01  $\pm$  3.318E-02 \\
    F06   & 8.232E+00  $\pm$  3.308E+00$-$ & \textbf{5.188E+00 $\pm$ 2.396E+00} & 5.516E+00  $\pm$  2.093E+00$-$ & \textbf{3.252E+00 $\pm$ 2.296E+00} \\
    F07   & 5.476E-02  $\pm$  1.109E-01$-$ & \textbf{1.670E-02 $\pm$ 2.070E-02} & 7.375E-03  $\pm$  1.217E-02$\approx$ & \textbf{5.594E-03  $\pm$  1.056E-02}  \\
\hline
\end{tabular}
\end{table*}

\renewcommand\thetable{S.\arabic{table}}
\begin{table*}[!htbp]
\scriptsize
\centering
\caption{Experimental results of ETI-DEs and AEPD-DEs for functions F01-F30 at $\emph{D}=30$.}\label{tablers2}
\begin{tabular}{|c|cc|cc|}
\hline
\textbf{Function} & \textbf{ETI-jDE ($\emph{NP}=100$)} & \textbf{AEPD-jDE ($\emph{NP}=20$)} & \textbf{ETI-JADE ($\emph{NP}=100$)} & \textbf{AEPD-JADE ($\emph{NP}=20$)}  \\\hline
F01   & \textbf{6.028E+04 $\pm$ 5.206E+04}$+$ & 1.632E+05 $\pm$ 1.329E+05 & \textbf{7.397E+02 $\pm$ 1.392E+03}$+$ & 4.800E+03 $\pm$ 3.662E+03 \\
    F02   & 0.000E+00 $\pm$ 0.000E+00$\approx$ & 0.000E+00 $\pm$ 0.000E+00 & 0.000E+00 $\pm$ 0.000E+00$\approx$ & 0.000E+00 $\pm$ 0.000E+00 \\
    F03   & 0.000E+00 $\pm$ 0.000E+00$\approx$ & 0.000E+00 $\pm$ 0.000E+00 & \textbf{6.664E-04 $\pm$ 1.563E-03}$+$ & 4.078E-03 $\pm$ 2.912E-02 \\\hline
    F04   & \textbf{1.510E+00 $\pm$ 8.925E-01}$+$ & 5.970E+01 $\pm$ 4.132E+01 & \textbf{0.000E+00 $\pm$ 0.000E+00}$+$ & 2.683E+00 $\pm$ 1.256E+01 \\
    F05   & \textbf{2.018E+01 $\pm$ 7.198E-02}$+$ & 2.032E+01 $\pm$ 6.389E-02 & \textbf{2.000E+01 $\pm$ 6.736E-03}$+$ & 2.007E+01 $\pm$ 3.318E-02 \\
    F06   & \textbf{1.196E+00 $\pm$ 1.339E+00}$+$ & 5.188E+00 $\pm$ 2.396E+00 & \textbf{5.091E-01 $\pm$ 6.621E-01}$+$ & 3.252E+00 $\pm$ 2.296E+00 \\
    F07   & \textbf{0.000E+00 $\pm$ 0.000E+00}$+$ & 1.670E-02 $\pm$ 2.070E-02 & \textbf{1.450E-04 $\pm$ 1.036E-03}$+$ & 5.594E-03 $\pm$ 1.056E-02 \\
\hline
\end{tabular}
\end{table*}

\renewcommand\thefigure{S.\arabic{table}}
\begin{table*}[!htbp]
\scriptsize
\centering
\caption{\emph{WIN}-\emph{LOSE} results of R1-ETI-DEs, R2-ETI-DEs, NoR-ETI-DEs (in Section IV.F) and ETI-DEs for functions F01-F30 at $\emph{D}=30$.}\label{tablera}
\begin{tabular}{|c|c|c|c|c|c|}
\hline
 \textbf{ETI-DEs} & \textbf{\emph{vs.} R1-ETI-DE/rand/1/bin} & \textbf{\emph{vs.} R1-ETI-DE/best/1/bin} & \textbf{\emph{vs.} R1-ETI-jDE} & \textbf{\emph{vs.} R1-ETI-JADE} & \textbf{\emph{vs.} R1-ETI-CoDE}  \\
 & 11-3 & 9-3 & 11-1 & 10-3 & 5-1 \\
 & \textbf{\emph{vs.} R1-ETI-SaDE} & \textbf{\emph{vs.} R1-ETI-ODE} & \textbf{\emph{vs.} R1-ETI-EPSDE} & \textbf{\emph{vs.} R1-ETI-SHADE} & \textbf{\emph{vs.} R1-ETI-OXDE}  \\
 & 9-1 & 3-0 & 6-5 & 9-5 & 11-0 \\\hline
 \textbf{ETI-DEs} & \textbf{\emph{vs.} R2-ETI-DE/rand/1/bin} & \textbf{\emph{vs.} R2-ETI-DE/best/1/bin} & \textbf{\emph{vs.} R2-ETI-jDE} & \textbf{\emph{vs.} R2-ETI-JADE} & \textbf{\emph{vs.} R2-ETI-CoDE}   \\
 & 8-8 & 4-2 & 9-0 & 13-1 & 11-1 \\
 & \textbf{\emph{vs.} R2-ETI-SaDE} & \textbf{\emph{vs.} R2-ETI-ODE} & \textbf{\emph{vs.} R2-ETI-EPSDE} & \textbf{\emph{vs.} R2-ETI-SHADE} & \textbf{\emph{vs.} R2-ETI-OXDE}  \\
 & 9-0 & 17-0 & 16-2 & 7-2 & 7-2 \\\hline
 \textbf{ETI-DEs} & \textbf{\emph{vs.} NoR-ETI-DE/rand/1/bin} & \textbf{\emph{vs.} NoR-ETI-DE/best/1/bin} & \textbf{\emph{vs.} NoR-ETI-jDE} & \textbf{\emph{vs.} NoR-ETI-JADE} & \textbf{\emph{vs.} NoR-ETI-CoDE}  \\
 & 13-3 & 7-2 & 13-0 & 12-1 & 10-0 \\
 & \textbf{\emph{vs.} NoR-ETI-SaDE} & \textbf{\emph{vs.} NoR-ETI-ODE} & \textbf{\emph{vs.} NoR-ETI-EPSDE} & \textbf{\emph{vs.} NoR-ETI-SHADE} & \textbf{\emph{vs.} NoR-ETI-OXDE}  \\
 & 9-4 & 12-2 & 13-1 & 12-1 & 14-0 \\\hline
\end{tabular}
\end{table*}

\clearpage

\renewcommand\thefigure{S.\arabic{table}}
\begin{table*}[!htbp]
\scriptsize
\centering
\caption{Experimental results of DE/rand/1/bin, DE/best/1/bin, jDE, JADE and the related ETI-based variants for functions F01-F30 at $\emph{D}=50$.}\label{table50d1}
\begin{tabular}{|c|cc|cc|}
\hline
\textbf{Function} & \textbf{DE/rand/1/bin} & \textbf{ETI-DE/rand/1/bin} & \textbf{DE/best/1/bin} & \textbf{ETI-DE/best/1/bin}  \\\hline
  F01   & 1.399E+06 $\pm$ 5.347E+05 & \textbf{8.417E+05 $\pm$ 2.873E+05}$+$ & 6.354E+07 $\pm$ 1.882E+07 & \textbf{1.226E+06 $\pm$ 4.545E+05}$+$ \\
    F02   & \textbf{1.913E+02 $\pm$ 7.531E+02} & 3.045E+03 $\pm$ 4.719E+03$-$ & \textbf{1.981E+02 $\pm$ 2.502E+02} & 6.963E+03 $\pm$ 7.455E+03$-$ \\
    F03   & \textbf{2.863E-01 $\pm$ 1.144E+00} & 3.588E-01 $\pm$ 7.479E-01$\approx$ & 8.240E+03 $\pm$ 2.083E+03 & \textbf{1.079E+02 $\pm$ 1.516E+02}$+$ \\\hline
    F04   & \textbf{7.729E+01 $\pm$ 2.730E+01} & 8.793E+01 $\pm$ 8.976E+00$\approx$ & 9.408E+01 $\pm$ 6.231E+00 & \textbf{5.922E+01 $\pm$ 3.859E+01}$+$ \\
    F05   & 2.113E+01 $\pm$ 3.349E-02 & \textbf{2.042E+01 $\pm$ 1.022E-01}$+$ & 2.112E+01 $\pm$ 3.629E-02 & \textbf{2.006E+01 $\pm$ 1.176E-01}$+$ \\
    F06   & \textbf{1.215E+00 $\pm$ 1.384E+00} & 1.786E+00 $\pm$ 1.671E+00$\approx$ & \textbf{4.841E+00 $\pm$ 2.272E+00} & 6.051E+00 $\pm$ 2.877E+00$-$ \\
    F07   & 1.450E-04 $\pm$ 1.036E-03 & 1.450E-04 $\pm$ 1.036E-03$+$ & 2.222E-03 $\pm$ 4.808E-03 & \textbf{1.352E-03 $\pm$ 4.520E-03}$\approx$ \\
\hline
\end{tabular}
\end{table*}

\renewcommand\thetable{S.\arabic{table}}
\begin{table*}[!htbp]
\scriptsize
\centering
\caption{Experimental results of CoDE, SaDE, ODE, EPSDE and the related ETI-based variants for functions F01-F30 at $\emph{D}=50$.}\label{table50d2}
\begin{tabular}{|c|cc|cc|}
\hline
\textbf{Function} & \textbf{CoDE} & \textbf{ETI-CoDE} & \textbf{SaDE} & \textbf{ETI-SaDE}  \\\hline
F01   & 4.059E+05 $\pm$ 1.669E+05 & \textbf{4.052E+05 $\pm$ 1.736E+05}$\approx$ & \textbf{7.958E+05 $\pm$ 2.986E+05} & 8.604E+05 $\pm$ 2.710E+05$\approx$ \\
    F02   & \textbf{1.190E-03 $\pm$ 2.515E-03} & 4.765E-02 $\pm$ 1.627E-01$-$ & \textbf{3.436E+03 $\pm$ 3.353E+03} & 4.227E+03 $\pm$ 3.214E+03$\approx$ \\
    F03   & \textbf{9.130E-02 $\pm$ 1.776E-01} & 5.349E-01 $\pm$ 1.479E+00$-$ & \textbf{7.512E+02 $\pm$ 6.082E+02} & 1.328E+03 $\pm$ 1.112E+03$-$ \\\hline
    F04   & 3.357E+01 $\pm$ 3.812E+01 & \textbf{3.066E+01 $\pm$ 3.596E+01}$\approx$ & \textbf{8.122E+01 $\pm$ 4.663E+01} & 9.818E+01 $\pm$ 3.930E+01$-$ \\
    F05   & 2.079E+01 $\pm$ 6.439E-02 & \textbf{2.006E+01 $\pm$ 5.966E-02}$+$ & 2.077E+01 $\pm$ 4.449E-02 & \textbf{2.001E+01 $\pm$ 1.835E-02}$+$ \\
    F06   & 4.088E+00 $\pm$ 2.748E+00 & \textbf{3.784E+00 $\pm$ 2.429E+00}$\approx$ & 1.182E+01 $\pm$ 2.452E+00 & \textbf{1.155E+01 $\pm$ 2.673E+00}$\approx$ \\
    F07   & \textbf{1.450E-04 $\pm$ 1.036E-03} & 3.378E-04 $\pm$ 2.412E-03$-$ & \textbf{1.000E-02 $\pm$ 8.789E-03} & 1.004E-02 $\pm$ 1.193E-02$\approx$ \\
\hline
\end{tabular}
\end{table*}

\renewcommand\thetable{S.\arabic{table}}
\begin{table*}[!htbp]
\scriptsize
\centering
\caption{Experimental results of SHADE, OXDE and the related ETI-based variants for functions F01-F30 at $\emph{D}=50$.}\label{table50d3}
\begin{tabular}{|c|cc|cc|}
\hline
\textbf{Function} & \textbf{SHADE} & \textbf{ETI-SHADE} & \textbf{OXDE} & \textbf{ETI-OXDE}  \\\hline
F01   & \textbf{2.511E+04 $\pm$ 1.246E+04} & 5.168E+04 $\pm$ 2.653E+04$-$ & 1.047E+06 $\pm$ 4.357E+05 & \textbf{8.382E+05 $\pm$ 3.255E+05}$+$ \\
    F02   & 0.000E+00 $\pm$ 0.000E+00 & 0.000E+00 $\pm$ 0.000E+00$\approx$ & \textbf{6.528E+02 $\pm$ 1.547E+03} & 4.519E+03 $\pm$ 4.998E+03$-$ \\
    F03   & 0.000E+00 $\pm$ 0.000E+00 & 0.000E+00 $\pm$ 0.000E+00$\approx$ & \textbf{9.332E-02 $\pm$ 3.010E-01} & 2.155E-01 $\pm$ 5.330E-01$\approx$ \\\hline
    F04   & \textbf{1.209E+01 $\pm$ 3.176E+01} & 1.834E+01 $\pm$ 3.729E+01$\approx$ & 8.687E+01 $\pm$ 1.146E+01 & \textbf{8.564E+01 $\pm$ 1.452E+01}$\approx$ \\
    F05   & \textbf{2.027E+01 $\pm$ 2.287E-02} & 2.030E+01 $\pm$ 9.085E-02$-$ & 2.113E+01 $\pm$ 3.902E-02 & \textbf{2.038E+01 $\pm$ 1.104E-01}$+$ \\
    F06   & 1.029E+00 $\pm$ 8.136E-01 & \textbf{8.501E-01 $\pm$ 8.322E-01}$\approx$ & \textbf{1.110E+00 $\pm$ 1.314E+00} & 1.305E+00 $\pm$ 1.351E+00$\approx$ \\
    F07   & 1.450E-04 $\pm$ 1.036E-03 & 1.450E-04 $\pm$ 1.036E-03$+$ & 1.014E-03 $\pm$ 3.311E-03 & \textbf{4.833E-04 $\pm$ 1.973E-03}$+$ \\
\hline
\end{tabular}
\end{table*}

\renewcommand\thefigure{S.\arabic{table}}
\begin{table*}[!htbp]
\scriptsize
\centering
\caption{Experimental results of DE/rand/1/bin, DE/best/1/bin, jDE, JADE and the related ETI-based variants for functions F01-F30 at $\emph{D}=100$.}\label{table100d1}
\begin{tabular}{|c|cc|cc|}
\hline
\textbf{Function} & \textbf{DE/rand/1/bin} & \textbf{ETI-DE/rand/1/bin} & \textbf{DE/best/1/bin} & \textbf{ETI-DE/best/1/bin}  \\\hline
  F01   & 4.492E+06 $\pm$ 1.385E+06 & \textbf{3.839E+06 $\pm$ 1.121E+06}$+$ & 4.596E+08 $\pm$ 1.482E+08 & \textbf{4.562E+06 $\pm$ 1.549E+06}$+$ \\
    F02   & 2.156E+04 $\pm$ 2.371E+04 & \textbf{1.706E+04 $\pm$ 1.555E+04}$\approx$ & \textbf{5.527E+02 $\pm$ 7.737E+02} & 3.786E+04 $\pm$ 4.325E+04$-$ \\
    F03   & \textbf{1.399E+03 $\pm$ 1.081E+03} & 1.571E+03 $\pm$ 1.199E+03$\approx$ & 6.372E+04 $\pm$ 9.975E+03 & \textbf{9.564E+02 $\pm$ 1.045E+03}$+$ \\\hline
    F04   & \textbf{1.781E+02 $\pm$ 3.805E+01} & 1.795E+02 $\pm$ 3.485E+01$\approx$ & 1.885E+02 $\pm$ 2.923E+01 & \textbf{1.749E+02 $\pm$ 3.049E+01}$+$ \\
    F05   & 2.131E+01 $\pm$ 2.829E-02 & \textbf{2.046E+01 $\pm$ 1.129E-01}$+$ & 2.132E+01 $\pm$ 2.003E-02 & \textbf{2.002E+01 $\pm$ 2.780E-02}$+$ \\
    F06   & \textbf{1.186E+01 $\pm$ 3.390E+00} & 1.426E+01 $\pm$ 3.402E+00$-$ & \textbf{1.784E+01 $\pm$ 4.952E+00} & 2.497E+01 $\pm$ 4.701E+00$-$ \\
    F07   & 4.833E-04 $\pm$ 1.973E-03 & \textbf{4.351E-04 $\pm$ 1.758E-03}$+$ & 4.351E-04 $\pm$ 1.758E-03 & \textbf{2.900E-04 $\pm$ 1.450E-03}$+$ \\
\hline
\end{tabular}
\end{table*}

\renewcommand\thetable{S.\arabic{table}}
\begin{table*}[!htbp]
\scriptsize
\centering
\caption{Experimental results of CoDE, SaDE, ODE, EPSDE and the related ETI-based variants for functions F01-F30 at $\emph{D}=100$.}\label{table100d2}
\begin{tabular}{|c|cc|cc|}
\hline
\textbf{Function} & \textbf{CoDE} & \textbf{ETI-CoDE} & \textbf{SaDE} & \textbf{ETI-SaDE}  \\\hline
  F01   & \textbf{1.060E+06 $\pm$ 2.786E+05} & 1.197E+06 $\pm$ 3.761E+05$\approx$ & \textbf{4.668E+06 $\pm$ 8.056E+05} & 5.645E+06 $\pm$ 1.094E+06$-$ \\
    F02   & \textbf{7.691E-01 $\pm$ 2.231E+00} & 5.818E+01 $\pm$ 1.459E+02$-$ & \textbf{1.182E+04 $\pm$ 6.929E+03} & 1.277E+04 $\pm$ 6.811E+03$\approx$ \\
    F03   & \textbf{6.164E+02 $\pm$ 7.210E+02} & 6.626E+02 $\pm$ 6.408E+02$\approx$ & \textbf{4.495E+03 $\pm$ 2.815E+03} & 6.066E+03 $\pm$ 2.713E+03$-$ \\\hline
    F04   & \textbf{1.483E+02 $\pm$ 3.419E+01} & 1.588E+02 $\pm$ 3.651E+01$\approx$ & 2.453E+02 $\pm$ 5.654E+01 & \textbf{2.331E+02 $\pm$ 5.522E+01}$\approx$ \\
    F05   & 2.066E+01 $\pm$ 3.934E-01 & \textbf{2.004E+01 $\pm$ 4.420E-02}$+$ & 2.105E+01 $\pm$ 2.378E-02 & \textbf{2.000E+01 $\pm$ 1.347E-02}$+$ \\
    F06   & 2.909E+01 $\pm$ 5.433E+00 & \textbf{2.769E+01 $\pm$ 5.737E+00}$\approx$ & \textbf{6.068E+01 $\pm$ 4.398E+00} & 6.069E+01 $\pm$ 3.848E+00$\approx$ \\
    F07   & 1.450E-04 $\pm$ 1.036E-03 & \textbf{0.000E+00 $\pm$ 0.000E+00}$+$ & 4.245E-03 $\pm$ 8.185E-03 & \textbf{2.026E-03 $\pm$ 5.645E-03}$+$ \\
\hline
\end{tabular}
\end{table*}

\renewcommand\thetable{S.\arabic{table}}
\begin{table*}[!htbp]
\scriptsize
\centering
\caption{Experimental results of SHADE, OXDE and the related ETI-based variants for functions F01-F30 at $\emph{D}=100$.}\label{table100d3}
\begin{tabular}{|c|cc|cc|}
\hline
\textbf{Function} & \textbf{SHADE} & \textbf{ETI-SHADE} & \textbf{OXDE} & \textbf{ETI-OXDE}  \\\hline
F01   & \textbf{1.162E+05 $\pm$ 5.084E+04} & 1.442E+05 $\pm$ 6.459E+04$-$ & \textbf{2.280E+06 $\pm$ 5.801E+05} & 2.588E+06 $\pm$ 6.312E+05$-$ \\
    F02   & 0.000E+00 $\pm$ 0.000E+00 & 0.000E+00 $\pm$ 0.000E+00$\approx$ & 1.104E+04 $\pm$ 1.402E+04 & \textbf{1.056E+04 $\pm$ 1.126E+04}$\approx$ \\
    F03   & 5.657E-04 $\pm$ 1.287E-03 & \textbf{1.532E-05 $\pm$ 2.951E-05}$+$ & \textbf{9.700E+02 $\pm$ 7.702E+02} & 1.438E+03 $\pm$ 1.488E+03$\approx$ \\\hline
    F04   & \textbf{6.576E+01 $\pm$ 5.191E+01} & 7.304E+01 $\pm$ 4.874E+01$\approx$ & 1.796E+02 $\pm$ 3.941E+01 & \textbf{1.739E+02 $\pm$ 4.484E+01}$\approx$ \\
    F05   & 2.035E+01 $\pm$ 1.606E-02 & \textbf{2.008E+01 $\pm$ 9.485E-02}$+$ & 2.132E+01 $\pm$ 2.440E-02 & \textbf{2.038E+01 $\pm$ 1.537E-01}$+$ \\
    F06   & 2.354E+01 $\pm$ 4.314E+00 & \textbf{2.143E+01 $\pm$ 3.242E+00}$+$ & \textbf{1.817E+01 $\pm$ 3.616E+00} & 1.888E+01 $\pm$ 3.213E+00$\approx$ \\
    F07   & \textbf{2.021E-03 $\pm$ 7.745E-03} & 2.026E-03 $\pm$ 5.833E-03$-$ & 1.834E-03 $\pm$ 5.939E-03 & \textbf{1.111E-03 $\pm$ 3.663E-03}$+$ \\
\hline
\end{tabular}
\end{table*}

\renewcommand\thetable{S.\arabic{table}}
\begin{table*}[!htbp]
\scriptsize
\centering
\caption{Holm test on the fitness, reference algorithm = ETI-SHADE (rank=15.70) for functions F01-F30 at $\emph{D}=50$.}\label{hbtest50d}
\begin{tabular}{|c|c|c|c|c|c|c|}
\hline
\textbf{\emph{j}} & \textbf{Optimizer} & \textbf{Rank} & \textbf{$\emph{z}_\emph{j}$} & \textbf{$\emph{p}_\emph{j}$} & \textbf{$\bm\delta/\emph{j}$} & \textbf{Hypothesis} \\\hline
    1     & SHADE & 13.87 & -1.20E+00 & 1.15E-01 & 2.94E-03 & Accepted \\\hline
    2     & ETI-JADE & 13.67 & -1.33E+00 & 9.16E-02 & 6.25E-03 & Accepted \\\hline
    3     & ETI-OXDE & 13.50 & -1.44E+00 & 7.49E-02 & 2.63E-03 & Accepted \\\hline
    4     & ETI-DE/rand/1/bin & 13.47 & -1.46E+00 & 7.19E-02 & 2.50E-02 & Accepted \\\hline
    5     & ETI-CoDE & 13.47 & -1.46E+00 & 7.19E-02 & 5.00E-03 & Accepted \\\hline
    6     & ETI-jDE & 12.47 & -2.12E+00 & 1.71E-02 & 8.33E-03 & Accepted \\\hline
    7     & ETI-EPSDE & 11.70 & -2.62E+00 & 4.41E-03 & 3.13E-03 & Accepted \\\hline
    8     & CoDE  & 10.93 & -3.12E+00 & 9.03E-04 & 5.56E-03 & Rejected \\\hline
    9     & JADE  & 10.67 & -3.30E+00 & 4.92E-04 & 7.14E-03 & Rejected \\\hline
    10    & ETI-ODE & 10.63 & -3.32E+00 & 4.55E-04 & 3.57E-03 & Rejected \\\hline
    11    & jDE   & 10.20 & -3.60E+00 & 1.59E-04 & 1.00E-02 & Rejected \\\hline
    12    & DE/rand/1/bin & 8.93  & -4.43E+00 & 4.72E-06 & 5.00E-02 & Rejected \\\hline
    13    & OXDE  & 8.63  & -4.63E+00 & 1.86E-06 & 2.78E-03 & Rejected \\\hline
    14    & EPSDE & 8.40  & -4.78E+00 & 8.81E-07 & 3.33E-03 & Rejected \\\hline
    15    & ETI-SaDE & 8.33  & -4.82E+00 & 7.08E-07 & 4.17E-03 & Rejected \\\hline
    16    & ODE   & 8.33  & -4.82E+00 & 7.08E-07 & 3.85E-03 & Rejected \\\hline
    17    & SaDE  & 6.93  & -5.74E+00 & 4.76E-09 & 4.55E-03 & Rejected \\\hline
    18    & ETI-DE/best/1/bin & 6.63  & -5.94E+00 & 1.46E-09 & 1.25E-02 & Rejected \\\hline
    19    & DE/best/1/bin & 3.53  & -7.96E+00 & 8.33E-16 & 1.67E-02 & Rejected \\\hline
\end{tabular}
\end{table*}

\renewcommand\thetable{S.\arabic{table}}
\begin{table*}[!htbp]
\scriptsize
\centering
\caption{Holm test on the fitness, reference algorithm = ETI-SHADE (rank=15.50) for functions F01-F30 at $\emph{D}=100$.}\label{hbtest100d}
\begin{tabular}{|c|c|c|c|c|c|c|}
\hline
\textbf{\emph{j}} & \textbf{Optimizer} & \textbf{Rank} & \textbf{$\emph{z}_\emph{j}$} & \textbf{$\emph{p}_\emph{j}$} & \textbf{$\bm\delta/\emph{j}$} & \textbf{Hypothesis} \\\hline
    1     & SHADE & 14.50 & -6.55E-01 & 2.56E-01 & 2.94E-03 & Accepted \\\hline
    2     & ETI-CoDE & 13.97 & -1.00E+00 & 1.58E-01 & 5.00E-03 & Accepted \\\hline
    3     & ETI-JADE & 13.23 & -1.48E+00 & 6.89E-02 & 6.25E-03 & Accepted \\\hline
    4     & ETI-jDE & 13.00 & -1.64E+00 & 5.09E-02 & 8.33E-03 & Accepted \\\hline
    5     & CoDE  & 12.57 & -1.92E+00 & 2.74E-02 & 5.56E-03 & Accepted \\\hline
    6     & ETI-OXDE & 12.30 & -2.09E+00 & 1.81E-02 & 2.63E-03 & Accepted \\\hline
    7     & ETI-DE/rand/1/bin & 11.33 & -2.73E+00 & 3.19E-03 & 2.50E-02 & Rejected \\\hline
    8     & jDE   & 11.33 & -2.73E+00 & 3.19E-03 & 1.00E-02 & Rejected \\\hline
    9     & JADE  & 10.93 & -2.99E+00 & 1.40E-03 & 7.14E-03 & Rejected \\\hline
    10    & ETI-EPSDE & 10.87 & -3.03E+00 & 1.21E-03 & 3.13E-03 & Rejected \\\hline
    11    & ETI-ODE & 9.63  & -3.84E+00 & 6.14E-05 & 3.57E-03 & Rejected \\\hline
    12    & ETI-SaDE & 9.23  & -4.10E+00 & 2.04E-05 & 4.17E-03 & Rejected \\\hline
    13    & EPSDE & 8.63  & -4.50E+00 & 3.47E-06 & 3.33E-03 & Rejected \\\hline
    14    & OXDE  & 8.20  & -4.78E+00 & 8.81E-07 & 2.78E-03 & Rejected \\\hline
    15    & DE/rand/1/bin & 8.03  & -4.89E+00 & 5.09E-07 & 5.00E-02 & Rejected \\\hline
    16    & ETI-DE/best/1/bin & 7.60  & -5.17E+00 & 1.16E-07 & 1.25E-02 & Rejected \\\hline
    17    & SaDE  & 7.60  & -5.17E+00 & 1.16E-07 & 4.55E-03 & Rejected \\\hline
    18    & ODE   & 7.27  & -5.39E+00 & 3.52E-08 & 3.85E-03 & Rejected \\\hline
    19    & DE/best/1/bin & 4.27  & -7.35E+00 & 9.62E-14 & 1.67E-02 & Rejected \\\hline
\end{tabular}
\end{table*}

\clearpage

\renewcommand\thefigure{S.\arabic{figure}}
\begin{figure*}[!htbp]
\begin{minipage}[t]{1\linewidth}
\centering
\includegraphics[width=12cm]{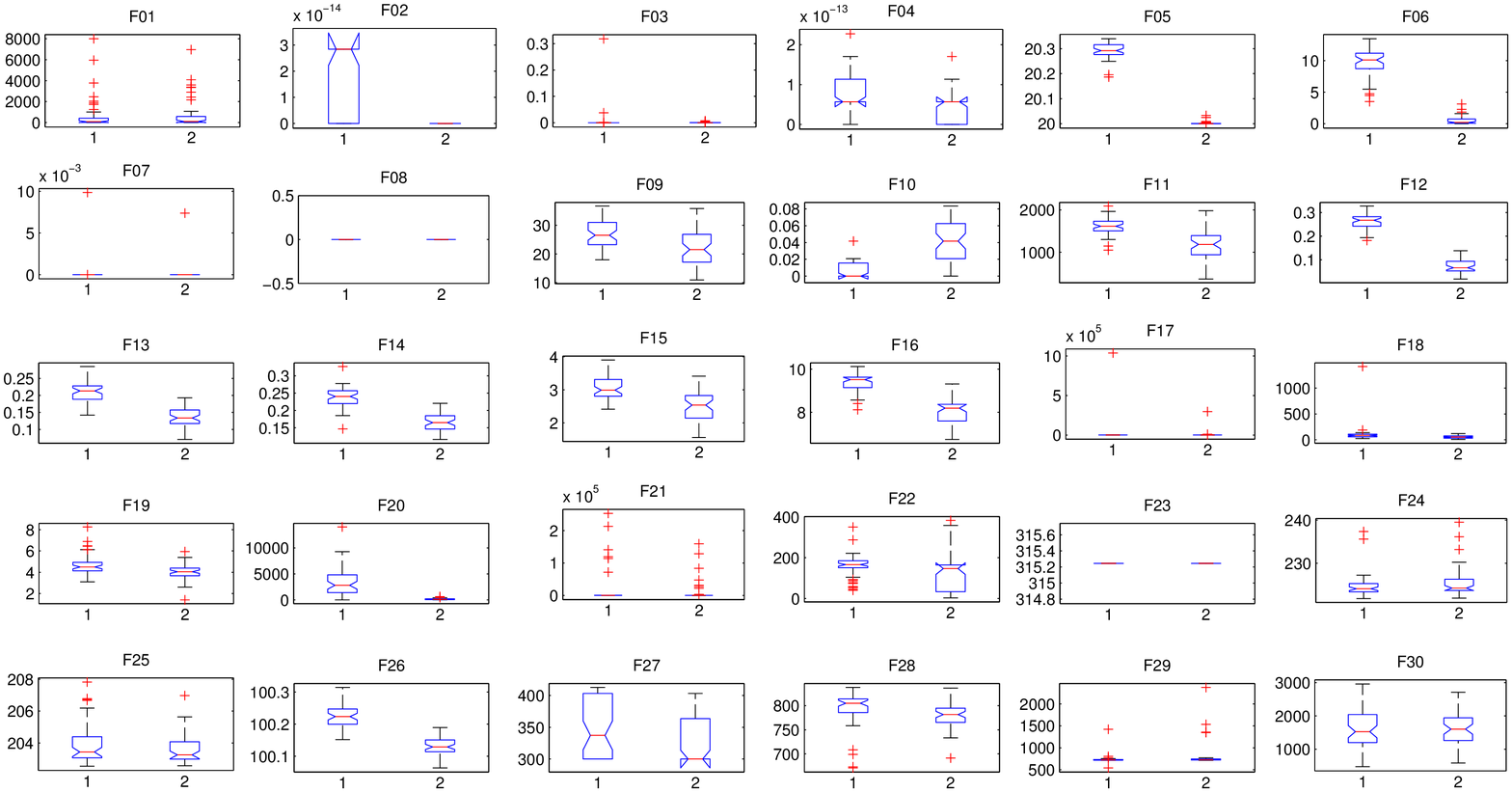}
\caption{Box plots for the results of JADE with/without ETI on CEC 2014 test suite at $\emph{D}=30$: 1--JADE; 2--ETI-JADE.} \label{jadebp30}
\end{minipage}
\end{figure*}

\renewcommand\thefigure{S.\arabic{figure}}
\begin{figure*}[!htbp]
\begin{minipage}[t]{1\linewidth}
\centering
\includegraphics[width=12cm]{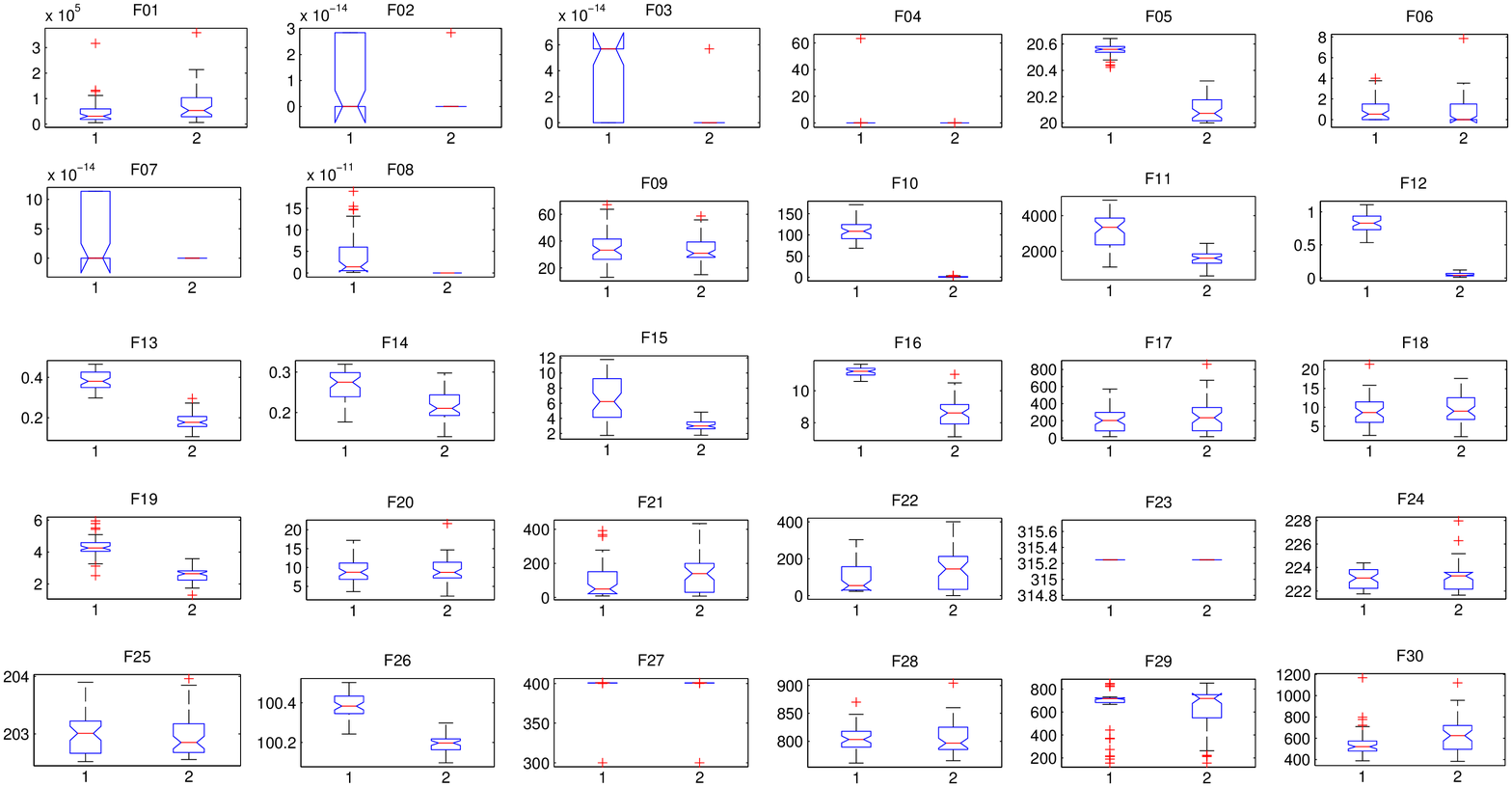}
\caption{Box plots for the results of CoDE with/without ETI on CEC 2014 test suite at $\emph{D}=30$: 1--CoDE; 2--ETI-CoDE.} \label{codebp30}
\end{minipage}
\end{figure*}

\renewcommand\thefigure{S.\arabic{figure}}
\begin{figure*}[!htbp]
\begin{minipage}[t]{1\linewidth}
\centering
\includegraphics[width=12cm]{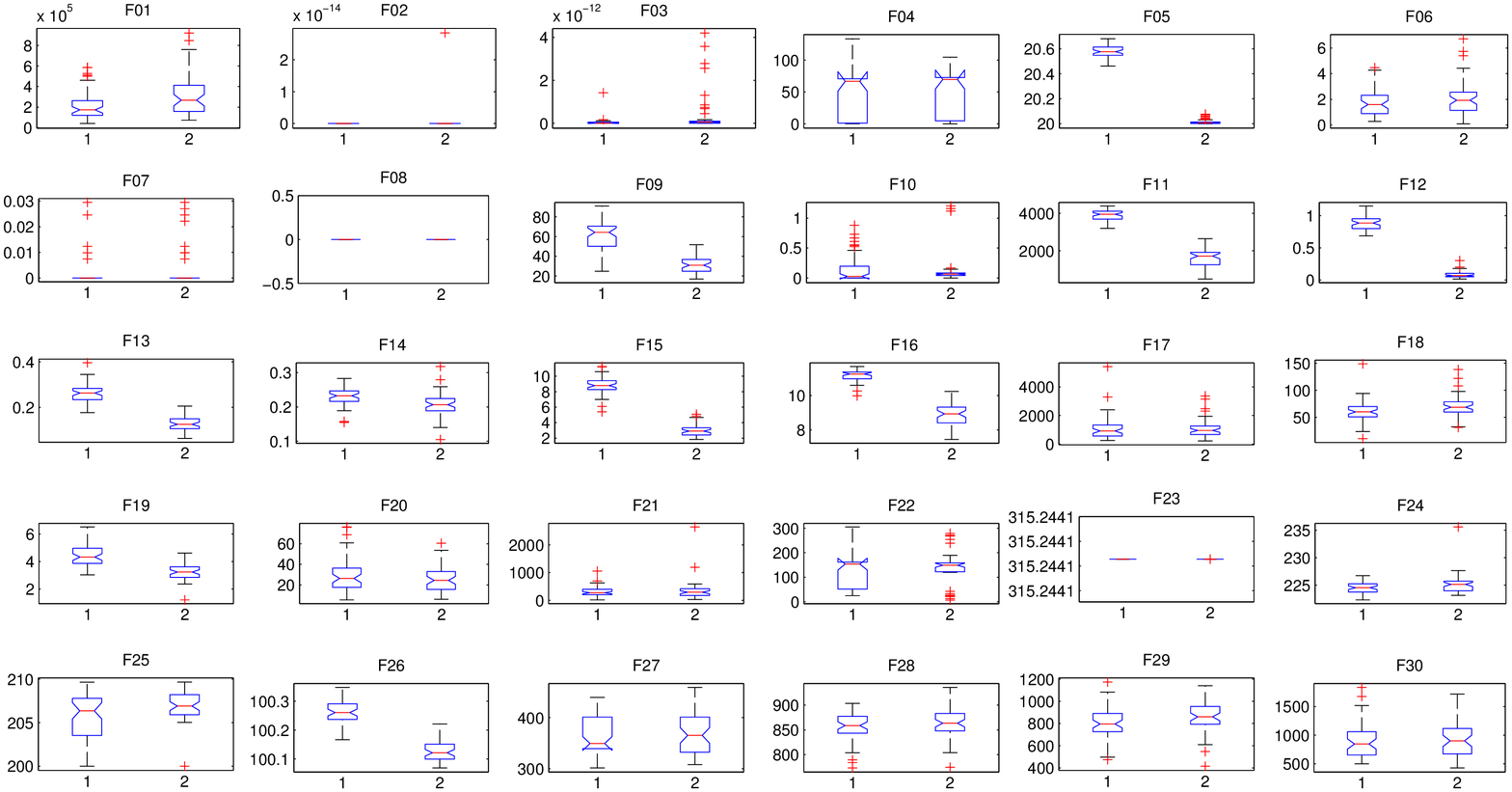}
\caption{Box plots for the results of SaDE with/without ETI on CEC 2014 test suite at $\emph{D}=30$: 1--SaDE; 2--ETI-SaDE.} \label{sadebp30}
\end{minipage}
\end{figure*}

\renewcommand\thefigure{S.\arabic{figure}}
\begin{figure*}[!htbp]
\begin{minipage}[t]{1\linewidth}
\centering
\includegraphics[width=12cm]{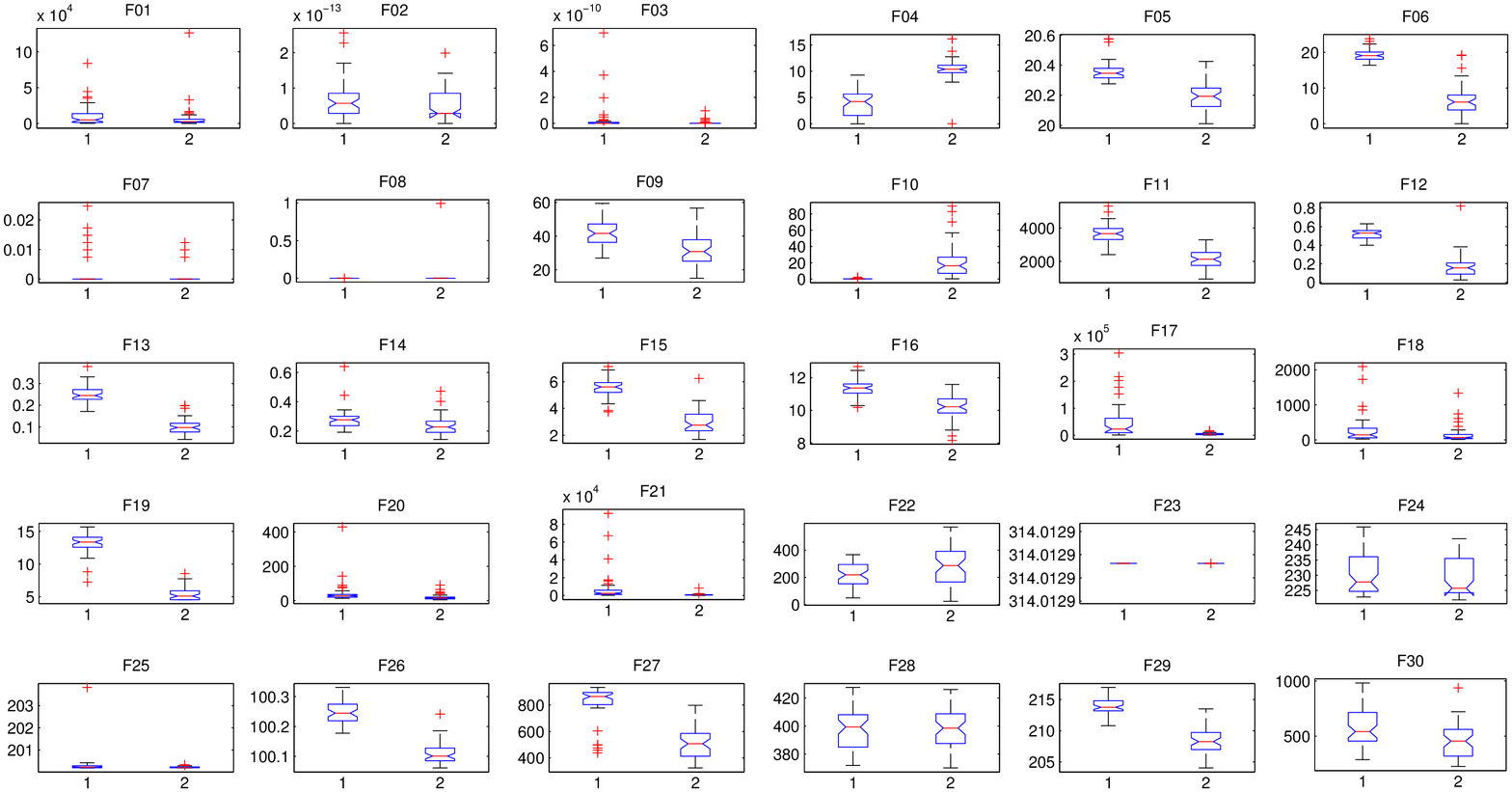}
\caption{Box plots for the results of EPSDE with/without ETI on CEC 2014 test suite at $\emph{D}=30$: 1--EPSDE; 2--ETI-EPSDE.} \label{epsdebp30}
\end{minipage}
\end{figure*}

\renewcommand\thefigure{S.\arabic{figure}}
\begin{figure*}[!htbp]
\centering
\subfigure[]{
\begin{minipage}[t]{0.3\textwidth}
\label{fig:subfig:a} 
\includegraphics[width=2.1in]{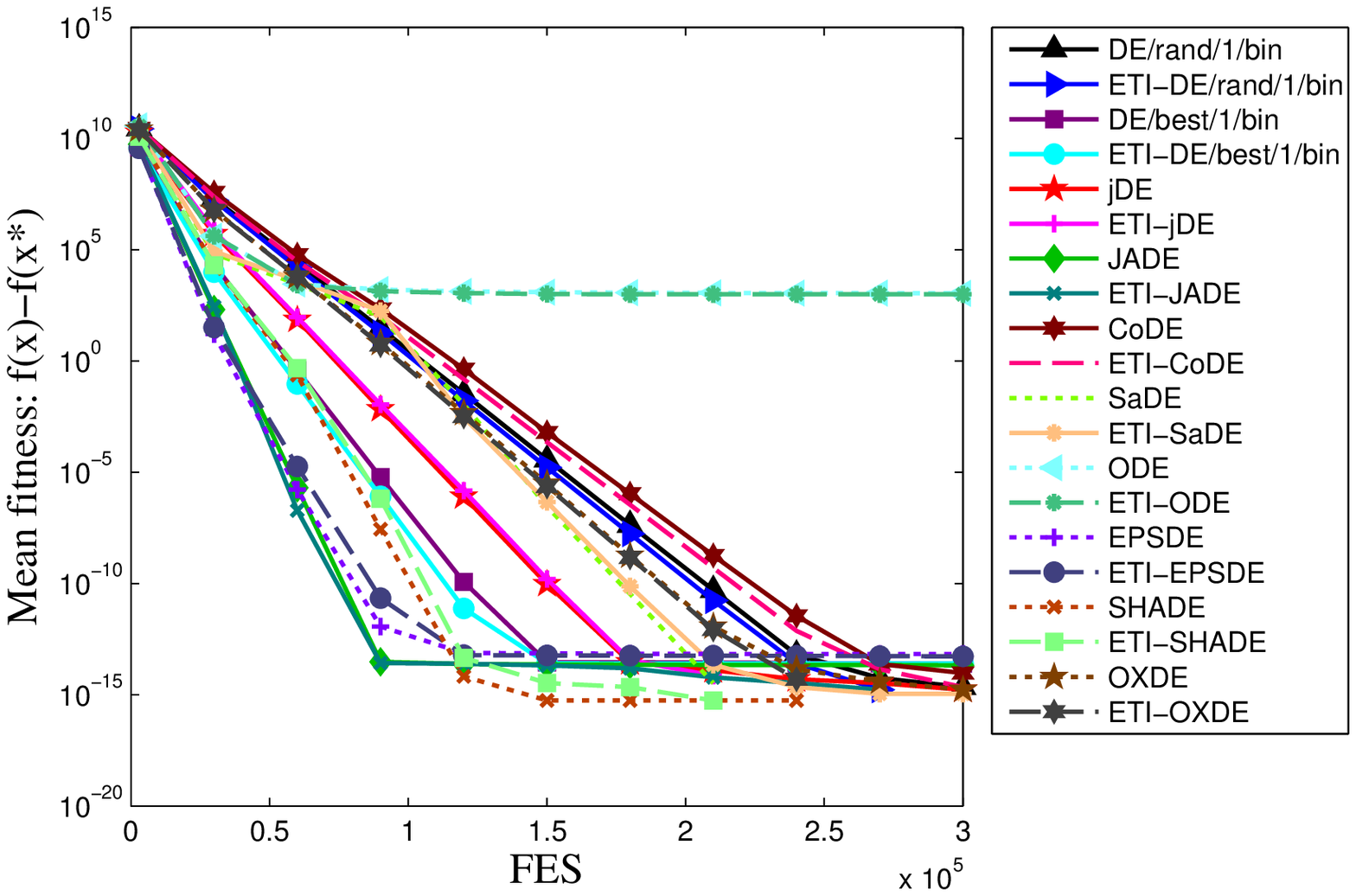}
\end{minipage}}
\centering
\subfigure[]{
\begin{minipage}[t]{0.3\textwidth}
\label{fig:subfig:b} 
\includegraphics[width=2.1in]{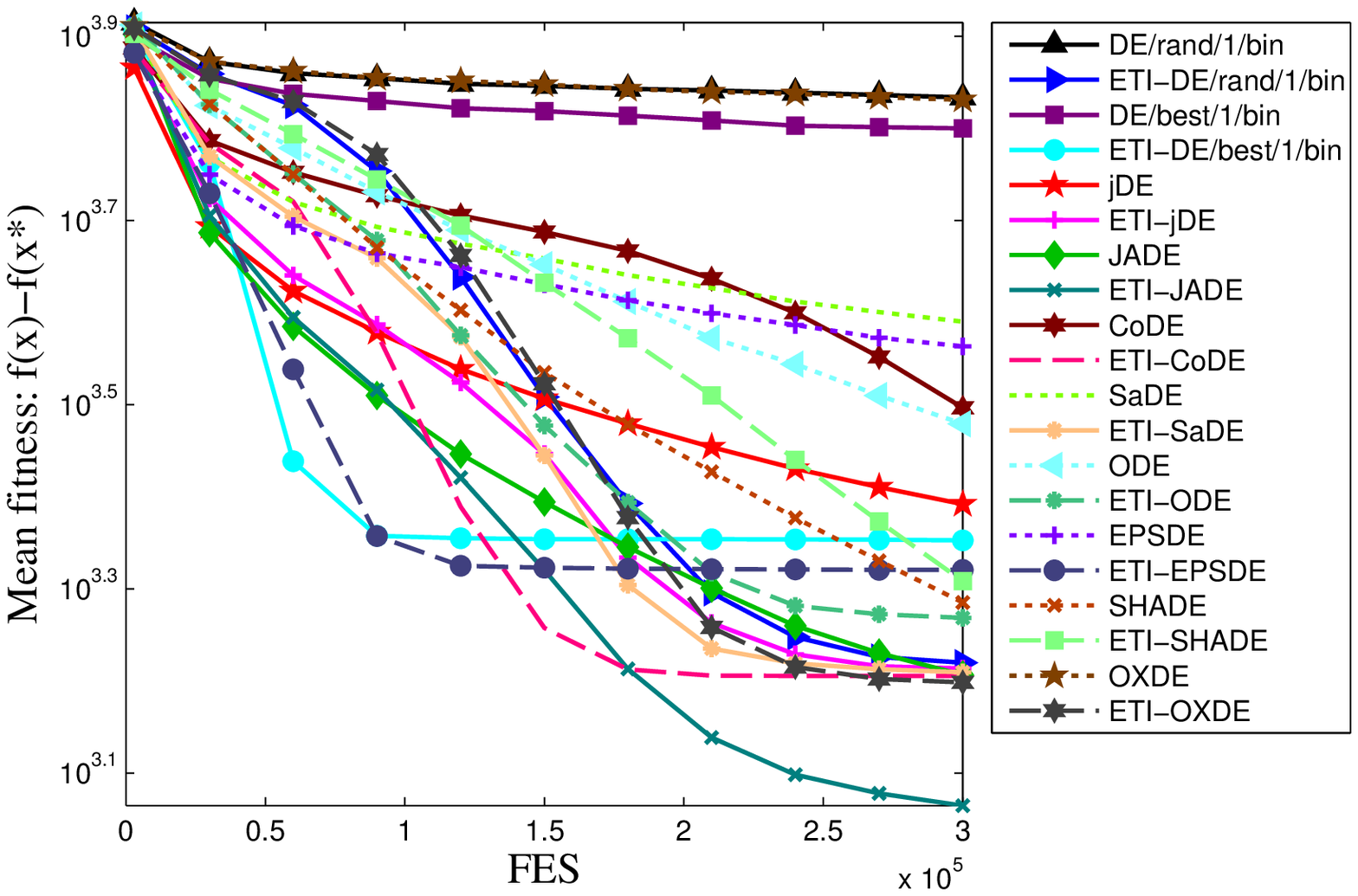}
\end{minipage}}
\centering
\subfigure[]{
\begin{minipage}[t]{0.3\textwidth}
\label{fig:subfig:c} 
\includegraphics[width=2.1in]{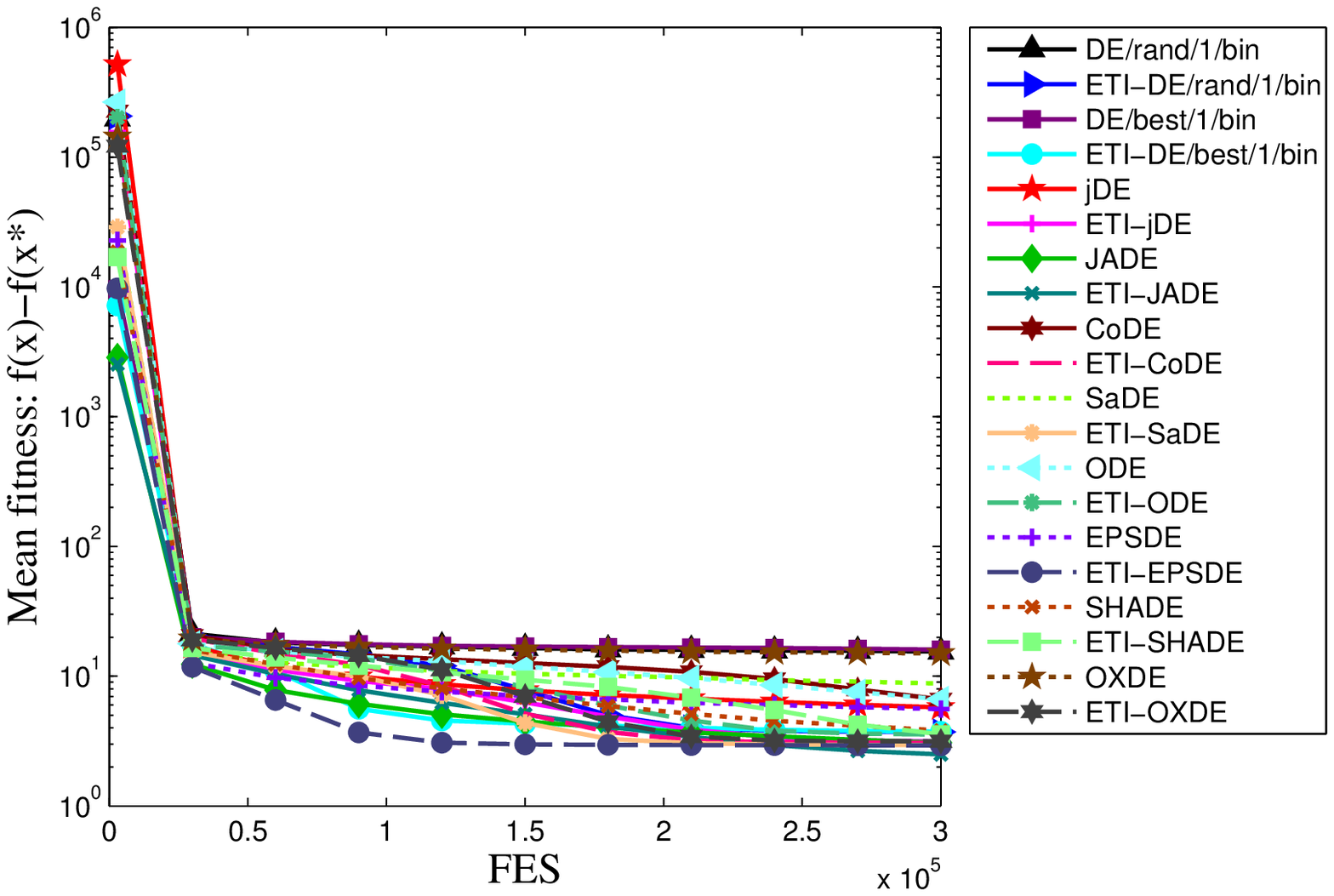}
\end{minipage}}
\centering
\subfigure[]{
\begin{minipage}[t]{0.3\textwidth}
\label{fig:subfig:d} 
\includegraphics[width=2.1in]{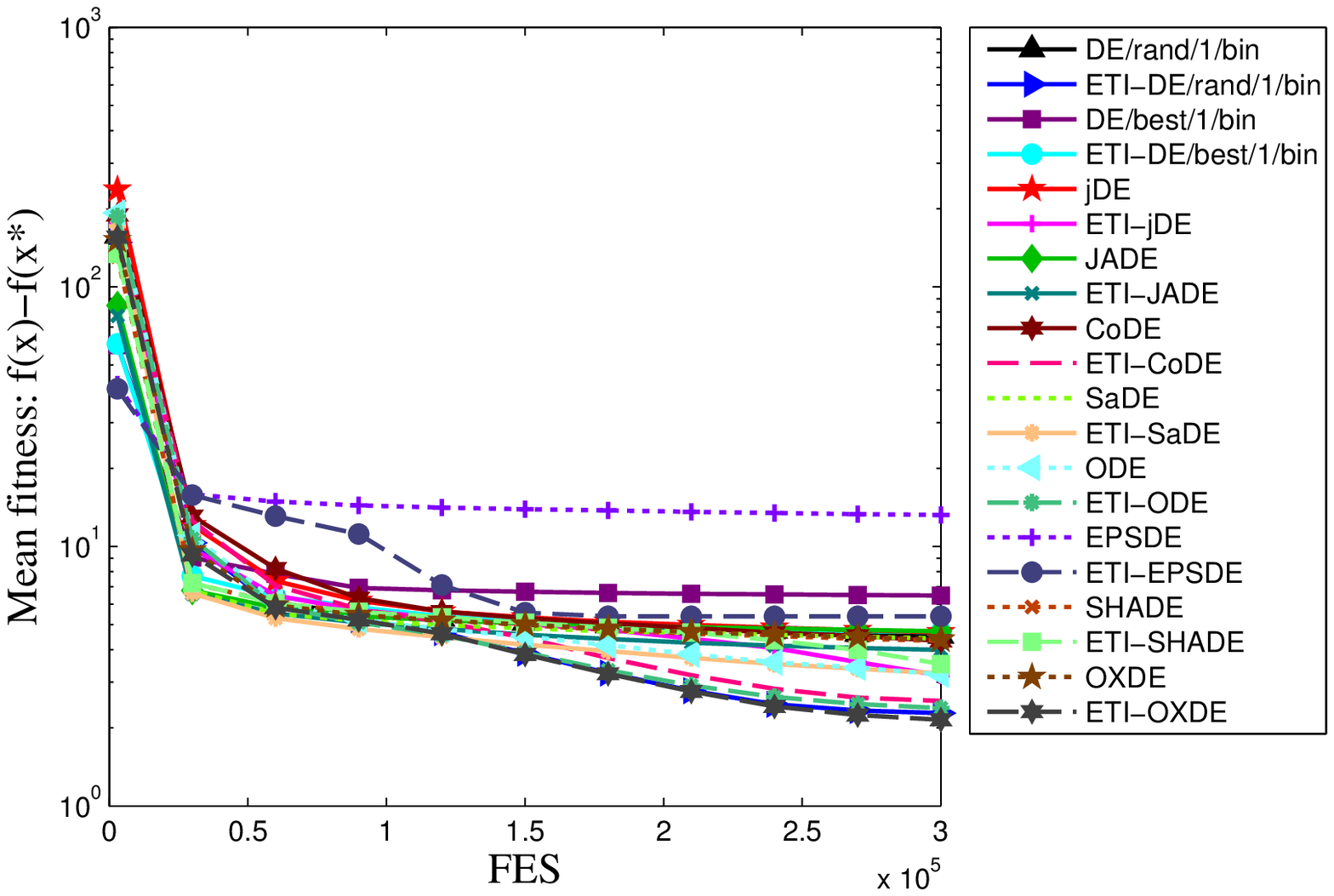}
\end{minipage}}
\centering
\subfigure[]{
\begin{minipage}[t]{0.3\textwidth}
\label{fig:subfig:e} 
\includegraphics[width=2.1in]{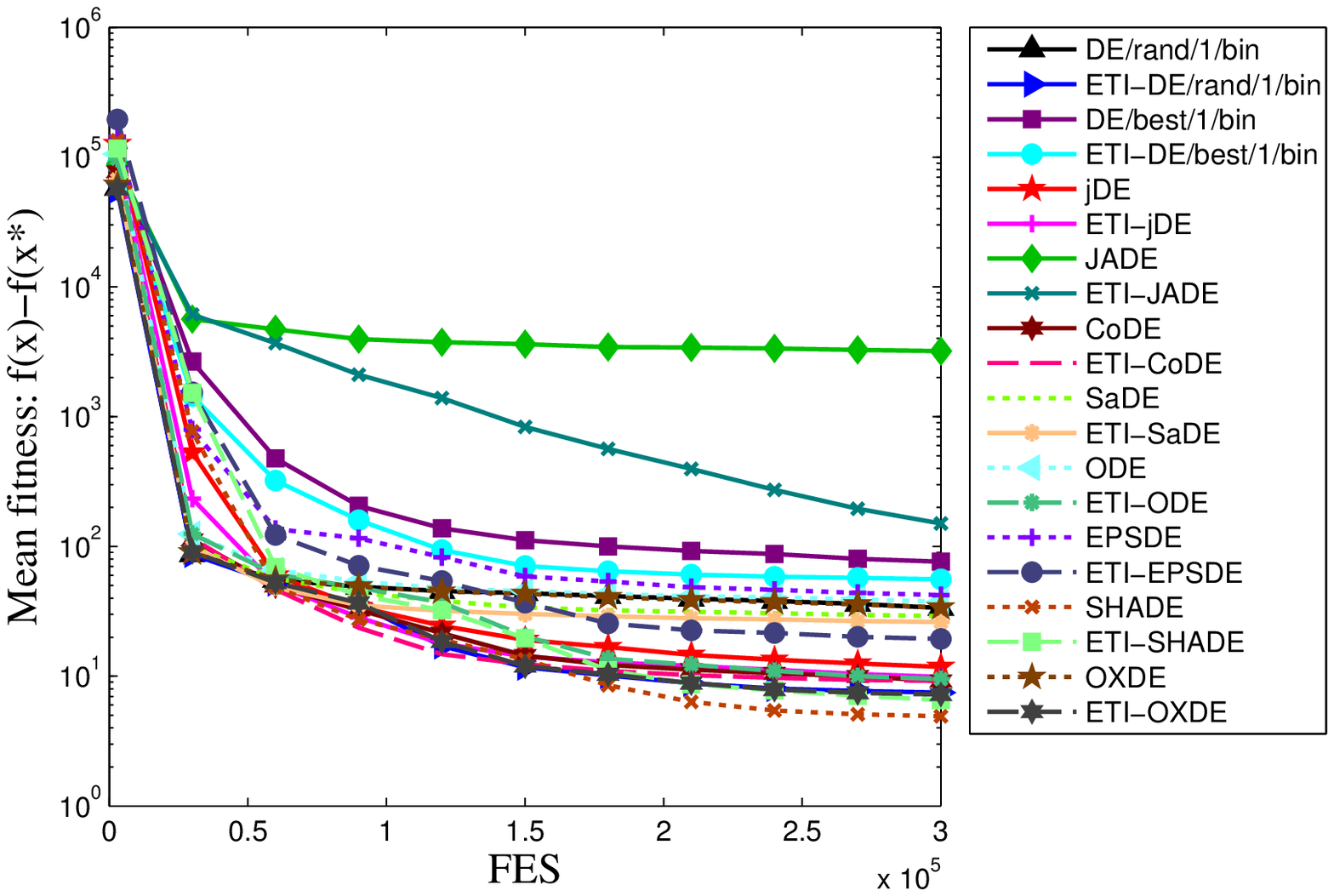}
\end{minipage}}
\centering
\subfigure[]{
\begin{minipage}[t]{0.3\textwidth}
\label{fig:subfig:e} 
\includegraphics[width=2.1in]{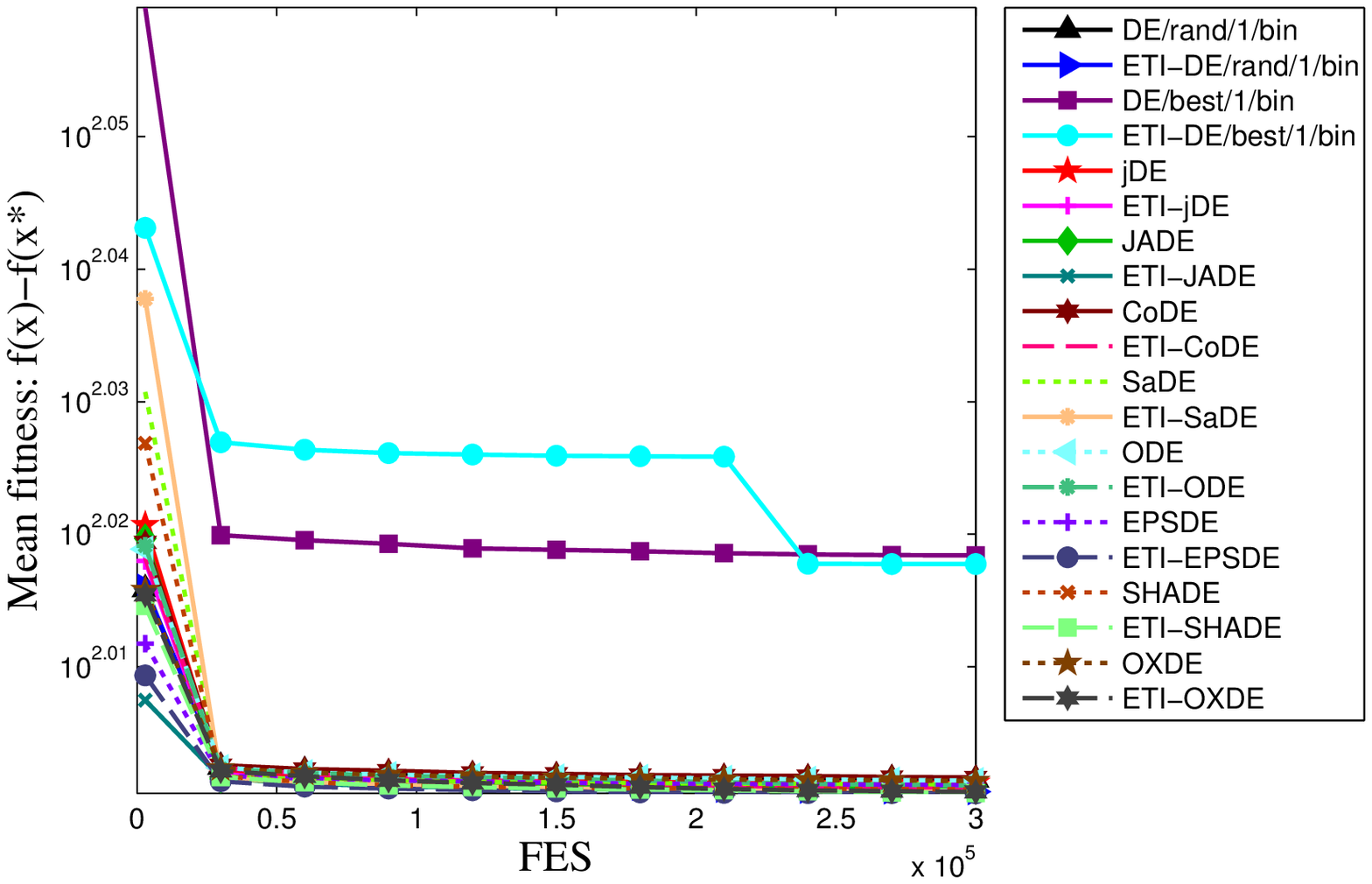}
\end{minipage}}
\caption{Evolution of the mean function error values obtained from the algorithms versus the number of FES on six 30-dimensional test functions. (a) F02; (b) F11; (c) F15; (d) F19; (e) F20; (f) F26.} \label{figcf}
\end{figure*}

\renewcommand\thefigure{S.\arabic{figure}}
\begin{figure*}[!htbp]
\centering
\subfigure[]{
\begin{minipage}[t]{0.4\textwidth}
\label{fig:subfig:a} 
\includegraphics[width=2.7in]{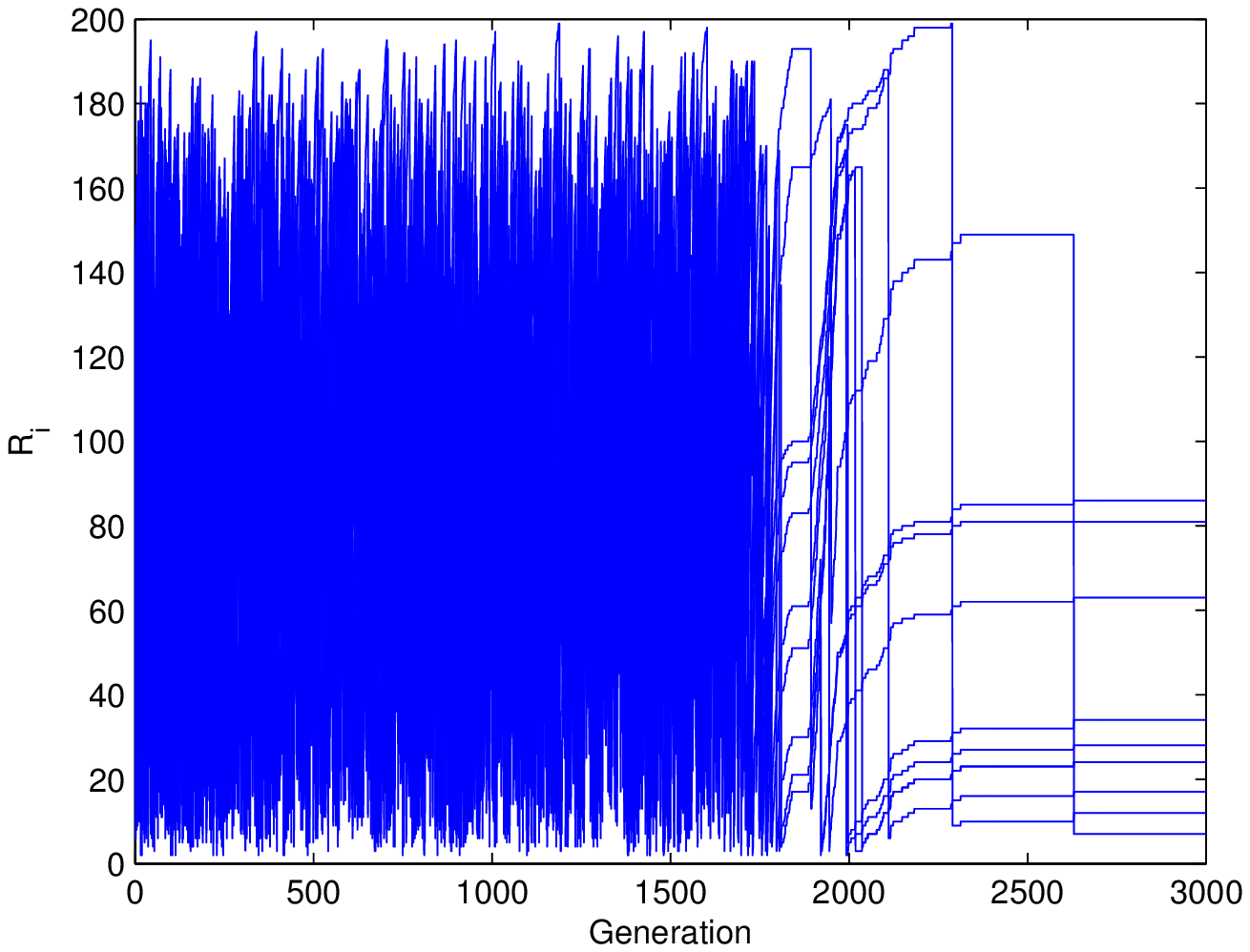}
\end{minipage}}
\centering
\subfigure[]{
\begin{minipage}[t]{0.4\textwidth}
\label{fig:subfig:b} 
\includegraphics[width=2.7in]{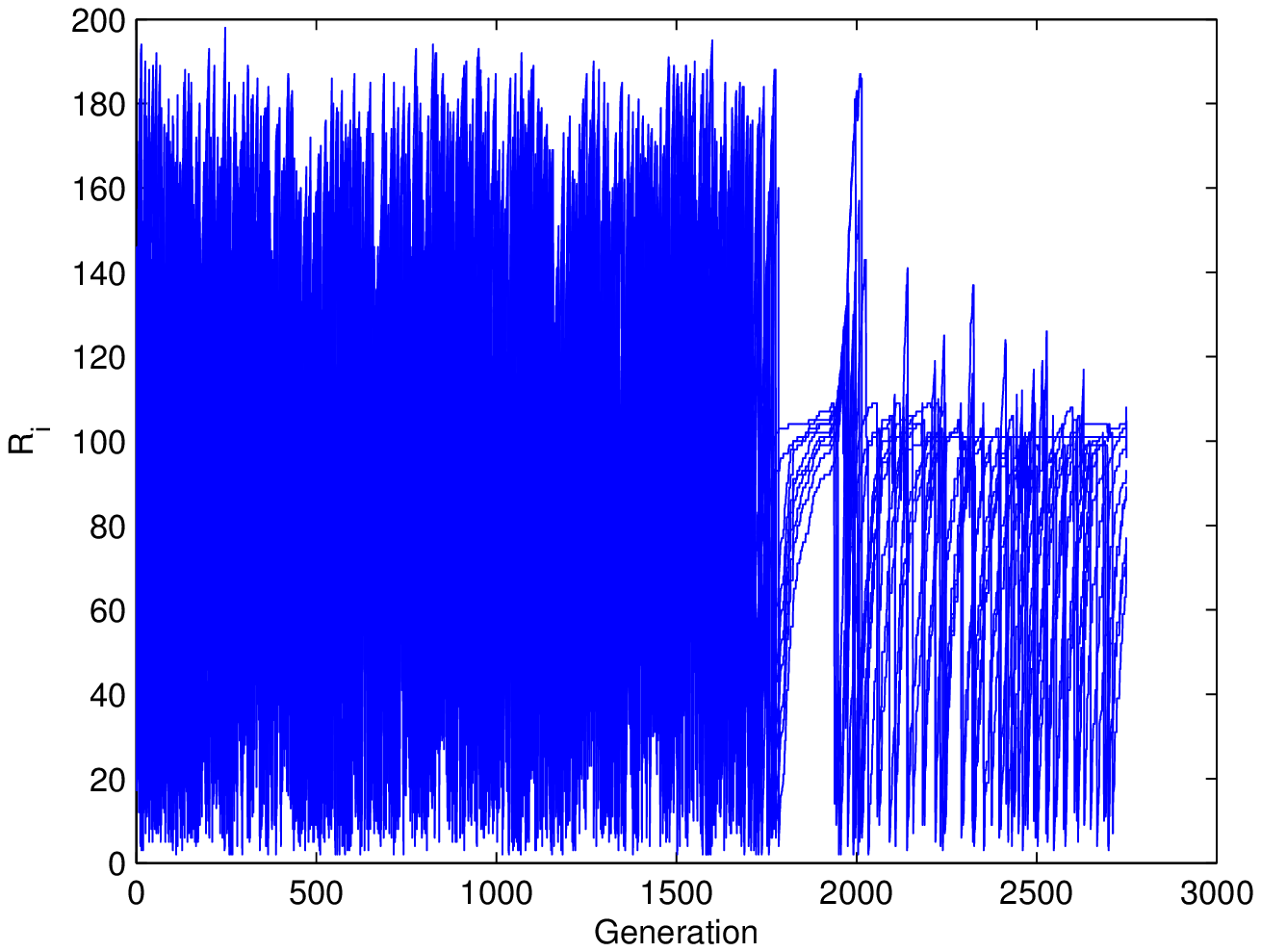}
\end{minipage}}
\centering
\subfigure[]{
\begin{minipage}[t]{0.4\textwidth}
\label{fig:subfig:c} 
\includegraphics[width=2.7in]{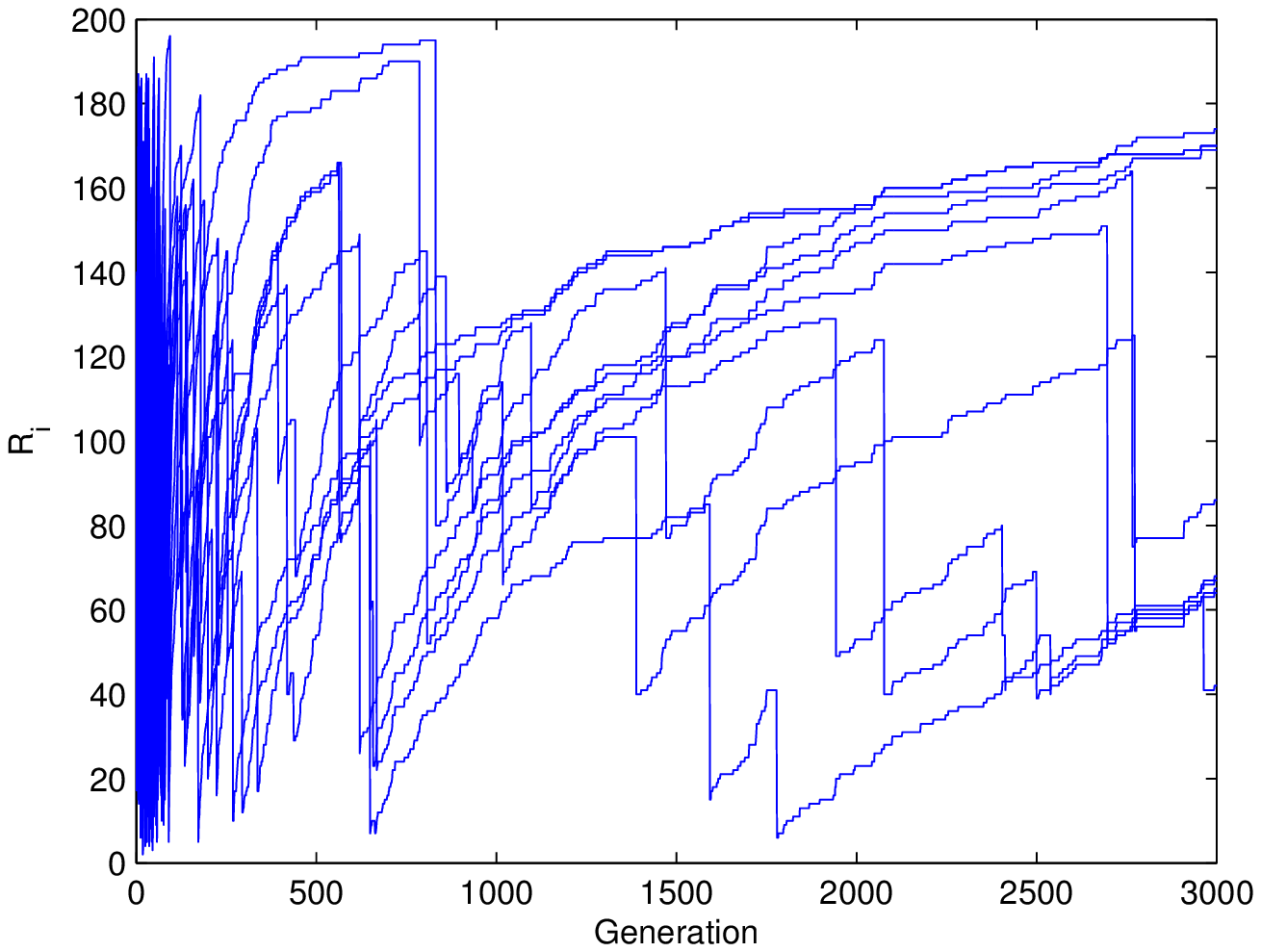}
\end{minipage}}
\centering
\subfigure[]{
\begin{minipage}[t]{0.4\textwidth}
\label{fig:subfig:d} 
\includegraphics[width=2.7in]{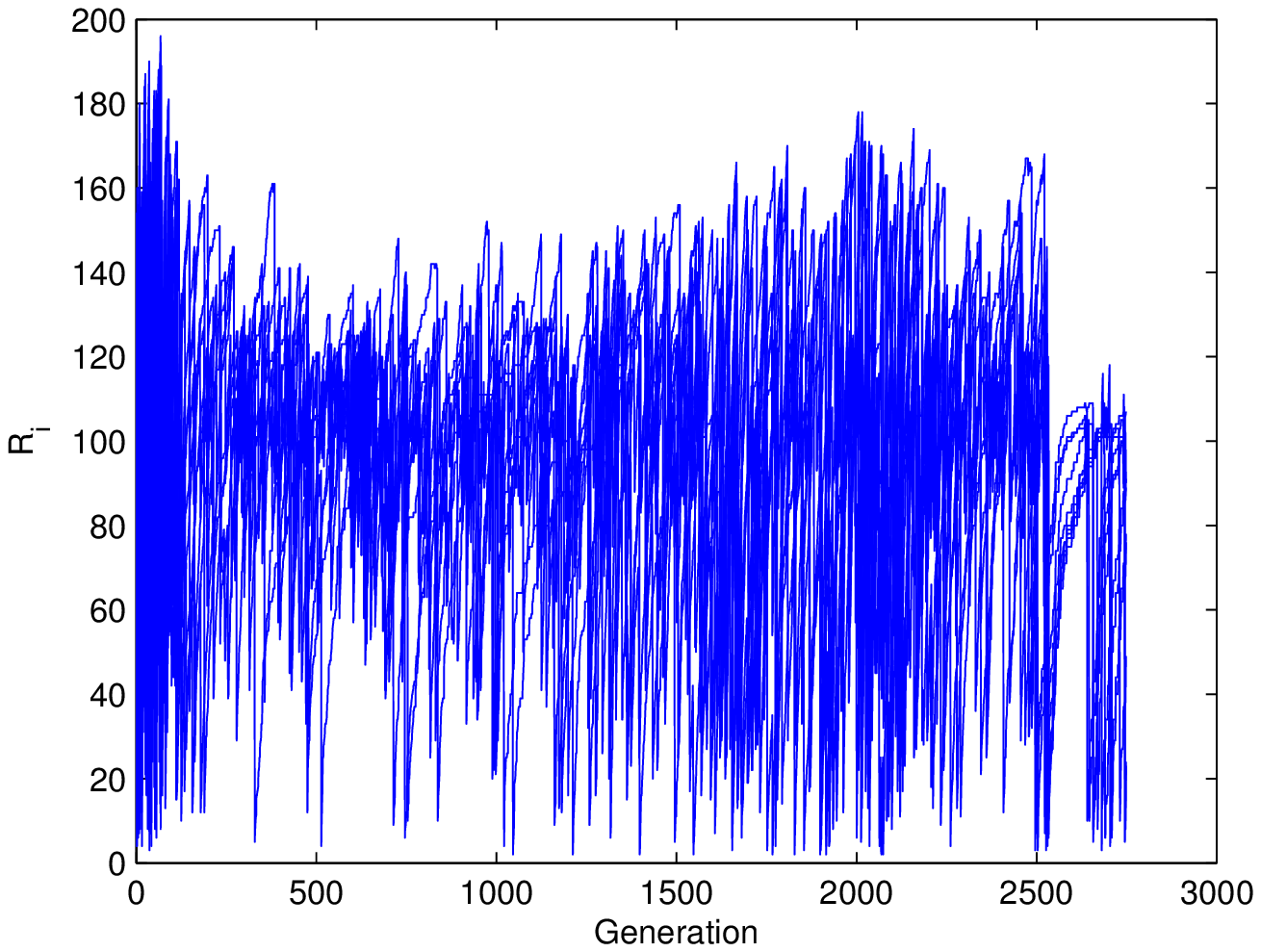}
\end{minipage}}
\caption{Evolution of $\emph{R}_\emph{i}$ by jDE and ETI-jDE on F02 and F13 at $\emph{D}=30$. (a) Evolution of ten random individuals' $\emph{R}_\emph{i}$ by jDE on F02. (b) Evolution of ten random individuals' $\emph{R}_\emph{i}$ by ETI-jDE on F02. (c) Evolution of ten random individuals' $\emph{R}_\emph{i}$ by jDE on F13. (d) Evolution of ten random individuals' $\emph{R}_\emph{i}$ by ETI-jDE on F13.} \label{figR}
\end{figure*}

\renewcommand\thefigure{S.\arabic{figure}}
\begin{figure*}[!htbp]
\begin{minipage}[t]{1\linewidth}
\centering
\includegraphics[width=16cm]{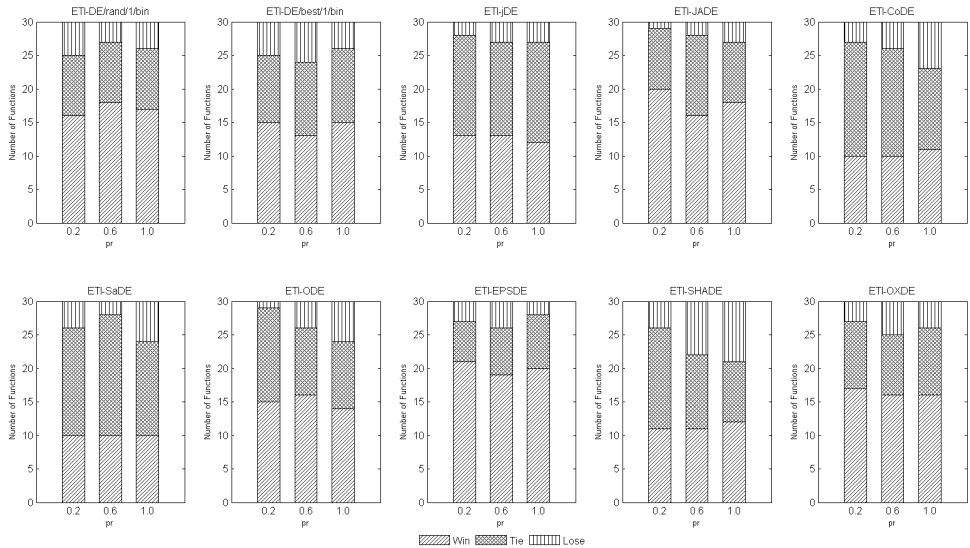}
\caption{The number of functions that ETI-DEs with different \emph{pr} values ($\emph{LN}=1,\emph{UN}=\emph{NP}$) are significantly better than, equal to and worse than the original DEs on CEC 2014 test suite at $\emph{D}=30$. (The results of adding the \emph{win/tie/lose} numbers for all the algorithms when using the same value of \emph{pr}: $\emph{pr}=0.2: 148/121/31; \emph{pr}=0.6: 142/117/41; \emph{pr}=1.0: 145/107/48$.)} \label{prplot}
\end{minipage}
\end{figure*}

\renewcommand\thefigure{S.\arabic{figure}}
\begin{figure*}[!htbp]
\begin{minipage}[t]{1\linewidth}
\centering
\includegraphics[width=16cm]{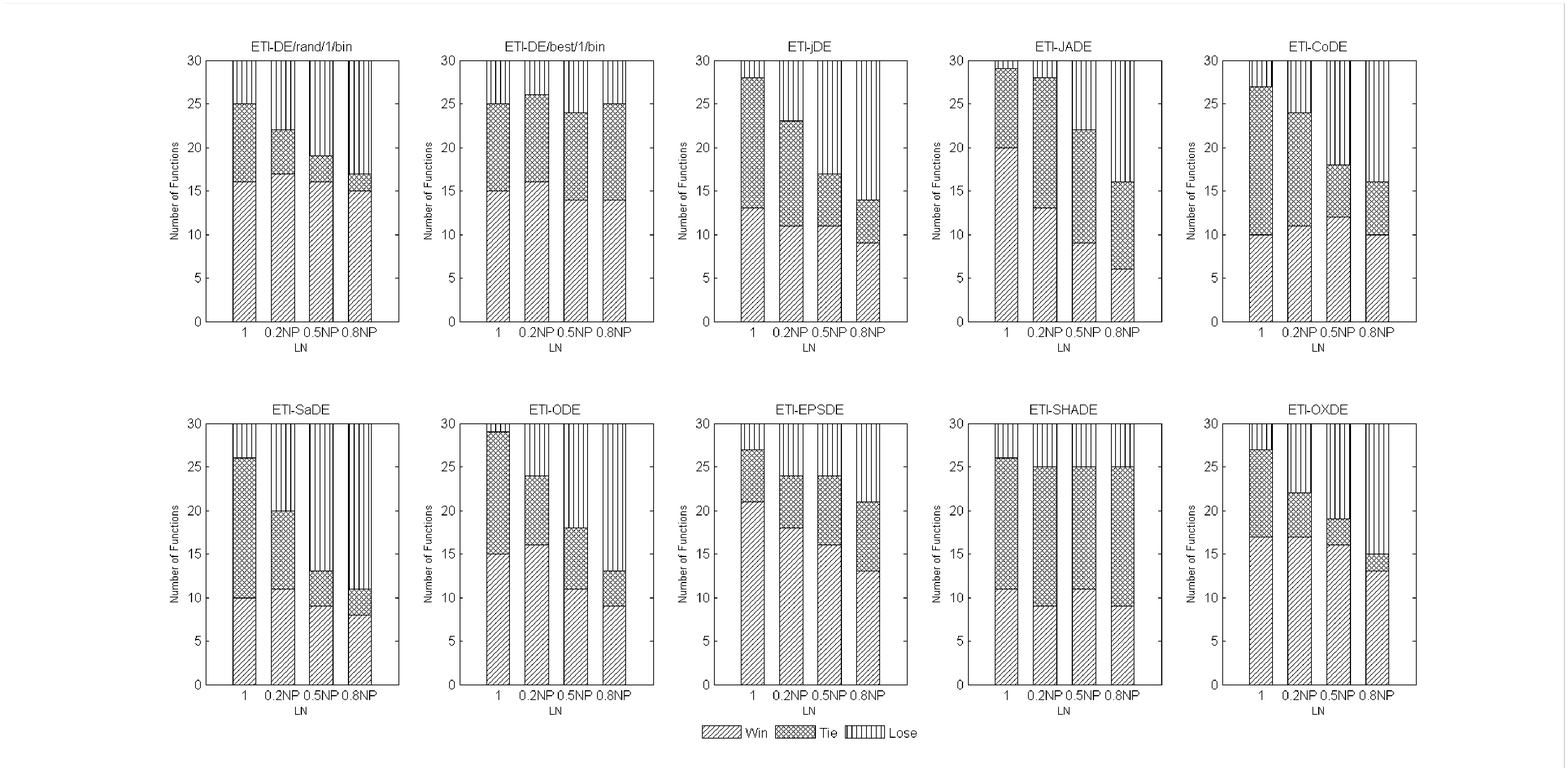}
\caption{The number of functions that ETI-DEs with different \emph{LN} values ($\emph{pr}=0.2,\emph{UN}=\emph{NP}$) are significantly better than, equal to and worse than the original DEs on CEC 2014 test suite at $\emph{D}=30$. (The results of adding the \emph{win/tie/lose} numbers for all the algorithms when using the same value of \emph{LN}: $\emph{LN}=1: 148/121/31; \emph{LN}=0.2\emph{NP}: 139/99/62; \emph{LN}=0.5\emph{NP}: 125/74/101; \emph{LN}=0.8\emph{NP}: 106/67/127$.)} \label{lnplot}
\end{minipage}
\end{figure*}

\renewcommand\thefigure{S.\arabic{figure}}
\begin{figure*}[!htbp]
\begin{minipage}[t]{1\linewidth}
\centering
\includegraphics[width=16cm]{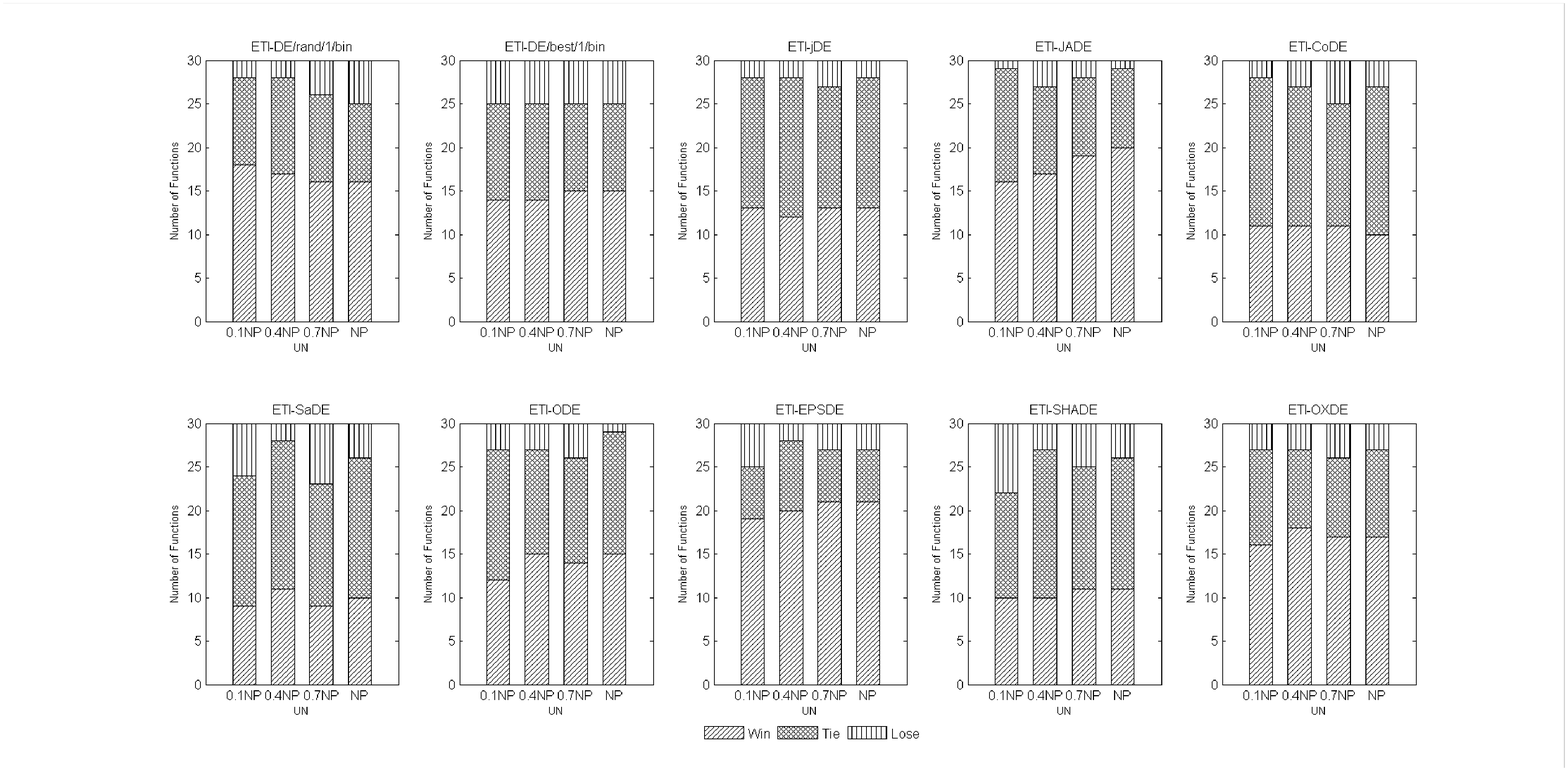}
\caption{The number of functions that ETI-DEs with different \emph{UN} values ($\emph{pr}=0.2,\emph{LN}=1$) are significantly better than, equal to and worse than the original DEs on CEC 2014 test suite at $\emph{D}=30$. (The results of adding the \emph{win/tie/lose} numbers for all the algorithms when using the same value of \emph{UN}: $\emph{UN}=0.1\emph{NP}: 138/125/37; \emph{UN}=0.4\emph{NP}: 145/127/28; \emph{UN}=0.7\emph{NP}: 146/112/42; \emph{UN}=\emph{NP}: 148/121/31$.)} \label{unplot}
\end{minipage}
\end{figure*}

\renewcommand\thefigure{S.\arabic{figure}}
\begin{figure*}[!htbp]
\begin{minipage}[t]{1\linewidth}
\centering
\includegraphics[width=12cm]{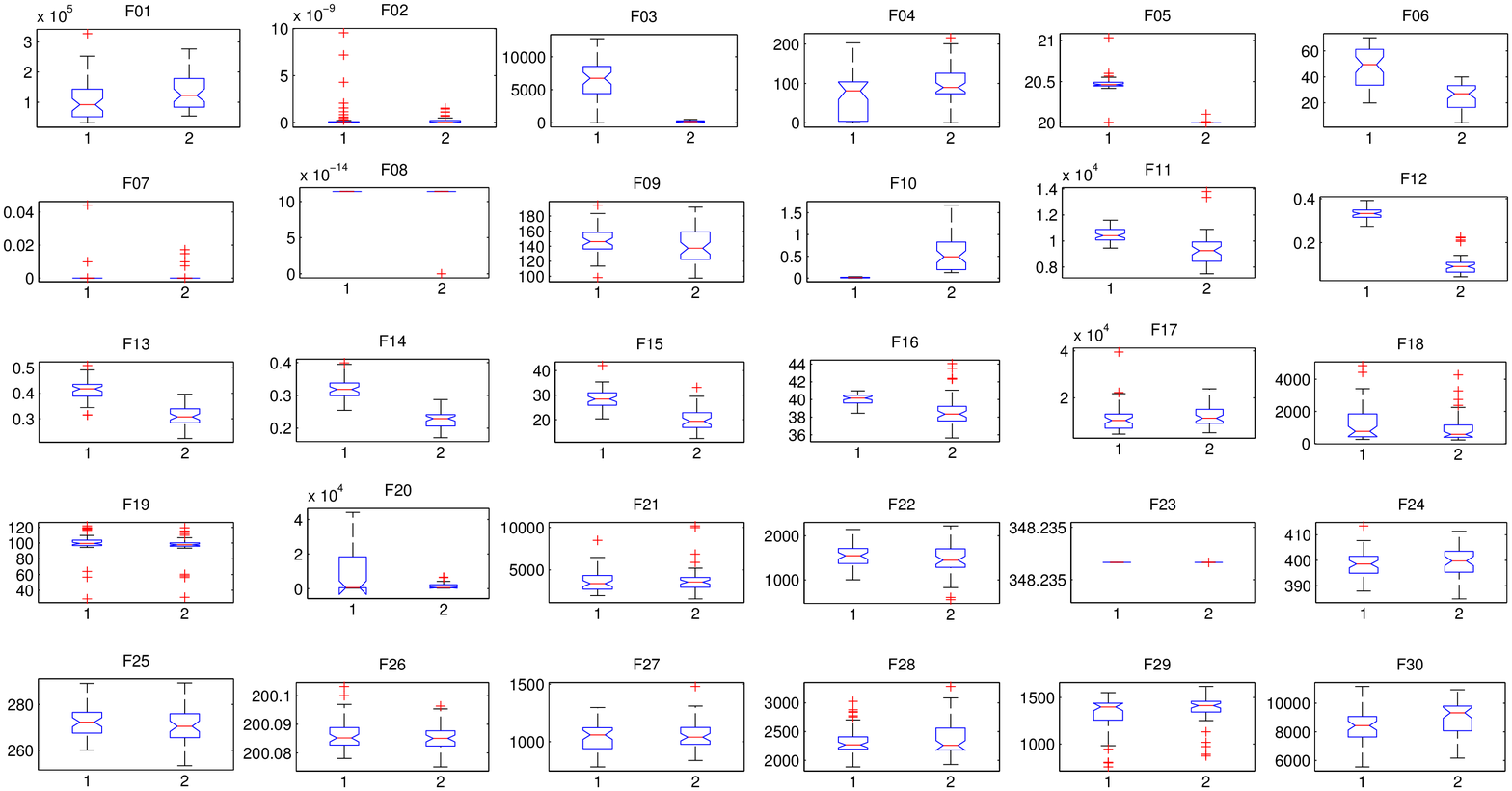}
\caption{Box plots for the results of JADE with/without ETI on CEC 2014 test suite at $\emph{D}=100$: 1--JADE; 2--ETI-JADE.} \label{jadebp100}
\end{minipage}
\end{figure*}

\renewcommand\thefigure{S.\arabic{figure}}
\begin{figure*}[!htbp]
\begin{minipage}[t]{1\linewidth}
\centering
\includegraphics[width=12cm]{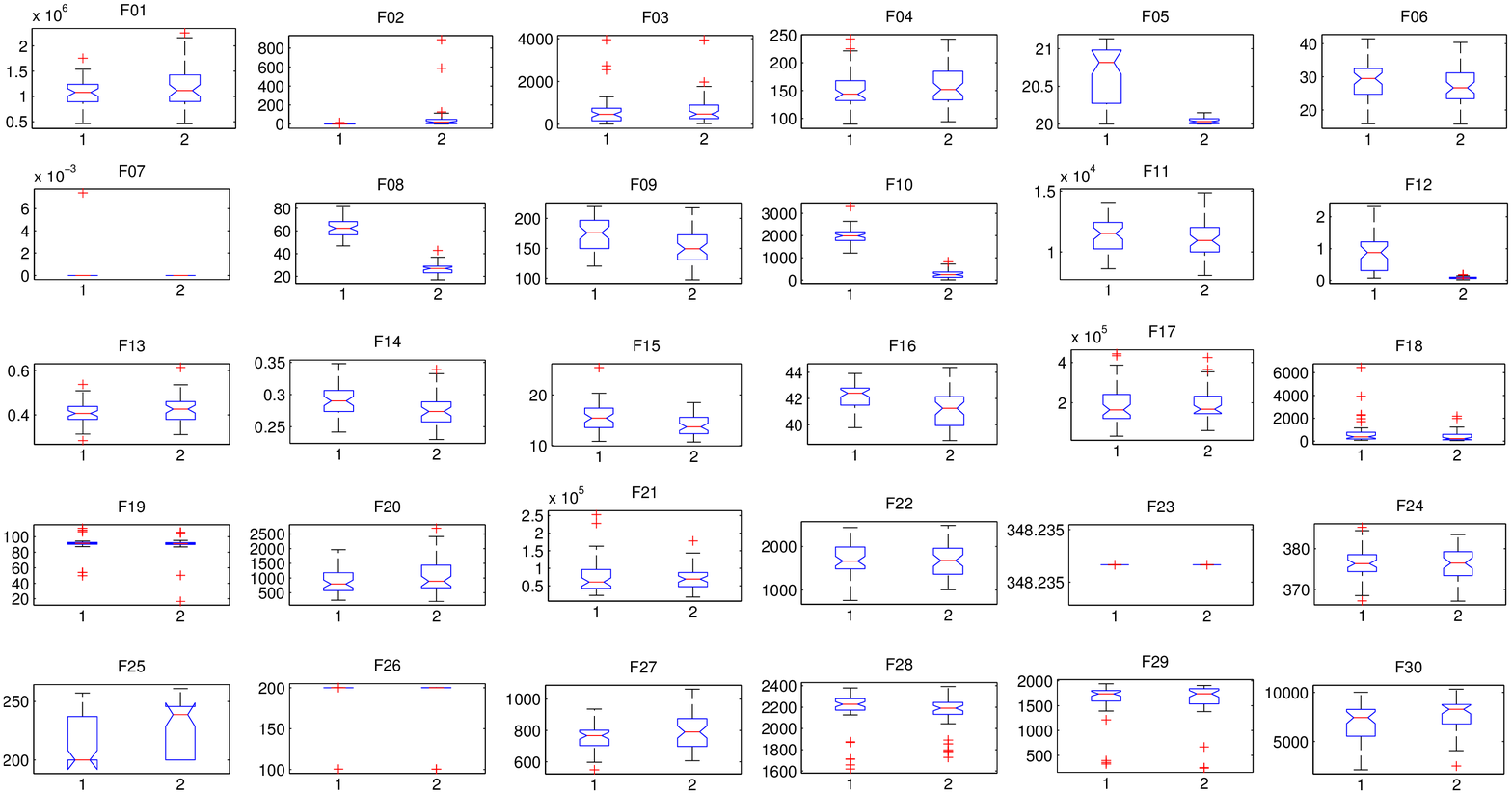}
\caption{Box plots for the results of CoDE with/without ETI on CEC 2014 test suite at $\emph{D}=100$: 1--CoDE; 2--ETI-CoDE.} \label{codebp100}
\end{minipage}
\end{figure*}

\renewcommand\thefigure{S.\arabic{figure}}
\begin{figure*}[!htbp]
\begin{minipage}[t]{1\linewidth}
\centering
\includegraphics[width=12cm]{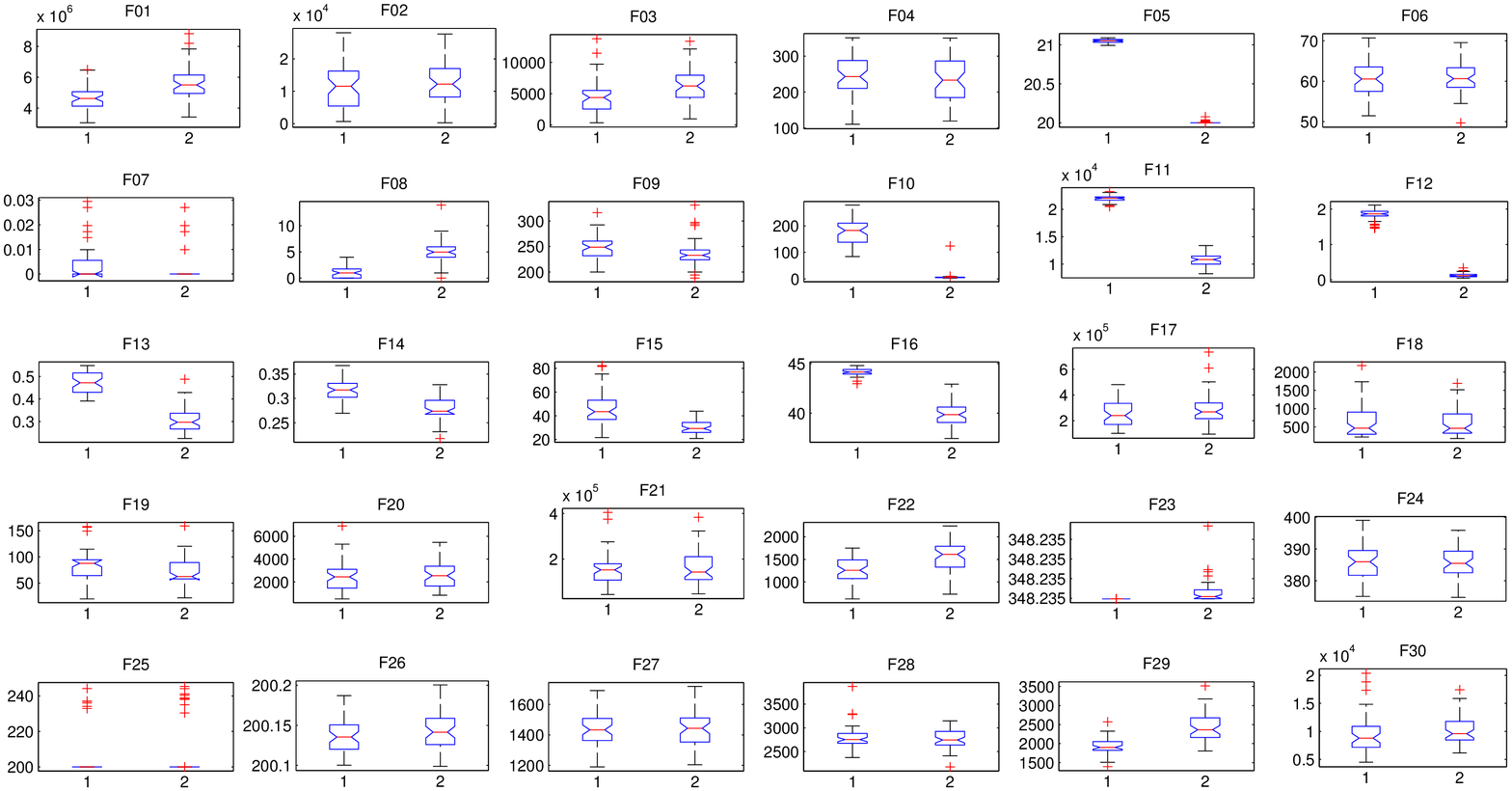}
\caption{Box plots for the results of SaDE with/without ETI on CEC 2014 test suite at $\emph{D}=100$: 1--SaDE; 2--ETI-SaDE.} \label{sadebp100}
\end{minipage}
\end{figure*}

\renewcommand\thefigure{S.\arabic{figure}}
\begin{figure*}[!htbp]
\begin{minipage}[t]{1\linewidth}
\centering
\includegraphics[width=12cm]{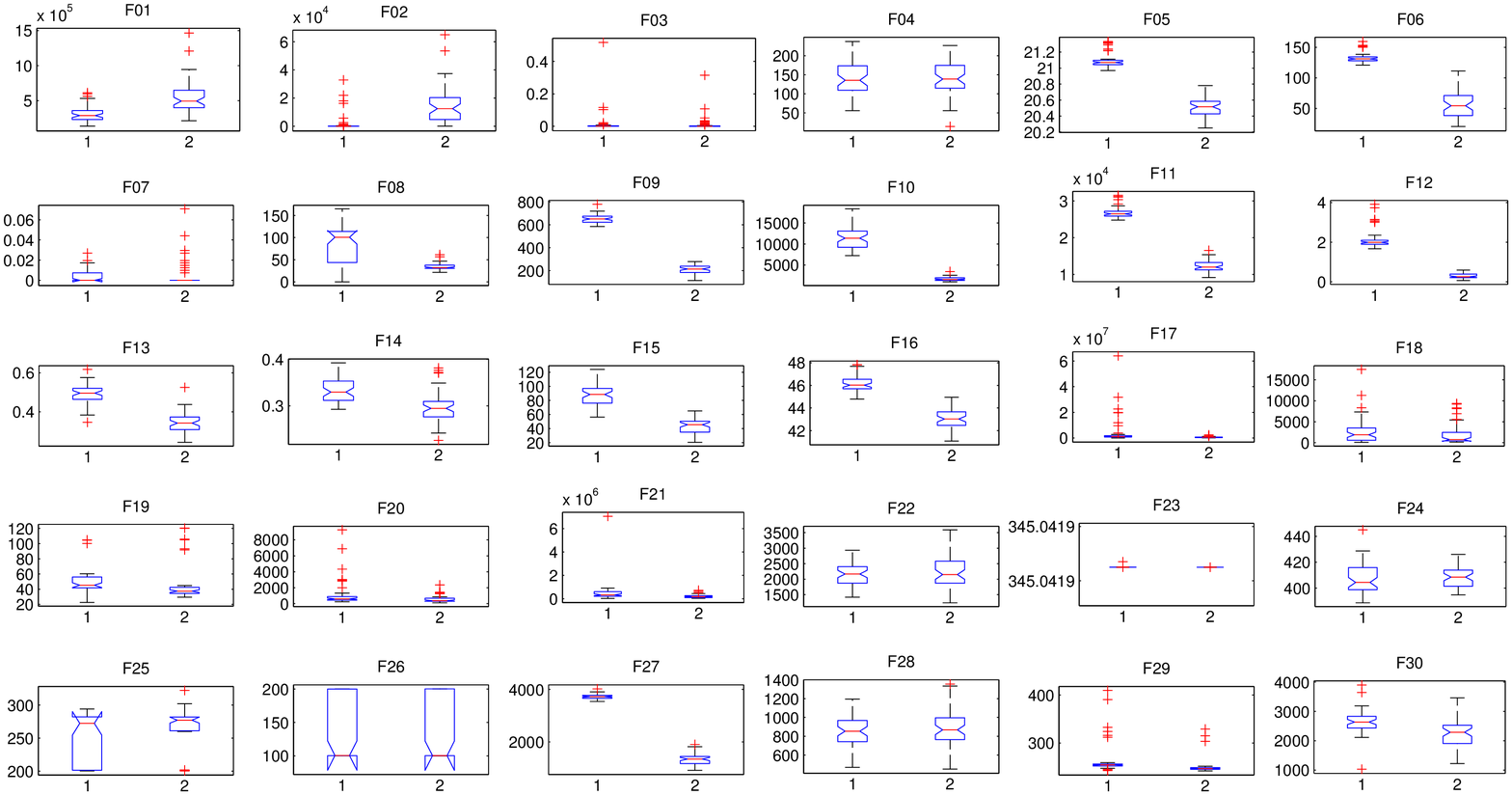}
\caption{Box plots for the results of EPSDE with/without ETI on CEC 2014 test suite at $\emph{D}=100$: 1--EPSDE; 2--ETI-EPSDE.} \label{epsdebp100}
\end{minipage}
\end{figure*}

\renewcommand\thefigure{S.\arabic{figure}}
\begin{figure*}[!htbp]
\begin{minipage}[t]{1\linewidth}
\centering
\includegraphics[width=12cm]{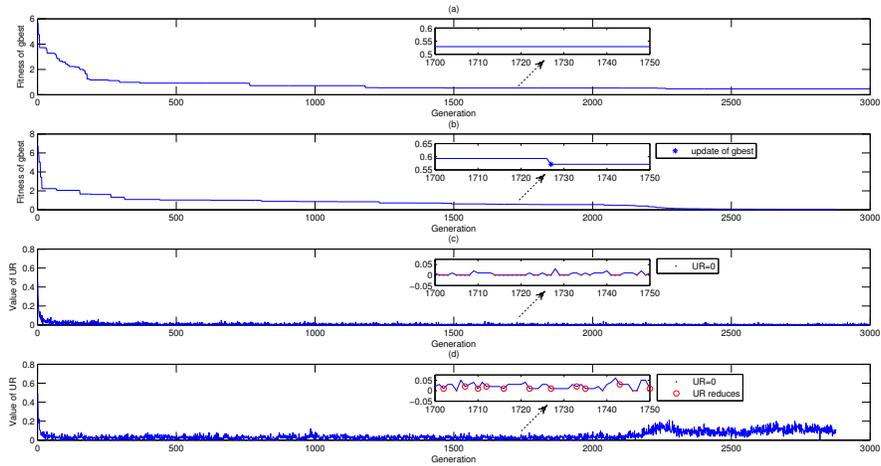}
\caption{Working mechanism of ETI by jDE and ETI-jDE on F12 at $\emph{D}=30$. (a) Change of the fitness value of \emph{gbest} of F12 optimized by jDE. (b) Change of the fitness value of \emph{gbest} of F12 optimized by ETI-jDE. (c) Change of the value of \emph{UR} of F12 optimized by jDE. (d) Change of the value of \emph{UR} of F12 optimized by ETI-jDE.} \label{wm}
\end{minipage}
\end{figure*}

\end{document}